\documentclass[11pt]{article}
\usepackage[utf8]{inputenc}
\usepackage{amsmath,amsfonts,amsthm,amssymb,url}
\usepackage{xcolor}
\usepackage{fullpage}
\usepackage{thmtools, thm-restate}
\usepackage{algorithmic}
\usepackage{algorithm}
\usepackage[T1]{fontenc}
\usepackage[colorlinks=true, linkcolor=red, urlcolor=blue, citecolor=gray]{hyperref}
\usepackage{mathtools}
\usepackage{graphicx}
\usepackage{caption}
\usepackage{tikz}
\usetikzlibrary{calc,arrows.meta,positioning}
\usetikzlibrary{decorations.pathreplacing}
\usetikzlibrary{arrows}

\usepackage{booktabs}       
\usepackage{nicefrac}       
\usepackage{microtype}      
\usepackage{xspace}

\usepackage{latexsym}
\usepackage{subfigure}
\usepackage{mathtools}
\usepackage{xcolor}
\usepackage{mleftright}
\usepackage[normalem]{ulem}
\usepackage{physics}
\usepackage{array}
\usepackage{makecell}
\usepackage{multicol}
\usepackage{enumitem}

\usepackage{natbib}

  \usepackage{nth}
  \usepackage{intcalc}

\definecolor{darkgreen}{HTML}{1b7837}

\definecolor{MyBlue}{HTML}{069af3}
\definecolor{MyGreen}{HTML}{15b01a}
\definecolor{MyPurple}{HTML}{9a0eea}

\newcommand{\bv}[1]{\mathbf{#1}}

\newcommand{\nystrom}{Nystr\"om\xspace}

\newcommand{\R}{\mathbb{R}}

\DeclareUnicodeCharacter{2212}{-}

\newcommand\blfootnote[1]{%
  \begingroup
  \renewcommand\thefootnote{}\footnote{#1}%
  \addtocounter{footnote}{-1}%
  \endgroup
}

\renewcommand{\cite}{\citep}

\title{Sublinear Time Approximation of Text Similarity Matrices}

\author{Archan Ray, Nicholas Monath$^\dag$, Andrew McCallum, Cameron Musco\\University of Massachusetts Amherst\\\texttt{\{ray, nmonath, mccallum, cmusco\}@cs.umass.edu}}
\date{\today}

\usepackage{sidecap}
\usepackage{adjustbox}
\begin{document}
\sloppy
\maketitle

\begin{abstract}
\blfootnote{$\dag$ Now at Google.}

We study algorithms for approximating pairwise similarity matrices that arise in natural language processing. Generally, computing a similarity matrix for $n$ data points requires $\Omega(n^2)$ similarity computations.
This quadratic scaling is a significant bottleneck, especially when similarities are computed via expensive functions, e.g., via transformer models.  Approximation methods reduce this quadratic complexity, often by using a small subset of exactly computed similarities to approximate the remainder of the complete pairwise similarity matrix.

Significant  work focuses on the efficient approximation of positive semidefinite (PSD) similarity matrices, which arise e.g., in kernel methods. However, much less is understood about indefinite (non-PSD) similarity matrices, which often  arise in  NLP. Motivated by the observation that many of these matrices are still somewhat close to PSD, we introduce a generalization of the popular \emph{\nystrom method} to the indefinite setting. Our algorithm can be applied to any similarity matrix and runs in sublinear time in the size of the matrix, producing a rank-$s$ approximation with just $O(ns)$ similarity computations.

We show that our method, along with a simple variant of CUR decomposition, performs very well in approximating a variety of similarity matrices arising in NLP tasks. We demonstrate high accuracy of the approximated similarity matrices in the downstream tasks of document classification, sentence similarity, and cross-document coreference. 
\end{abstract}


\section{Introduction}\label{sec:introduction}

Many machine learning tasks center around the computation of pairwise similarities between data points using an appropriately chosen similarity function. E.g., in kernel methods, a non-linear kernel inner product is used to measure similarity, and often to construct a pairwise kernel similarity matrix. 
In natural language processing, document or sentence similarity functions (e.g., cross-encoder transformer models \cite{devlin2018bert} or word mover’s distance \cite{piccoli2014generalized,kusner2015word})) are key components of cross-document coreference \cite{cattan2020streamlining} and passage retrieval for question answering \cite{karpukhin-etal-2020-dense}. String-similarity functions are used to model name aliases \cite{tam-etal-2019-optimal} and for morphology \cite{rastogi2016weighting}.

Computing all pairwise similarities for a data set with $n$ points requires $\Omega(n^2)$ similarity computations. This can be a major runtime bottleneck, especially when each computation requires the evaluation of a neural network or other expensive operation. 
One approach to avoid this bottleneck is to produce a compressed approximation to the $n \times n$ pairwise similarity matrix $\bv K$ for the data set, but avoid ever fully forming this matrix and run in sub-quadratic time (i.e., with running time less than $O(n^2)$, or sublinear in the size of $\bv{K}$). 
The compressed approximation, $\bv{\tilde K}$, can be used in place of  $\bv{K}$ to quickly access approximate pairwise similarities, and in methods for near neighbor search, clustering, and regression, which would typically involve $\bv{K}$.

\subsection{Existing Methods}

Similarity matrix approximation is very well-studied, especially in the context of accelerating kernel methods and Gaussian process regression. Here, $\bv{K}$ is typically positive semidefinite (PSD). This structure is leveraged by techniques like the random Fourier features and \nystrom methods \cite{rahimi2007random,le2013fastfood,williams2001using,yang2012nystrom}, which approximate $\bv{K}$ via a rank-$s$ approximation $\bv{\tilde K} = \bv{Z} \bv{Z}^T$, for $s \ll n$ and $\bv{Z} \in \R^{n \times s}$. These methods have runtimes scaling linearly in $n$ and sublinear in the  matrix size. They have been very successful in practice \cite{huang2014kernel,meanti2020kernel}, and often come with  strong theoretical bounds \cite{gittens2016revisiting,musco2017recursive,musco2017sublinear}. 

Unfortunately, most similarity matrices arising in natural language processing, such as those based on cross-encoder transformers \cite{devlin2018bert} or word mover’s distance \cite{piccoli2014generalized}, are \emph{indefinite} (i.e., non-PSD). For such matrices, much less is known. 
Sublinear time methods have been studied for certain classes of similarities  \cite{bakshi18sublinear,oglic2019scalable,indyk2019sample}, but do not apply more generally.
Classic techniques like low-rank approximation via the SVD or fast low-rank approximation via random sketching \cite{frieze2004fast,sarlos2006improved,drineas2008relative} generally must form all of $\bv{K}$  to approximate it, and so run in $\Omega(n^2)$ time. There are generic sublinear time sampling methods, like CUR decomposition \cite{drineas2006fast,wang2016towards}, which are closely related to \nystrom approximation. However, as we will see, the performance of these methods varies greatly depending on the application. 

\subsection{Our Contributions}

\noindent\textbf{Algorithmic.} Our first contribution is a simple variant of the \nystrom method  that applies to symmetric indefinite similarity matrices\footnote{While \emph{asymmetric} similarity matrices do arise, we focus on the symmetric case. In our experiments, simply symmetrizing and then approximating these matrices yields good performance.}. The \nystrom method \cite{williams2001using} approximates a PSD similarity matrix $\bv K$ by sampling a set of $s \ll n$ \emph{landmark points} from the dataset, computing their similarities with all other points (requiring $O(ns)$ similarity computations), and then using this sampled set of similarities to reconstruct all of $\bv{K}$. See Sec. \ref{sec:mot}.

Our algorithm is motivated by the observation that many indefinite similarity matrices arising in NLP  are \emph{somewhat close to PSD} -- they have relatively few negative eigenvalues.
Thus, a natural approach would be simply to apply \nystrom to them. However, even for matrices with just a few small negative eigenvalues, this fails completely.
We instead show how to `minimally correct' our matrix to be closer to PSD, before applying Nystr\"{o}m. Specifically, we apply an eigenvalue shift based on the minimum eigenvalue of a small random principal submatrix of $\bv{K}$. We call our method \emph{Submatrix-Shifted \nystrom}, or \emph{SMS-\nystrom}. SMS-\nystrom is extremely efficient, and, while we do not give rigorous approximation bounds, it recovers the strong performance of the \nystrom method on many near PSD-matrices.

\smallskip

\noindent\textbf{Empirical.} Our second contribution is a systematic evaluation of a number of sublinear time matrix approximation methods in NLP applications. We consider three applications involving indefinite similarity matrices:
1) computing document embeddings using word mover's distance \cite{kusner2015word}, for four different text classification tasks; 2) approximating  similarity matrices generated using cross-encoder BERT \cite{devlin2018bert} and then comparing performance in  three GLUE tasks: STS-B \cite{cer2017semeval}, MRPC \cite{mrpc} and RTE \cite{rte}, which require predicting similarity, semantic equivalence, and entailment between sentences; 3) approximating the similarity function used to determine coreference relationships across documents in a corpus of news articles mentioning entities and events \cite{cybulska2014using,cattan2020streamlining}.

We show that both {SMS-\nystrom}, and a simple variant of CUR decomposition yield accurate approximations that maintain downstream task performance in all these tasks while greatly reducing the time and space required as compared to the exact similarity matrix. They typically significantly outperform the classic \nystrom method and other CUR variants.

\subsection{Other Related Work}

Our work fits into a vast literature on randomized methods for matrix approximation \cite{mahoney2011randomized,woodruff2014sketching}. There is significant work on different sampling distributions and theoretical bounds for both the Nystr\"{o}m  and CUR methods \cite{goreinov1997theory,drineas2005nystrom, drineas2008relative,zhang2008improved,kumar2012sampling,wang2013improving,talwalkar2014matrix}. However, more advanced methods generally require reading all of $\bv{K}$ and so do not avoid $\Omega(n^2)$ time. In fact, any method with non-trivial worst-case  guarantees on general matrices cannot run less than $O(n^2)$ time. If the entire mass of the matrix is placed on a single  entry, all entries must be accessed  to find it.

A number of works apply \nystrom  variants to indefinite matrices. \citet{belongie2002spectral} show that the \nystrom method can be effectively applied to eigenvector approximation for indefinite matrices, specifically in application to  spectral partitioning. However, they do not investigate the behavior of the method in approximating the similarity matrix itself.
\citet{gisbrecht2015metric} shows that, in principal, the classic \nystrom approximation converges to the true matrix  when the similarity function is continuous over $\R$. However, we observe poor finite sample performance of this method on text similarity matrices. 
Other work exploits assumptions on the input  points -- e.g. that they lie in a small number of labeled classes, or in a low-dimensional space where distances correlate with the similarity  \cite{schleif2018supervised}. This later assumption is made implictly in recent work on anchor-net based \nystrom  \cite{cai2021fast}, and while it may hold in many settings, in  NLP applications, it is often not clear how to find such a low-dimensional representation.

By removing the above assumptions, our work is well suited for applications in NLP, which often feed two inputs (e.g., sentences) into a neural network (e.g., transformer or MLP) to compute similarities.

There is also significant related work on modifying indefinite similarity matrices to be PSD, including via eigenvalue transformations and shifts \cite{chen2009learning,gisbrecht2015metric}. These modifications would allow the matrix to be approximated with the classic \nystrom method. However, this work does not focus on sublinear runtime, typically using modifications that require $\Omega(n^2)$ time.

Finally, outside of similarity matrix approximation, there are many methods that seek to reduce the cost of similarity computation.
One approach is to reduce the number of similarity computations.
Examples include locality sensitive hashing \cite{gionis1999similarity,lv2007multi}, distance preserving embeddings \cite{hwang2012fast}, and graph based algorithms \cite{orchard1991fast,dong2011efficient} for near-neighbor search. 
Another approach is to reduce the cost of each similarity computation, e.g., via model distillation for cross-encoder-based similarity   \cite{sanh2019distilbert,jiao2019tinybert,michel2019sixteen,lan2019albert,zafrir2019q8bert, humeau2019poly}. However, model distillation requires significant additional training time to fit the reduced model, unlike our proposed approach which requires only $O(ns)$ similarity computations.
There is also work on random features methods and other alternatives to expensive similarity functions, such as those based on the word-movers distance \cite{cuturi2013sinkhorn,wu2018word,wu2019scalable}.

\section{Submatrix-Shifted \nystrom}\label{sec:mot}

In this section, we introduce the \nystrom method for PSD matrix approximation, and describe our modification of this method for application to indefinite similarity matrices.

\subsection{The \nystrom Method}

Let $\mathcal{X} = \{x_i\}_{i=1}^n$ be a dataset with $n$ datapoints, $\Delta: \mathcal{X}\times\mathcal{X}\to\mathbb{R}$ be a similarity function, and $\bv K\in \mathbb{R}^{n\times n}$ be the corresponding similarity matrix with
$\bv K_{ij} = \Delta(x_i, x_j)$.

The \nystrom method samples $s$ landmark points -- let $\bv{S} \in \mathbb{R}^{n\times s}$ be the matrix performing this sampling. 
$\bv S$ has a single randomly positioned $1$ in each column. Thus $\bv{KS}$ is an $\mathbb{R}^{n \times s}$ submatrix of $\bv K$ consisting of randomly sampled columns corresponding to the similarities between all $n$ datapoints and the $s$ landmark points. The key idea is to approximate all pairwise similarities using just this sampled set. In particular, the \nystrom approximation of $\bv{K}$ is given as:
\begin{align}
    \label{eq: nystrom approximation}
    \tilde{\bv{K}} = \bv{KS}(\bv{S}^T\bv{KS})^{-1}\bv{S}^T\bv{K}.
\end{align}

\noindent\textbf{Running Time.} Observe that the \nystrom approximation of  \eqref{eq: nystrom approximation} requires just $O(ns)$ evaluations of the similarity function to compute $\bv{KS} \in \R^{n \times s}$.
We typically do not form $\bv{\tilde K}$ directly, as it would take at least $n^2$ time to even write down. 
Instead, we store this matrix in `factored form', computing 
    $\bv{Z} = \bv{KS}(\bv{S}^T\bv{KS})^{-1/2}.$
In this way, we have $\bv{ZZ}^T = \bv{\tilde K}$. I.e., the approximate similarity between points $x_i$ and $x_j$ is simply the inner product between the $i^{th}$ and $j^{th}$ rows of $\bv{Z}$, which can be thought of as embeddings of the points into $\R^s$.
Computing $\bv{Z}$ requires  
computing $(\bv{S}^T  \bv{K} \bv{S})^{-1/2}$ --  the matrix squareroot of $(\bv{S}^T \bv{KS})^{-1}$ which takes $O(s^3)$ time using e.g., Cholesky decomposition\footnote{If $\bv{S}^T\bv{KS}$ is singular, the pseudoinverse $(\bv{S}^T\bv{KS})^+$ can be used.}. Multiplying by $\bv{KS}$ then takes $O(ns^2)$ time, which is the dominant cost since $n > s$. 

\smallskip

\noindent\textbf{Intuition.}
In \eqref{eq: nystrom approximation}, $\bv{S}^T\bv{KS} \in \R^{s \times s}$ is the principal submatrix of $\bv{K}$ containing the similarities between the landmark points themselves.
To gain some intuition behind the approximation, consider removing the  $(\bv{S}^T\bv{KS})^{-1}$ term and approximating $\bv{K}$ with $\bv{KS S}^T \bv{K}$. That is, we approximate the similarity between any  two points $x_i$ and $x_j$ by the inner product between their corresponding rows in $\bv{K}\bv{S}$ -- i.e. the vector in $\R^s$ containing their similarities with the landmarks. This would be a reasonable approach -- when $x_i$ and $x_j$ are more similar, we expect these rows to have higher dot products.

The $(\bv{S}^T \bv{KS})^{-1}$ term intuitively `corrects for' similarities between the landmark points. Formally, when $\bv{K}$ is PSD, it can be written as $\bv{K} = \bv{BB}^T$ for some matrix $\bv{B} \in \mathbb{R}^{n \times n}$. Thus $\bv{K}_{ij} = \langle \bv{b}_i, \bv{b}_j \rangle$. 
Equation \eqref{eq: nystrom approximation} is equivalent to  projecting all rows of $\bv{B}$ onto the subspace spanned by the rows corresponding to the landmark points to produce $\bv{\tilde B}$, and then letting $\bv{\tilde K} = \bv{\tilde B} \bv{\tilde B}^T.$ If e.g., $\rank(\bv{K}) \le s$, then $\rank(\bv{B}) = \rank(\bv{K}) \le s$ and so as long as the rows of $\bv{B}$ corresponding to the landmark points are linearly independent, we will have $\bv{\tilde B} = \bv{B}$ and thus $\bv{\tilde K} = \bv{K}$. If $\bv{K}$ is close to low-rank, as is often the case in practice, $\bv{\tilde K}$ will still generally yield a very good approximation. 

\subsection{\nystrom for Indefinite Matrices}\label{sec:obv}

Our extension of the \nystrom method to indefinite matrices is motivated by two observations.

\begin{figure*}[h]
    \centering%
    \includegraphics[width=0.32\textwidth]{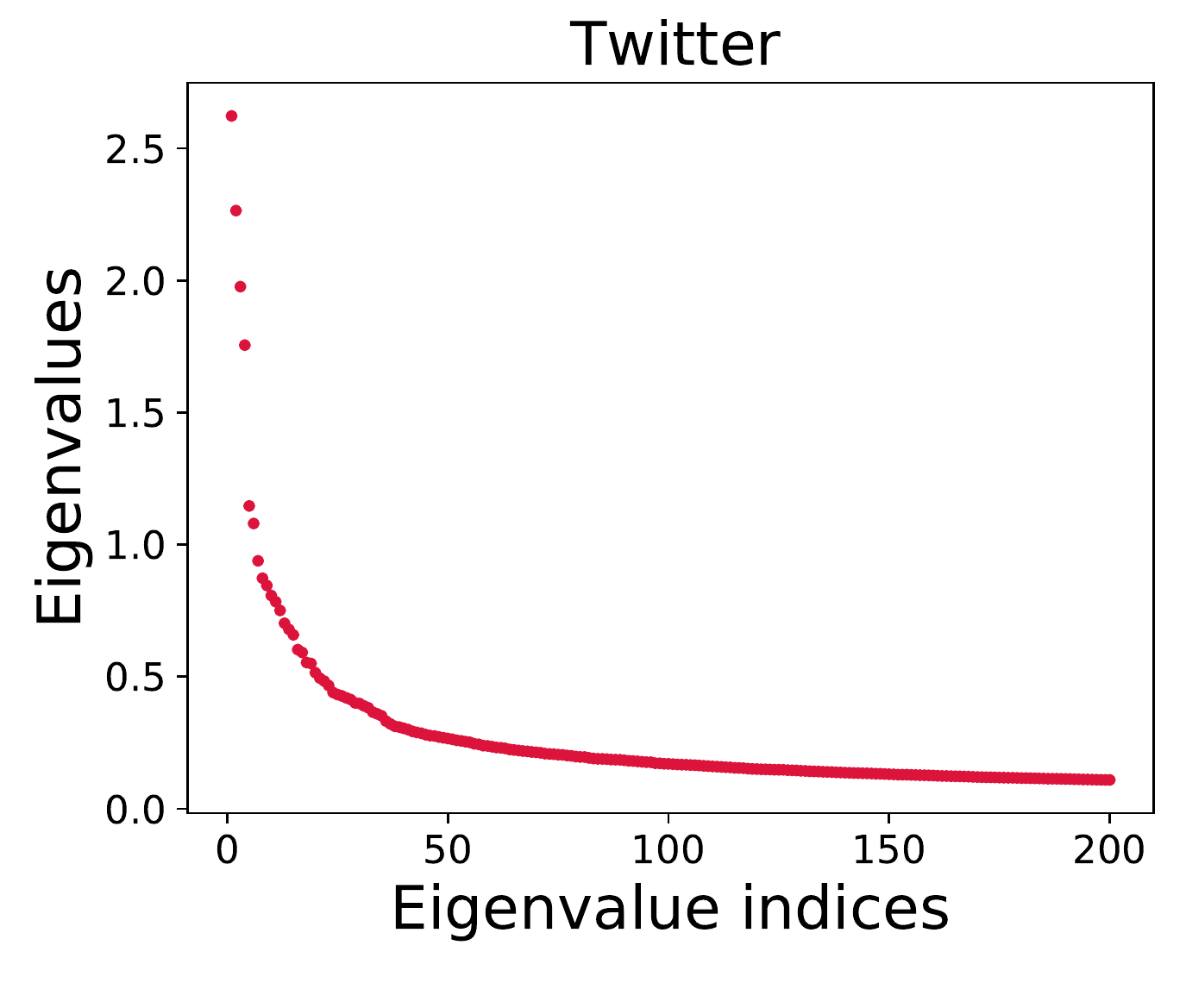}
    \includegraphics[width=0.32\textwidth]{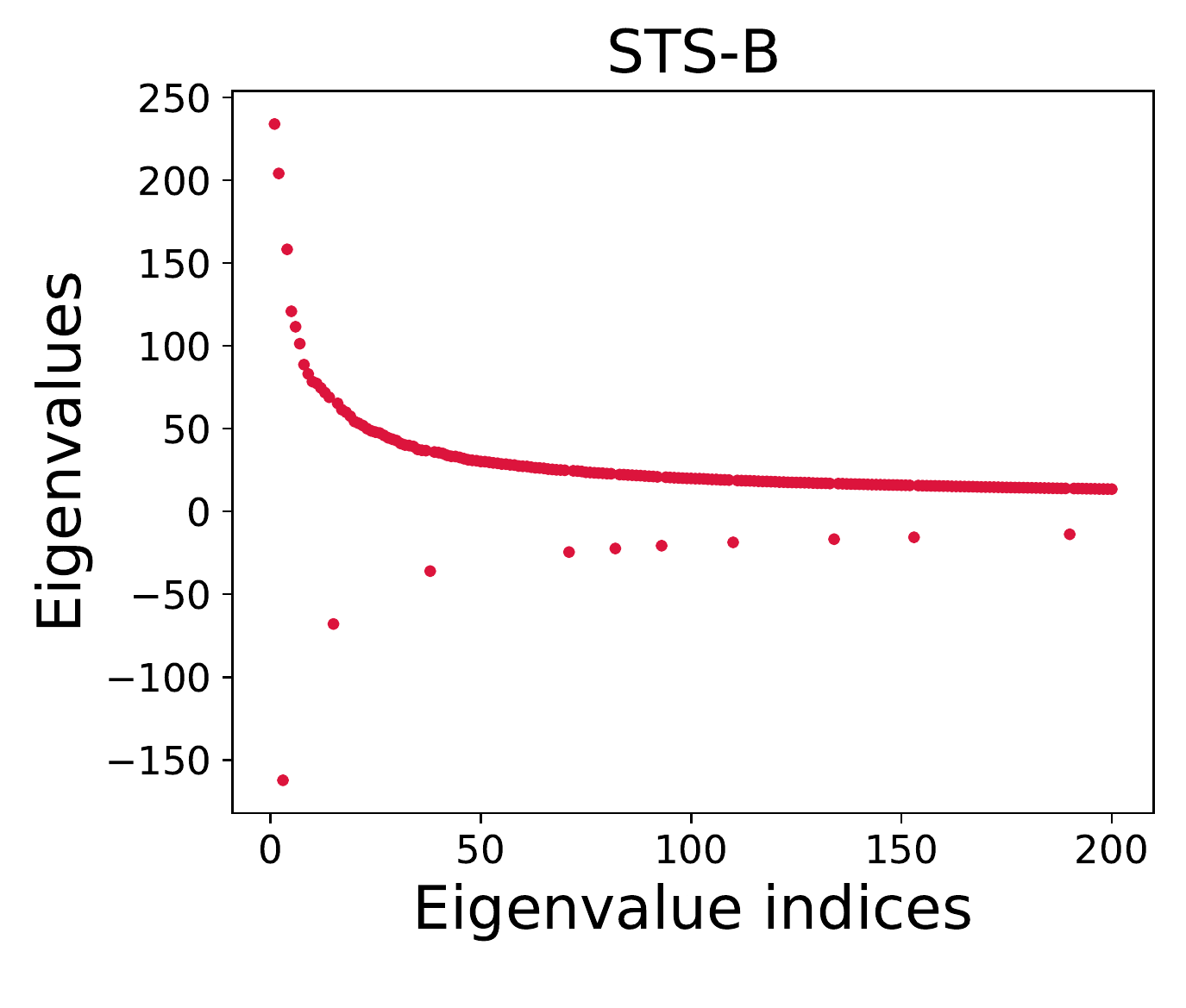}
    \includegraphics[width=0.32\textwidth]{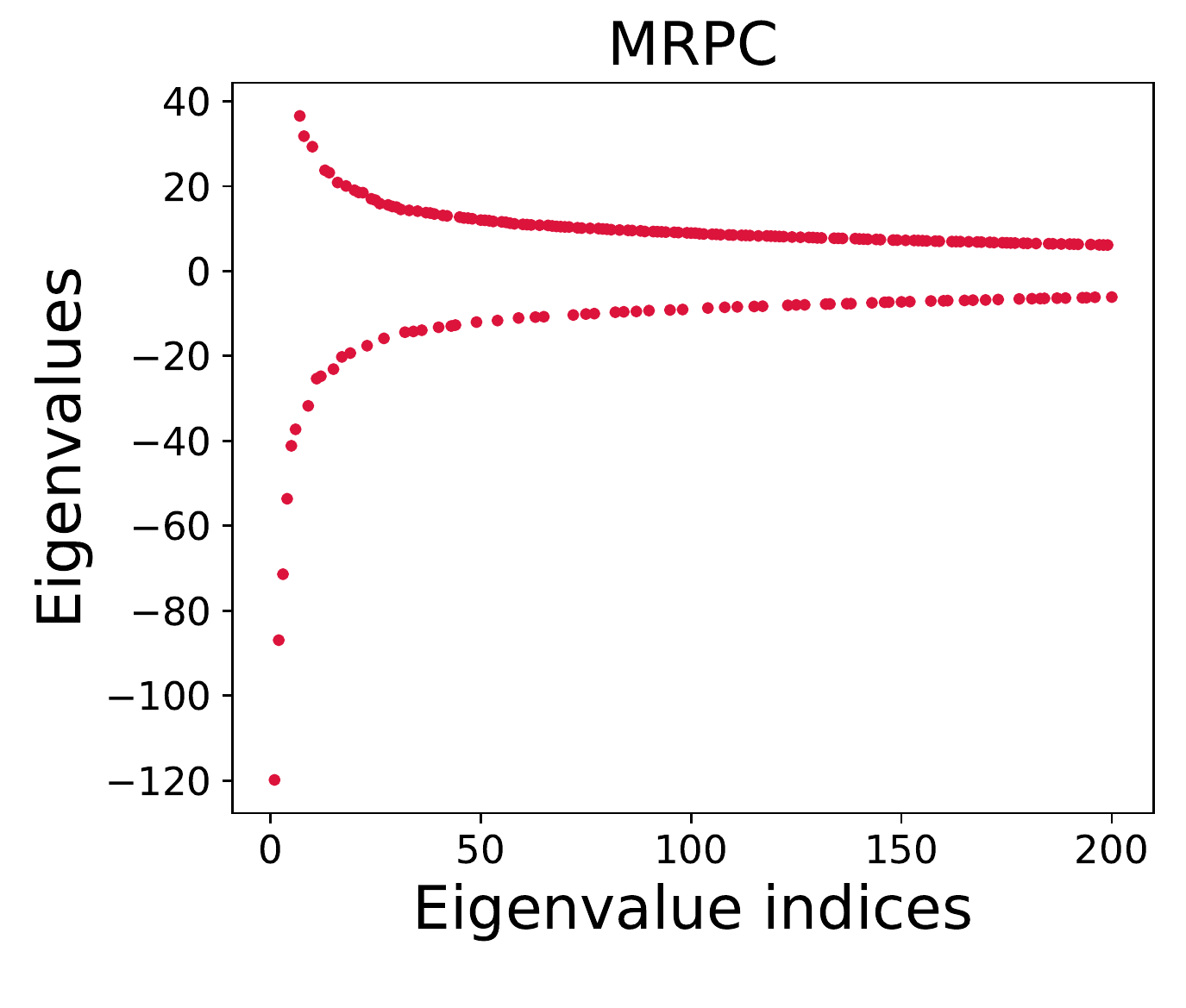}
    \vspace{-3mm}
    \caption{\textbf{Eigenspectrums of language similarity matrices}. The eigenspectrums of many text similarity matrices have relatively few negative eigenvalues -- i.e., they are  relatively close to PSD. Left: similarity matrix arising from the exponentiation of Word Mover's Distance \cite{kusner2015word} -- see Section \ref{sec:wme}. Middle and Right: symmetrized cross-encoder BERT sentence and document similarity matrices \cite{devlin2018bert}. Eigenvalues are plotted in decreasing order of magnitude from rank 2 to 201. The magnitude of the top eigenvalue is typically very large, and so excluded for better visualization.}
    \label{fig: Eigenvalue plots}
    \vspace{-1em}
\end{figure*}

\smallskip 
\noindent\textbf{Obs. 1: Text Similarity Matrices are Often Close to PSD.}
Without some form of structure, we cannot approximate a general $n \times n$ matrix in less than $O(n^2)$ time. Fortunately, while many similarity functions used  in natural language processing do not lead to matrices with PSD structure, they do lead to matrices that are close to PSD, in that they have relatively few negative eigenvalues, and very few negative eigenvalues of large magnitude. See Figure \ref{fig: Eigenvalue plots}.

\smallskip

\noindent\textbf{Obs. 2: Classic \nystrom  Fails on Near-PSD Matrices.}
Given Observation 1, it  is natural to hope that perhaps the \nystrom method is directly useful in approximating many indefinite similarity matrices arising in NLP applications. Unfortunately, this is not the case -- the classic Nystr\"{o}m method becomes very unstable and leads to large approximation errors when applied to indefinite matrices, unless they are very close to PSD.
See Figure \ref{fig: naive nystrom vs CUR variants}. 

\begin{figure*}[h]
    \centering%
    \includegraphics[width=0.32\textwidth]{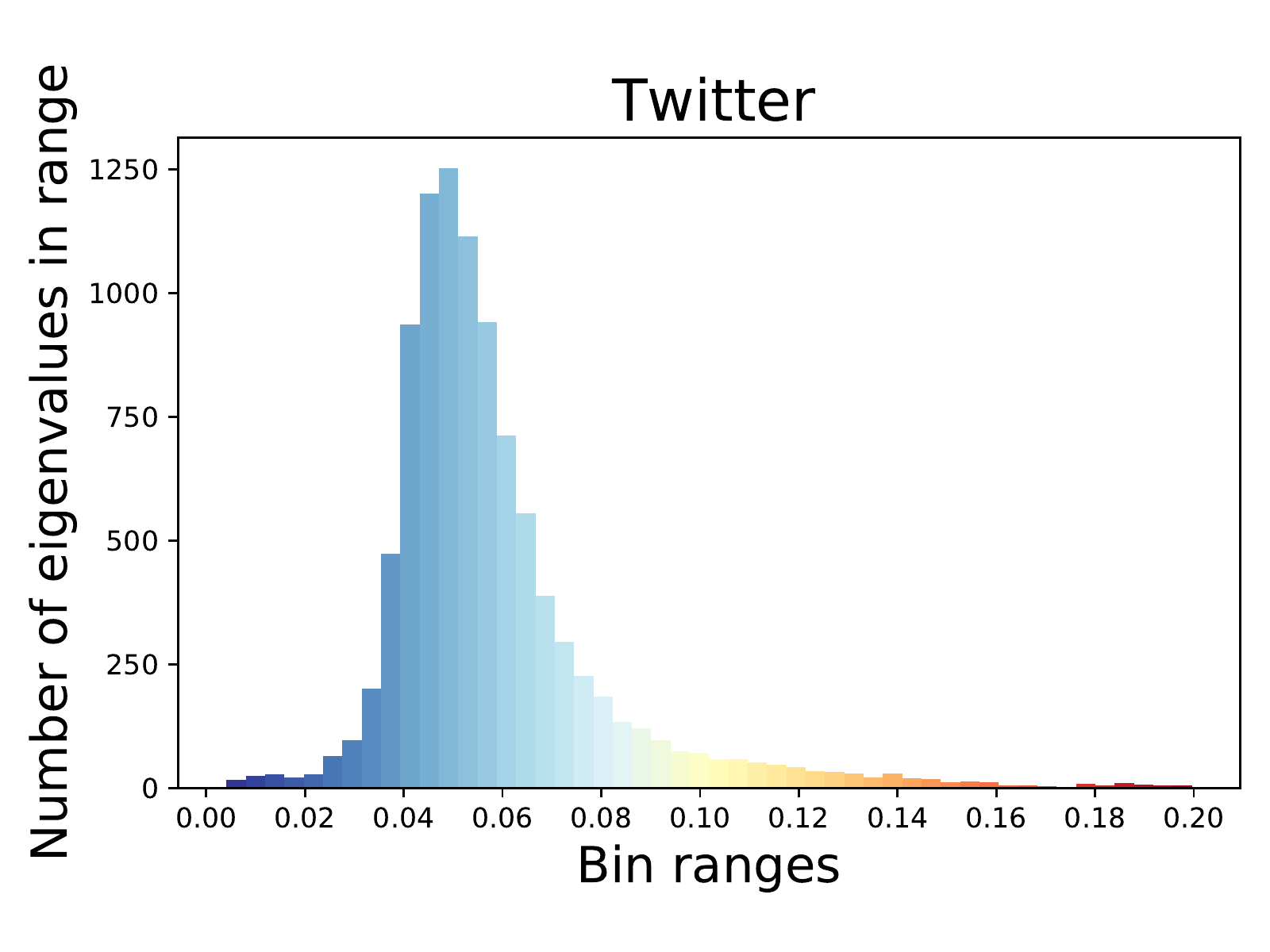}
    \includegraphics[width=0.32\textwidth]{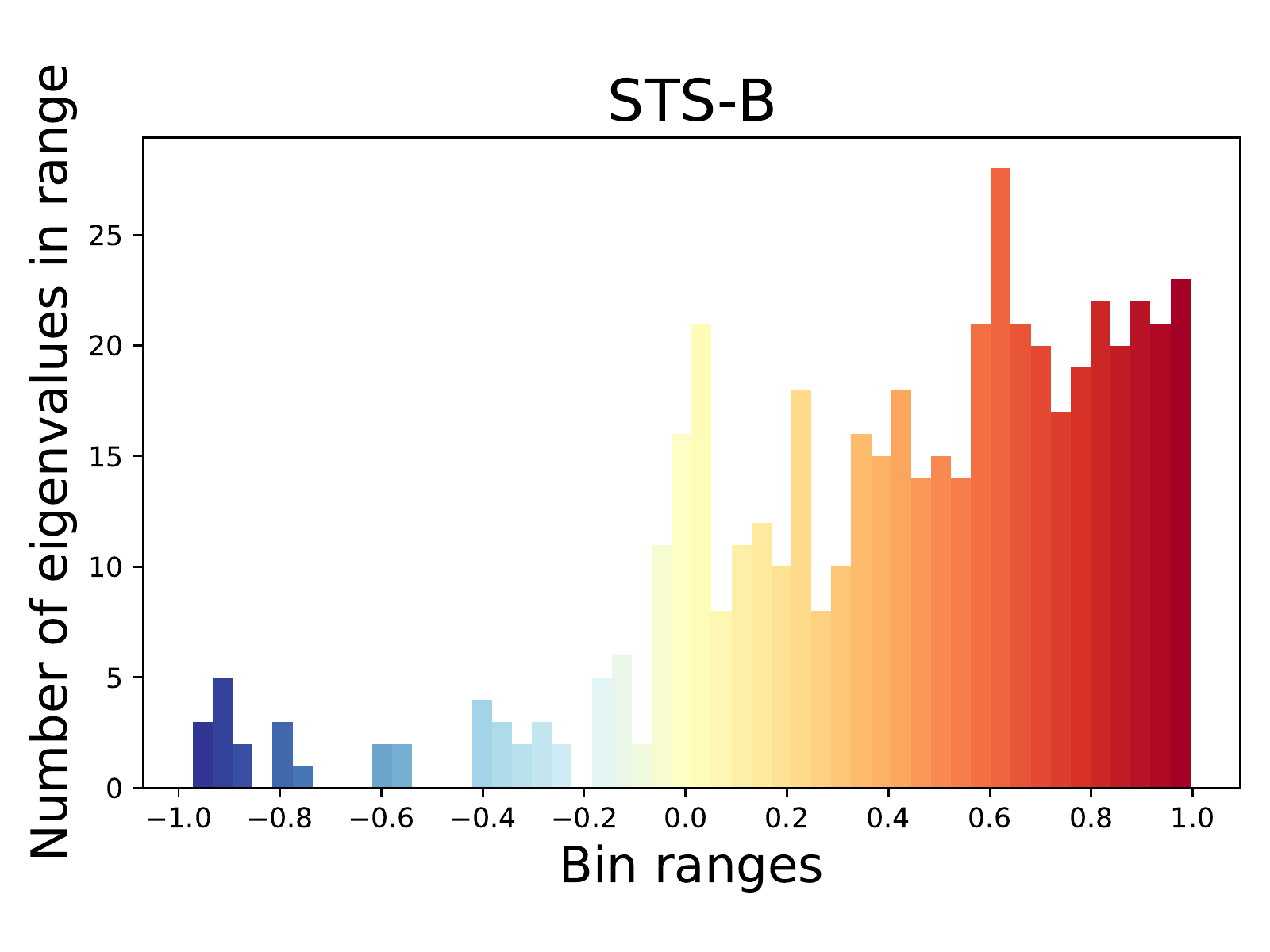}
    \includegraphics[width=0.32\textwidth]{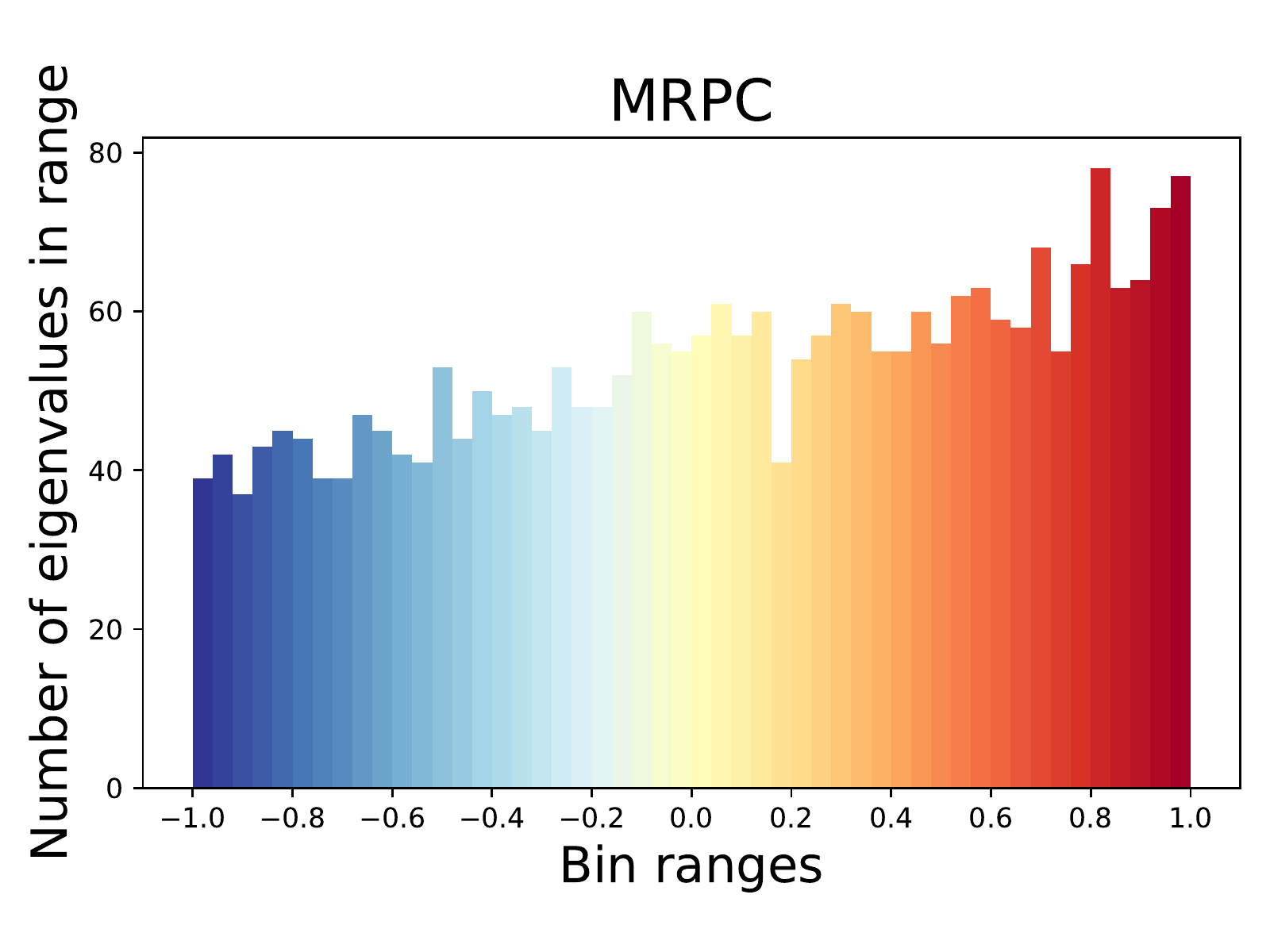}
    \vspace{-3mm}
    \caption{\textbf{Eigenvalue histogram plots}. To understand why \nystrom fails in indefinite matrices, even when they are relatively near-PSD, we independently sample $\bv S^T\bv K \bv S$ with sample size of 200, 50 times. For each sample we compute all eigenvalues, combine, and plot them in a histogram. As we can see, for the STS-B and MRPC matrices, $\bv S^T\bv K \bv S$ often has eigenvalues very close to zero. For Twitter, which is very near-PSD, there are many fewer eigenvalues very close to zero. As we can see in Figure \ref{fig: naive nystrom vs CUR variants}, classic \nystrom performs well on Twitter, but fails on the other two matrices.}
    \label{fig: Eigenvalue histogram plots}
    \vspace{-1em}
\end{figure*}

A major reason for this instability seems to be that $\bv{S}^T \bv{K} \bv{S}$ tends to be ill-conditioned, with several very small eigenvalues that are `blown up' in $(\bv{S}^T \bv{K} \bv{S})^{-1}$ and lead to significant approximation error. See Figure \ref{fig: Eigenvalue histogram plots}. Several error bounds for the classic \nystrom method and the related pseudo-skeleton approximation method (where the sampled sets of rows and columns may be different) applied to indefinite matrices depend on $\lambda_{min}(\bv{S}^T \bv{K} \bv{S})^{-1}$, and thus grow large when $\bv{S}^T \bv{K} \bv{S}$ has eigenvalues near zero \cite{cai2021fast,goreinov1997theory,kishore2017literature}.
When $\bv{K}$ is PSD, by the Cauchy interlacing theorem, $\bv{S}^T \bv{KS}$ is at least as well conditioned as $\bv{K}$. However, this is not the case when $\bv{K}$ is indefinite. When $\bv{K}$ is indefinite, there may exist well-conditioned principal submatrices. Indeed, a number of methods attempt to select $\bv{S}$ such that $\bv{S}^T \bv K \bv S$ is well conditioned \cite{cai2021fast}. However, it is not clear how this can be done in sublinear time in general, without further assumptions.

\subsection{Submatrix-Shifted \nystrom}

Given the above observations, our goal is to give an extension of the \nystrom method that can be  applied to near-PSD matrices. 
Our approach is based on a simple idea: if we let $\lambda_{\min}(\bv{K})$ denote the minimum eigenvalue of $\bv{K}$, then $\bv{\bar K} = \bv{K} - \lambda_{\min}(\bv{K}) \cdot \bv{I}_{n \times n}$ is PSD. $\bv{\bar K}$ can thus be approximated with classic Nystr\"{o}m, and if $|\lambda_{\min}(\bv K)|$ is not too large, this should  yield a good approximation to $\bv{K}$ itself. 

There are two issues with the above approach however: (1) $\lambda_{\min}(\bv{K})$ cannot be computed without fully forming $\bv{K}$ and (2) when $\lambda_{\min}(\bv{K})$ is relatively large in magnitude, the shift can have a significant negative impact on the approximation quality -- this often occurs in practice -- see Figure \ref{fig: Eigenvalue plots}.

We resolve these issues by instead sampling a small principal submatrix of $\bv{K}$, computing its minimum eigenvalue, and using this value to shift $\bv{K}$. Specifically, consider the \nystrom approximation  $\bv{KS}_1 (\bv{S}_1^T \bv{K} \bv{S}_1)^{-1} \bv{KS}_1$ generated by sampling a set of $s_1$ indices $S_1 \subseteq [n]$. We let $S_2$ be a superset of $S_1$, with size $s_2$. We typically simply set $s_2 = 2 \cdot s_1$. We then compute $e = \lambda_{\min}(\bv{S}_2^T \bv{K} \bv{S}_2)$ and apply the \nystrom method to $\bv{\bar K} = \bv{K} - e \cdot \bv{I}_{n \times n}$. 

Since $\bv{S}_2^T \bv{K} \bv{S}_2$ is a principal submatrix of $\bv{K}$, $e = \lambda_{\min}(\bv{S}_2^T \bv{K} \bv{S}_2) \ge \lambda_{\min}(\bv{K})$ and thus $\bv{\bar K}$ will \emph{generally not be PSD}. However, we do have $e  \le \lambda_{\min}(\bv{S}_1^T \bv{K} \bv{S}_1)$, since  $\bv{S}_1^T \bv{K} \bv{S}_1$ is a submatrix of $\bv{S}_2^T \bv{K} \bv{S}_2$. Thus, $\bv{S}_1^T \bv{K} \bv{S}_1 - e \cdot \bv{I}_{n \times n}$ will always be PSD. We also do not expect this matrix  to have any very  small eigenvalues, since we expect a fairly large gap between $\lambda_{\min}(\bv{S}_2^T \bv{K} \bv{S}_2)$ and $\lambda_{\min}(\bv{S}_1^T \bv{K} \bv{S}_1)$ when $s_2$ is significantly larger than $s_1$ -- e.g. $s_2 = 2 \cdot s_1$. 
To further insure this, we can multiply $e$ by a small constant factor $\alpha > 1$ (we typically use $\alpha = 1.5)$ before applying the shift.

Since $(\bv{S}_1^T \bv{K} \bv{S}_1 - e \cdot \bv{I}_{n \times n})^{-1}$ is exactly the joining matrix in the \nystrom approximation of $\bv{\bar K} $, our method resolves the issue of small eigenvalues discussed in Sec. \ref{sec:obv}. As we observe in Sec. \ref{sec: generalized nystrom}, it is enough to recover the strong performance of \nystrom on many near-PSD matrices. Since the minimum eigenvalue of $\bv{S}_2^T \bv{K} \bv{S}_2$ is typically much smaller in magnitude than $\lambda_{\min}(\bv{K})$, we often see improved accuracy  over the exact correction baseline as well.

We call our method Submatrix-shifted \nystrom ({SMS-Nystr\"{o}m}) and give full pseudocode in Algorithm \ref{alg:generalized_nystrom}.
{SMS-\nystrom} requires roughly the same number of similarity computations and running time as classsic \nystrom. We need to perform $(s_2-s_1)^2$ additional similarity computations to form $\bv{S}_2^T \bv{K} \bv{S}_2$ and must also compute  $\lambda_{\min}(\bv{S}_2^T \bv{K} \bv{S}_2)$, which takes $O(s_2^3)$ using a full eigendecomposition. However, this value can also be very efficiently approximated using iterative methods, and typically this additional computation is negligible compared to the full \nystrom running time.

\begin{algorithm}[tbh]
   \caption{Submatrix-Shifted \nystrom ({SMS-Nystr\"om})}
   \label{alg:generalized_nystrom}
\begin{algorithmic}[1]
   \STATE {\bfseries Input:} Data $\{x_i\}_{i=1}^n \in \mathcal{X}$, sample sizes $s_1,s_2$, with $s_2 \ge s_1$ scaling parameter $\alpha$, similarity function $\Delta:\mathcal{X}\times \mathcal{X}\to \mathbb{R}$.
   \STATE Draw at set of $s_2$ indices $S_2$ uniformly at random without replacement from $1,\ldots, n$.
    \STATE Draw at set of $s_1$ indices $S_1$ uniformly at random without replacement from $S_2$. 
   \STATE $\bv {KS}_1 = \Delta(\mathcal{X}, \mathcal{X}_{S_1})$, $\bv{S}_1^T\bv{K}\bv{S}_1 = \Delta(\mathcal{X}_{S_1}, \mathcal{X}_{S_1})$.
      \STATE $\bv{S}_2^T\bv{K}\bv{S}_2 = \Delta(\mathcal{X}_{S_2}, \mathcal{X}_{S_2})$.
      \STATE $e = -\alpha \cdot  \lambda_{\min}(\bv{S}_2^T\bv{K}\bv{S}_2)$.
   \STATE $\bv{KS}_1 = \bv{KS}_1 + e*\bv{I}_{n ,s_1}$, where $\bv{I}_{n \times s_1} \in \R^{n \times s_1}$ has $\bv{I}_{ij} = 1$ if $i = j$, $\bv{I}_{ij} = 0$ otherwise.
   \STATE $\bv{S}_1^T\bv{K}\bv{S}_1 = \bv{S}_1^T\bv{K}\bv{S}_1 + e* \cdot \bv{I}_{s_1 \times s_1}$ .
  
   \STATE \textbf{Return} $\bv{Z} = \bv{KS}_1 (\bv{S}_1^T \bv{K} \bv{S}_1)^{-1/2}$ with $\bv{ZZ}^T \approx \bv{K}$.
   
\end{algorithmic}
\end{algorithm}


\section{Matrix Approximation Results}\label{sec: generalized nystrom}

We now evaluate {SMS-\nystrom}  and several baselines in approximating a representative subset of matrices.

\smallskip

\noindent{\textbf{CUR Decomposition.} In addition to the classic \nystrom method, we consider a closely related family of \emph{CUR decomposition} methods \cite{mahoney2009cur,wang2016towards,pan2019cur}. In CUR decomposition, the matrix $\bv K \in \R^{n \times n}$ is approximated as the product of a small subset of columns $\bv{K}\bv{S}_1 \in \R^{n \times s_1}$, a small subset of rows $\bv{S}_2^T \bv{K} \in \R^{s_2 \times n}$, and a joining matrix $\bv{U} \in \R^{s_1 \times s_2}$. $\bv{KS}_1$ and $\bv{S}_2^T \bv{K}$ are generally sampled randomly -- the strongest theoretical bounds require sampling according to row/column norms or matrix leverage scores \cite{drineas2006fast,drineas2008relative}. However, these sampling probabilities require $\Omega(n^2)$ time to compute and thus we focus on the setting where the subsets of columns and rows are selected uniformly at random.

There are multiple possible options for the joining matrix $\bv{U}$. Most simply and analogously to the \nystrom method, we can set $\bv{U} = (\bv{S}_2^T \bv{K} \bv{S}_1)^+$ -- this is also called \emph{skeleton approximation} \cite{goreinov1997theory}. In fact, if $\bv{S}_1 = \bv{S}_2$, and $\bv{K}$ is symmetric this method is identical to \nystrom.  Alternatively, as suggested e.g., in \cite{drineas2006fast}, we can set $ s_1 = s_2 = s$ and $\bv{U} = \frac{n}{s} \cdot (\bv{KS}_1 \bv{S}_1^T \bv{K})^{-1} \bv{S}_1^T \bv{K} \bv{S}_2$. As we will see, these different choices yield very different performance.

\smallskip
\noindent{\textbf{Results.} 
We report matrix approximation error vs. sample size for several  CUR variants, along with \nystrom and {SMS-\nystrom} on the text similarity matrices from Fig. \ref{fig: Eigenvalue plots}, along with a random PSD matrix. Our results are shown in Fig. \ref{fig: naive nystrom vs CUR variants}. 
\begin{itemize}[topsep=0pt,itemsep=-1ex,partopsep=1ex,parsep=1ex,leftmargin=*]
\item \textbf{Nystr\"{o}m.} As discussed in Sec. \ref{sec:mot}, while \nystrom performs  well on the PSD matrix and the Twitter matrix, which is very near PSD, it completely fails on the other  matrices.
\item \textbf{SMS-Nystr\"{o}m.} Our simple \nystrom variant with $s_2 = 2 \cdot s_1$ and $\alpha = 1.5$ performs well on all test cases, matching the strong performance of \nystrom on the PSD and very near-PSD Twitter matrix, but  still performing well on the less-near PSD cases of STS-B and MRPC.
\item \textbf{Skeleton Approximation.} Similar results  to \nystrom are observed for the closely related skeleton approximation method when $\bv{U} = (\bv{S}_2^T \bv{K} \bv{S}_1)^+$, $s_1 = s_2$, and $\bv{S}_1,\bv{S}_2$ are sampled independently. This is unsurprising -- this method is quite similar to Nystr\"{o}m.
\item \textbf{SiCUR.} If we modify the skeleton approximation, using $s_2 > s_1$, we also obtain strong results. Many theoretical bounds for CUR with joining matrix $\bv{U} = (\bv{S}_2^T \bv{K} \bv{S}_1)^+$ require $s_2 > s_1$ (cf. \cite{drineas2008relative}), and this choice has a significant effect. It is similar  to how SMS-\nystrom  regularizes the inner matrix -- $\bv{S}_2^T \bv{K} \bv{S}_1$ is a rectangular matrix whose minimum singular value is unlikely to be too  small.  We find that setting $s_2 = 2 \cdot s_1$ yields good performance in all cases. To minimize similarity computations, we have $\bv{S}_1$ sample a random subset of  the indices in $\bv{S}_2$. There is very little performance difference if $\bv{S}_1$ and $\bv{S}_2$ are chosen entirely independently. We call this approach  SiCUR for `Simple CUR'.
\item \textbf{StaCUR}: Using the $\bv{U} = \frac{n}{s} \cdot (\bv{KS}_1 \bv{S}_1^T \bv{K})^{-1} \bv{S}_1^T \bv{K} \bv{S}_2$ variant of CUR with $s = s_1 = s_2$ yields what we call StaCUR for `Stable CUR'. StaCUR gives good results on all datasets, however is outperformed by \nystrom on PSD matrices and by SMS-\nystrom and SiCUR in most other cases. Unlike SMS-\nystrom and SiCUR however, StaCUR has  no parameters to tune. Unlike for skeleton approximation, setting $s_2 > s_1$ for this method seems to have little effect so we keep $s_1 = s_2$. In Figure \ref{fig: naive nystrom vs CUR variants} we report results for two variants StaCUR(s) and StaCUR(d), where $\bv{S}_1,\bv{S}_2$ are set equal  or to independent samples respectively. StaCUR(s) typically performs better and requires roughly half as many similarity computations, so we use this variant for the remainder of our evaluations.
\end{itemize}

\begin{figure}[h]
\vspace{-1cm}
    \centering%
    \includegraphics[width=0.48\textwidth]{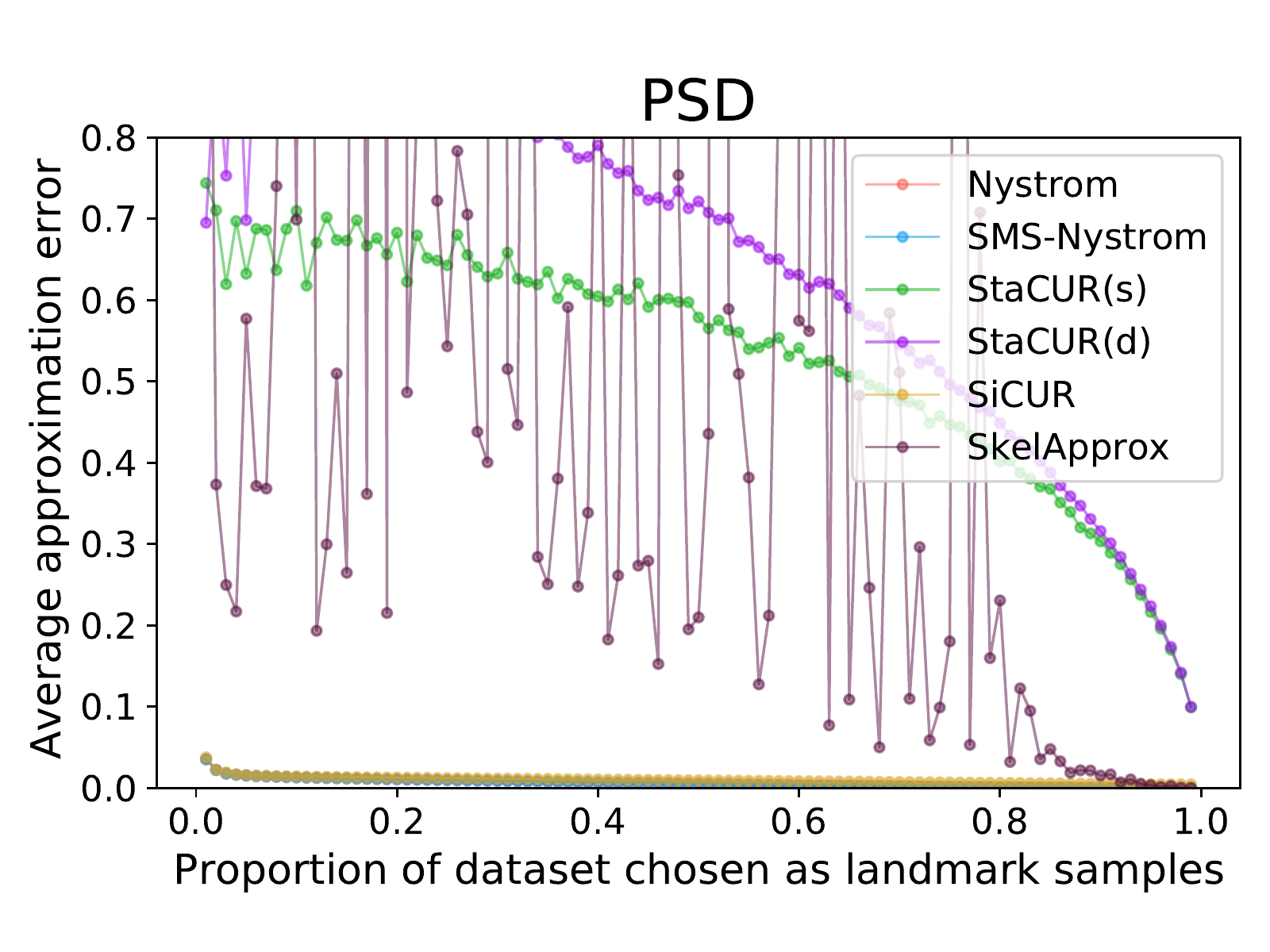}
    \vspace{-.1cm}
    \includegraphics[width=0.48\textwidth]{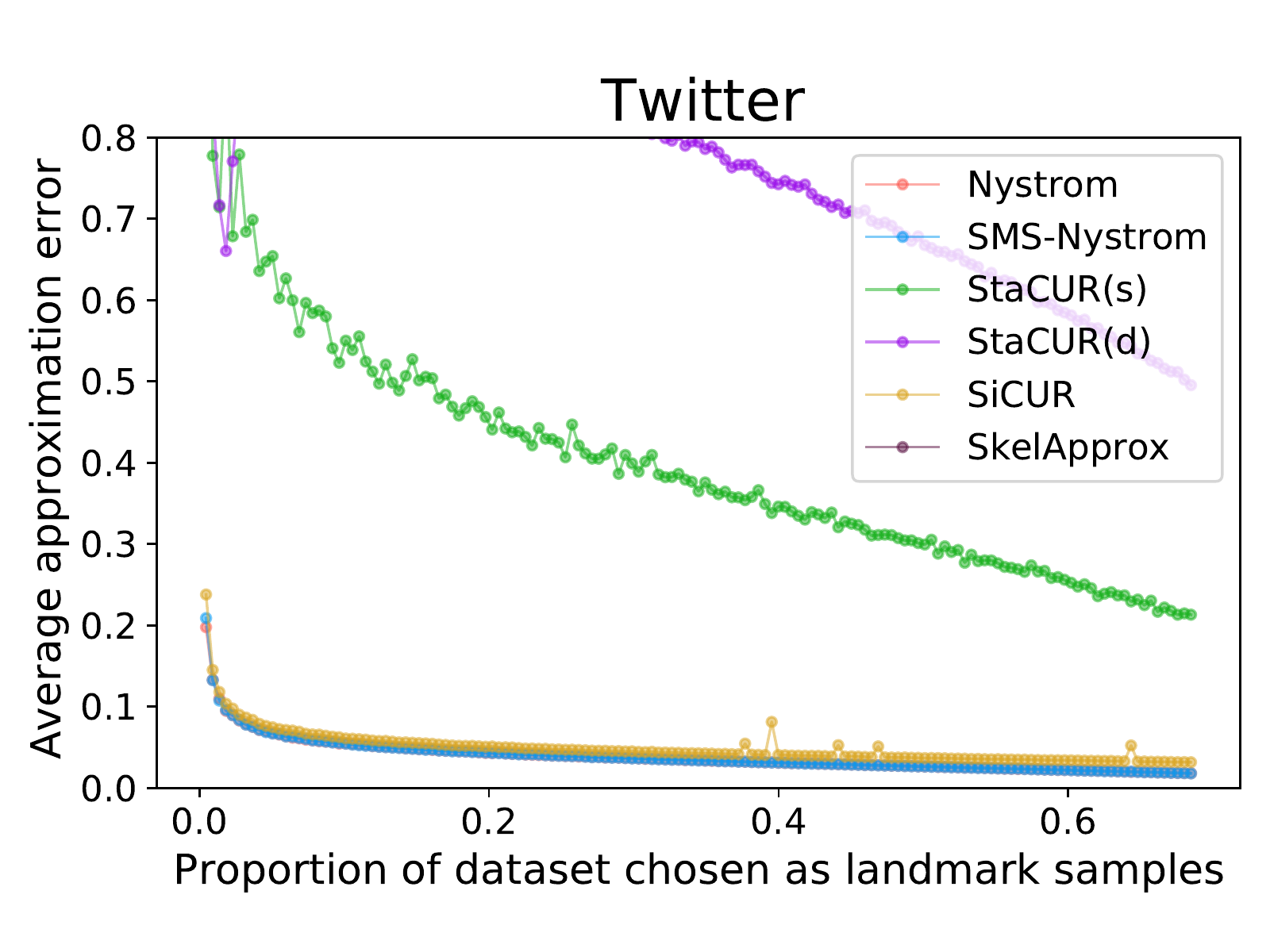}
    \vspace{-.1cm}
    \includegraphics[width=0.48\textwidth]{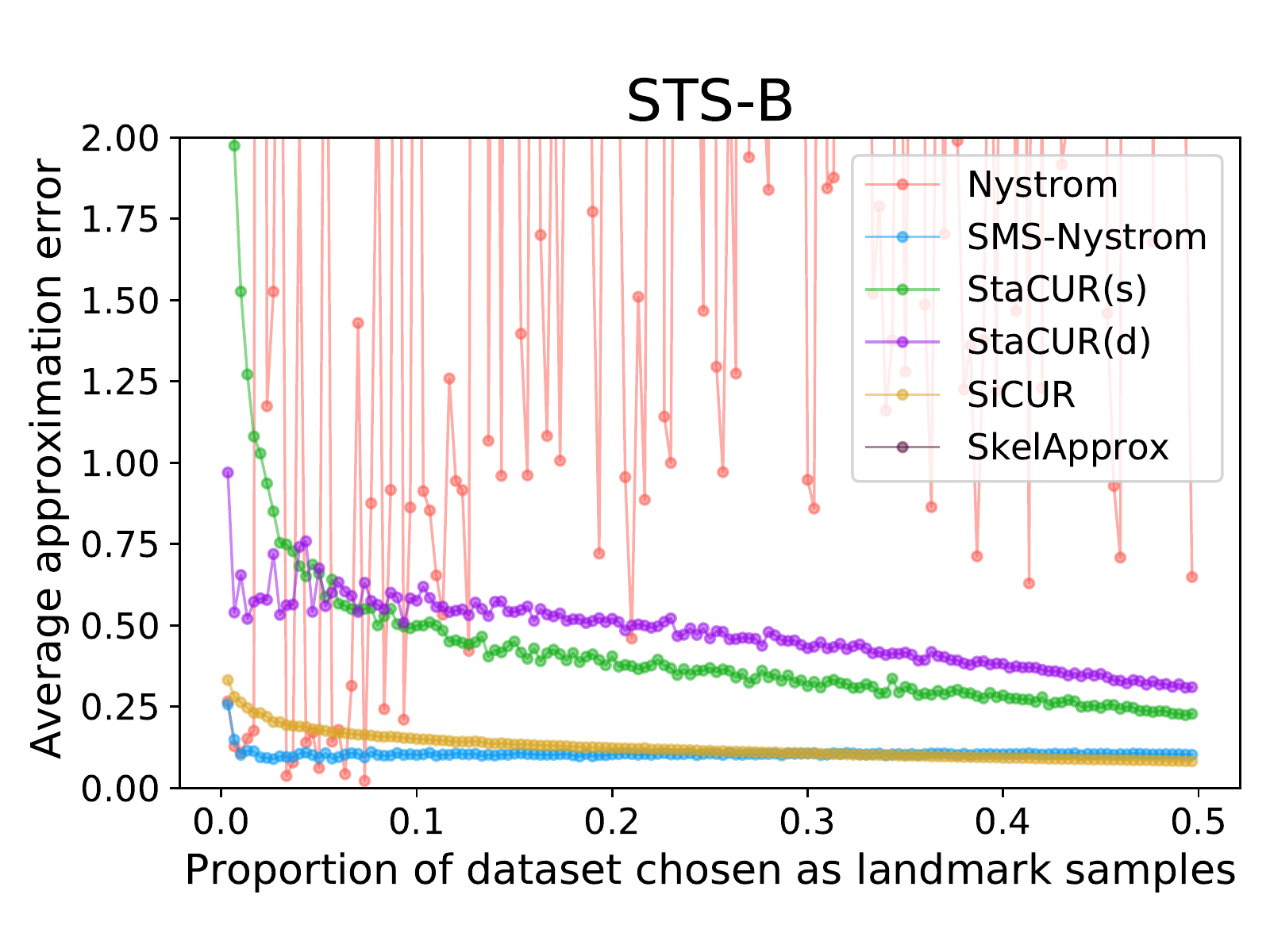}
    \vspace{-.1cm}
    \includegraphics[width=0.48\textwidth]{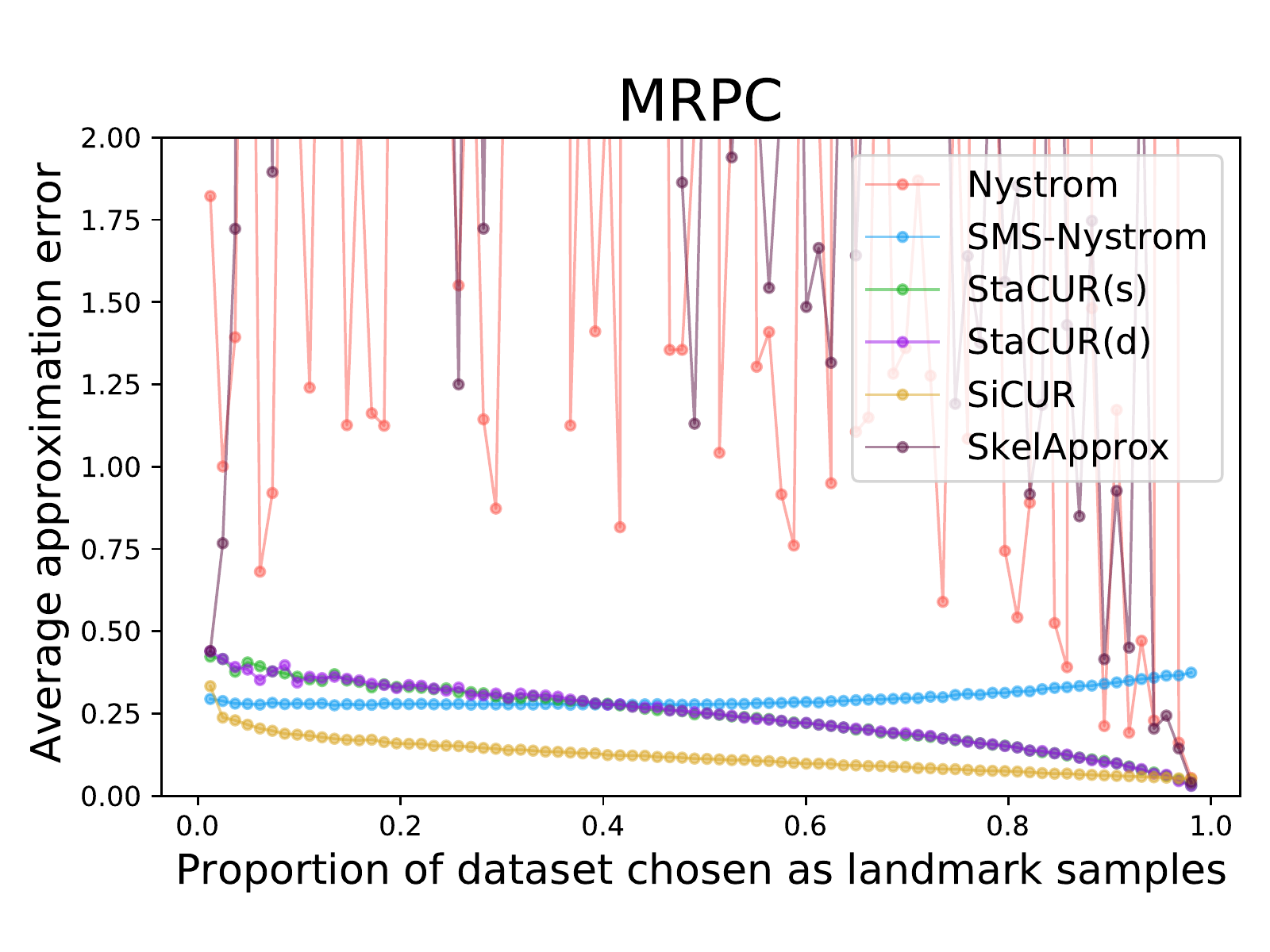}
    \caption{\textbf{Approximation error plots.} Evaluation of sublinear time \nystrom and CUR variants on the language similarity matrices described in Figure~\ref{fig: Eigenvalue plots}, and a test PSD matrix, $\bv{ZZ}^T$ with $\bv{Z} \in \R^{1000 \times 1000}$ having i.i.d. $\mathcal{N}(0,1)$ entries. 
    Error is reported as $\|\bv{K}-\bv{\tilde K}\|_F/\norm{\bv{K}}_F$ and averaged over 10 trials. The $x$-axis is $s/n$. For SiCUR, where $s_2 > s_1$, it is $s_2/n$. If a method does not appear, it may be that it had very large error, which is out of range. The error might increase with samples after a certain limit, we believe this is because the correction term overwhelms the approximation error. }
    \label{fig: naive nystrom vs CUR variants}
\end{figure}

\section{Empirical Evaluation}\label{sec: experiments}
We now evaluate SMS-\nystrom, along with SiCUR and StaCUR on approximating similarity matrices used in document classification, sentence similarity, and cross document coreference, focusing the downstream performance when using the approximated similarity matrix. 
In each application, we show that our approximation techniques can achieve downstream task performance that matches or is competitive with exact methods, using a fraction of the computation.

\subsection{Document Classification with WMD}\label{sec:wme}

Our first application is approximating Word mover's distance (WMD) \cite{kusner2015word}  in document classification. WMD is a variant on the Earthmover's distance, which measures how well words in two documents align, based on how far apart they are in a word embedding space. 
Computing the WMD between two documents with max length $L$ requires $O(L^3 \log(L))$ time  \cite{kusner2015word}, and hence computing a full pairwise distance matrix can be very expensive. 

\smallskip

\noindent\textbf{Word Movers Embedding.} \citet{wu2018word} suggests a PSD similarity function derived from WMD, for which the  similarity matrix $\bv{K}$ can be approximated very efficiently as $\bv{K} \approx \bv{ZZ}^T$ using a random features approximation. 
The resultant feature embeddings $\bv{Z}$ are called Word mover's embeddings (WME). 
Experiments show that WME outperforms true WMD in several classification tasks  \cite{wu2018word}.

\smallskip

\noindent\textbf{Our Approach.} 
Following  \cite{wu2018word}, we define a similarity function between two documents $x,\omega$ by $\Delta(x, \omega) = \exp(-\gamma \text{WMD}(x, \omega))$ for a scalar parameter $\gamma$. While this function does not seem to be PSD, it tends to produce near-PSD matrices -- see. e.g. the Twitter matrix in Figure \ref{fig: Eigenvalue plots}.
We then approximate the similarity matrix $\bv{K}$ using our \nystrom and CUR variants. For \nystrom, we write $\bv{\tilde K} = \bv{ZZ}^T$ and use $\bv{Z}$ as document embeddings (see Algorithm \ref{alg:generalized_nystrom}). For CUR, we  factor  $\bv{U}$ using its SVD $\bv{U}= \bv{W} \bv{S} \bv{V}^T$ as $(\bv{W} \bv{S}^{1/2})(\bv{S}^{1/2}\bv{V}^T)$, and use $\bv{C}\bv{W} \bv{S}^{1/2}$ as document embeddings.

\smallskip

\noindent\textbf{Datasets}. We use 4 different corpora drawn from \cite{kusner2015word, huang2016supervised} for this comparison and the statistics are listed in Table \ref{tab: dataset description WME}. Following \cite{wu2018word, kusner2015word}, for datasets without train/test split, we compute a 70/30 split of the available data. Each word of the documents are represented using Word2Vec word embedding\footnote{code from: \url{}{https://code.google.com/archive/p/word2vec}} \cite{mikolov2013efficient}. The datasets contain documents which are used to perform downstream tasks like sentiment analysis (e.g. Twitter) and topic clustering (e.g. Twitter, Recipe-L and 20News).
\begin{table*}[h]
    \begin{center}
    \begin{small}
    \begin{tabular}{lccccclc}
        \toprule
        \bf Dataset & \bf Classes & \bf Train & \bf Test & \bf BOW Dim & \bf Length & \bf Application\\
        \midrule
        TWITTER & 3 & 2176 & 932 & 6344 & 9.9 & Tweets categorized by sentiment\\
        RECIPE-L & 20 & 27841 & 11933 & 3590 & 18.5 & Recipe procedures labeled by origin\\
        OHSUMED & 10 & 3999 & 5153 & 31789 & 59.2 & medical Abstracts (class subsampled)\\
        20NEWS & 20 & 11293 & 7528 & 29671 & 72 & Canonical User-written posts dataset\\
        \bottomrule
    \end{tabular}
    \end{small}
    \end{center}
        \caption{\textbf{WMD dataset description.} Dataset description for approximating Word mover's distance.}
    \label{tab: dataset description WME}
\end{table*}

\smallskip

\noindent\textbf{Setup.}
We perform 10-fold cross validation on the train split to obtain the best set of hyper-parameters. For WME these hyper-parameters include maximum document length $D_{\max}$, number of iterations for the convergence of random features method $R$, the exponential hyperparameter for the kernel $\gamma$, document embedding scheme and word weighting scheme. We then train SVM using the extracted features in each round. The SVM is implemented using LIBLINEAR~\cite{fan2008liblinear} keeping consistent with \cite{wu2018word}. The hyper-parameters of SVM, the cost of misclassification error and error margin, is also tuned in this process. We then run the final set of hyper-parameters once on the test set to obtain the best score. 

\smallskip

\noindent\textbf{Evaluation.}
We evaluate the performance of our embeddings in multi-class classification  for four different corpora drawn from \cite{huang2016supervised,kusner2015word} -- Twitter (2176 train, 932 test), Recipe-L (27841 train, 11933 test), Ohsumed (3999 train, 5153 test), and 20News (11293 train, 7528 test). For hyperparameter details see Appendix \ref{app:wme}. We evaluate performance over 20  runs of the respective approximation algorithms for the test set, and for each run we compute the average prediction accuracy and  standard deviation.

Following \cite{wu2018word} we compare the performance of the embeddings produced by WME, SMS-Nystr\"{o}m, SiCUR, and StaCUR at several  dimensions (sample sizes $s$). `Small Rank', is the dimension $\le 550$ for which the method achieves highest performance. `Large Rank' is the dimension $\le 4096$ (1500, and 2500 resp. for Twitter and Ohsumed) where the method achieves highest performance. 
See Table \ref{tab: wmd best rank comparison} in Appendix \ref{app:wme} for the exact values of these ranks. For all except WME, the optimal ranks are typically around the dimension limits. This is expected since the methods achieve higher accuracy in similarity approximation with higher samples.

As baselines, we also compare against (1) WMD-kernel, which uses the true similarity matrix with entries given by $\Delta(x, \omega) = \exp(-\gamma \text{WMD}(x, \omega))$ and (2) Optimal -- which uses the optimal rank-$k$ approximation to $\bv{K}$ computed with SVD. This method is inefficient, but can be thought of as giving a cap on the performance of our sublinear time  methods. 

\smallskip

\noindent\textbf{Results.} Our results are reported in Table \ref{tab: wmd accuracy}.  SMS-\nystrom  consistently outperforms all other methods, and even at relatively low-rank nears the `optimal' accuracy. In general, the similarity matrix approximation methods tend to outperform the WME baseline.  
Interestingly, while StaCUR tends to have  lower approximation quality on these similarity matrices (see Figure \ref{fig: naive nystrom vs CUR variants}), its performance in downstream classification is comparable to SMS-Nystr\"om and SiCUR. 

Observe that the approximation methods achieve much higher accuracy than previous work, WME, including an ~8 point improvement on 20News. Our approximation methods achieve results that are within ~2-4 points of accuracy of the expensive WMD-kernel true similarity matrix, while maintaining sublinear time and massive space reduction, (especially on corpora like Recipe-L which has tens of thousands of documents).} We also observe that SMS-Nystrom and SiCUR can achieve high accuracy for small ranks, compared to both WME and WMD-kernel. The amount of computation we save is considerable, e.g., we require just 14\% of the computation for Recipe-L as compared to WMD-kernel. For detailed comparison of rank to performance see Appendixs \ref{app:wme}.

\begin{table}[h]
\setlength{\tabcolsep}{2.5pt}
\footnotesize
    \centering
    \begin{tabular}{m{1em}c@{}cccc}
        \toprule
        \multicolumn{2}{l}{\textbf{Method}} & \textbf{Twitter} & \textbf{RecipeL} & \textbf{Ohsumed} & \textbf{20News}\\
        \midrule
        \rotatebox{90}{Small Rank} & \makecell[l]{WME \\ SMS-N \\ StaCUR \\ SiCUR \\Optimal} & \makecell{ $72.5 \pm 0.5$ \\ $\mathbf{75.3 \pm 1.3}$ \\ $73.8 \pm 1.5$ \\ $74.9\pm1.5$ \\$75.8$} & \makecell{$72.5 \pm 0.4$ \\ $\mathbf{77.7 \pm 1.3}$ \\ $74.9\pm1.0$  \\ $75.9\pm1.5$\\$78.8$} & \makecell{$55.8\pm0.3$ \\ $\mathbf{59.4\pm1.5}$ \\ $58.7\pm2.6$ \\ $59.3\pm1.9$\\$60.3$} & \makecell{$72.9$ \\ $\mathbf{79.3\pm1.3}$ \\ $76.8\pm1.6$ \\ $73.0\pm0.6$\\$82.2$}\\
        \midrule
        \rotatebox{90}{Large Rank}& \makecell[l]{WME \\ SMS-N \\ StaCUR \\ SiCUR \\Optimal } & \makecell{$74.5 \pm 0.5$ \\ $\mathbf{76.1\pm 1.2}$  \\ $71.9\pm2.3$ \\ $75.3\pm2.1$ \\ $76.9$} & \makecell{$79.2 \pm 0.3$ \\ $\mathbf{80.7 \pm 1.1}$ \\ $77.1\pm1.0$ \\ $79.5\pm1.7$\\ $81.3$} & \makecell{$64.5\pm0.2$ \\ $\mathbf{65.3\pm1.1}$ \\ $55.7\pm0.4$ \\ $63.3\pm2.9$\\ $68.2$} & \makecell{$78.3$ \\ $\mathbf{86.6 \pm 1.5}$ \\ $84.2\pm2.1$\\ $85.8\pm1.0$ \\$88.3$}\\
        \midrule
        \multicolumn{2}{l}{WMD-kernel} & \makecell{$78.21$} & \makecell{$82.17$} & \makecell{$69.03$} & \makecell{$89.37$}\\
        \bottomrule
    \end{tabular}
    \caption{Results on document classification task with WMD-based similarity. SMS-\nystrom is abbreviated as SMS-N.}
    \label{tab: wmd accuracy}
\end{table}

\begin{table*}[h]
\footnotesize
    \centering
    \begin{tabular}{m{1em}ccccc}
        \toprule
        & \textbf{Method} & \textbf{STS-B(P)} & \textbf{STS-B(S)} & \textbf{MRPC} & \textbf{RTE}\\
        \midrule
        \rotatebox{90}{SMS-Nys} & \makecell[l]{$@$Rank1 \\ $@$Rank2 \\ $@$Rank3 } & \makecell{ $\mathbf{75.61\pm 1.3}$@$250$ \\ $\mathbf{77.32\pm 1.8}$@$350$ \\ $\mathbf{79.36\pm1.5}$@$700$} & \makecell{$\mathbf{75.27\pm 1.5}$@$250$ \\ $\mathbf{76.91\pm1.8}$@$350$ \\ $\mathbf{78.56\pm1.3}$@$700$} & \makecell{$57.37\pm2.2$@$100$ \\ $63.93\pm2.7$@$250$ \\ $63.04\pm1.1$@$500$} & \makecell{$60.01\pm1.1$@$100$\\ $61.84\pm2.1$@$250$ \\ $60.23\pm1.1$@$450$}\\
        \midrule
        \rotatebox{90}{StaCUR} & \makecell[l]{$@$Rank1 \\ $@$Rank2 \\ $@$Rank3 } & \makecell{ $28.21\pm 2.3$@$250$ \\ $34.18\pm1.6$@$350$ \\ $45.87\pm1.1$@$700$} & \makecell{$46.77 \pm 2.1$@$250$ \\ $49.86\pm3.2$@$350$ \\ $51.73\pm1.4$@$700$} & \makecell{$53.78\pm 4.2$@$100$ \\ $64.41\pm0.5$@$250$ \\ $66.97\pm1.1$@$500$} & \makecell{$58.23\pm2.2$@$100$ \\$57.32\pm1.2$@$250$ \\ $61.37\pm0.1$@$450$}\\
        \midrule
        \rotatebox{90}{SiCUR} & \makecell[l]{$@$Rank1 \\ $@$Rank2 \\ $@$Rank3 } & \makecell{ $45.60\pm 3.1$@$250$ \\ $57.65\pm2.6$@$350$ \\ $68.84\pm0.2$@$700$} & \makecell{$44.91 \pm 2.8$@$250$ \\ $56.52\pm2.4$@$350$ \\ $68.97\pm0.4$@$700$} & \makecell{$\mathbf{69.42\pm 3.7}$@$100$ \\ $\mathbf{72.38\pm2.1}$@$250$ \\ $\mathbf{75.53\pm0.9}$@$500$} & \makecell{$\mathbf{61.11\pm2.2}$@$100$ \\ $\mathbf{62.67\pm1.5}$@$250$ \\ $\mathbf{63.28\pm0.3}$@$450$}\\
        \midrule
        & \makecell[l]{BERT \\ SYM-BERT} & \makecell{$85.09$ \\ $85.54$} & \makecell{$84.70$ \\ $85.13$} & \makecell{$83.30$ \\ $83.75$} & \makecell{$65.98$ \\ $66.10$}\\
        \bottomrule
    \end{tabular}
   \caption{Performance comparison  of original BERT similarities and approximated similarities on GLUE benchmarks. Ranks (i.e., sample size) are recorded next to each result. 
   }
    \label{tab: correlation comparison GLUE}
    \vspace{-1.5em}
\end{table*}

\begin{figure} 
    \centering
    \includegraphics[width=0.48\textwidth]{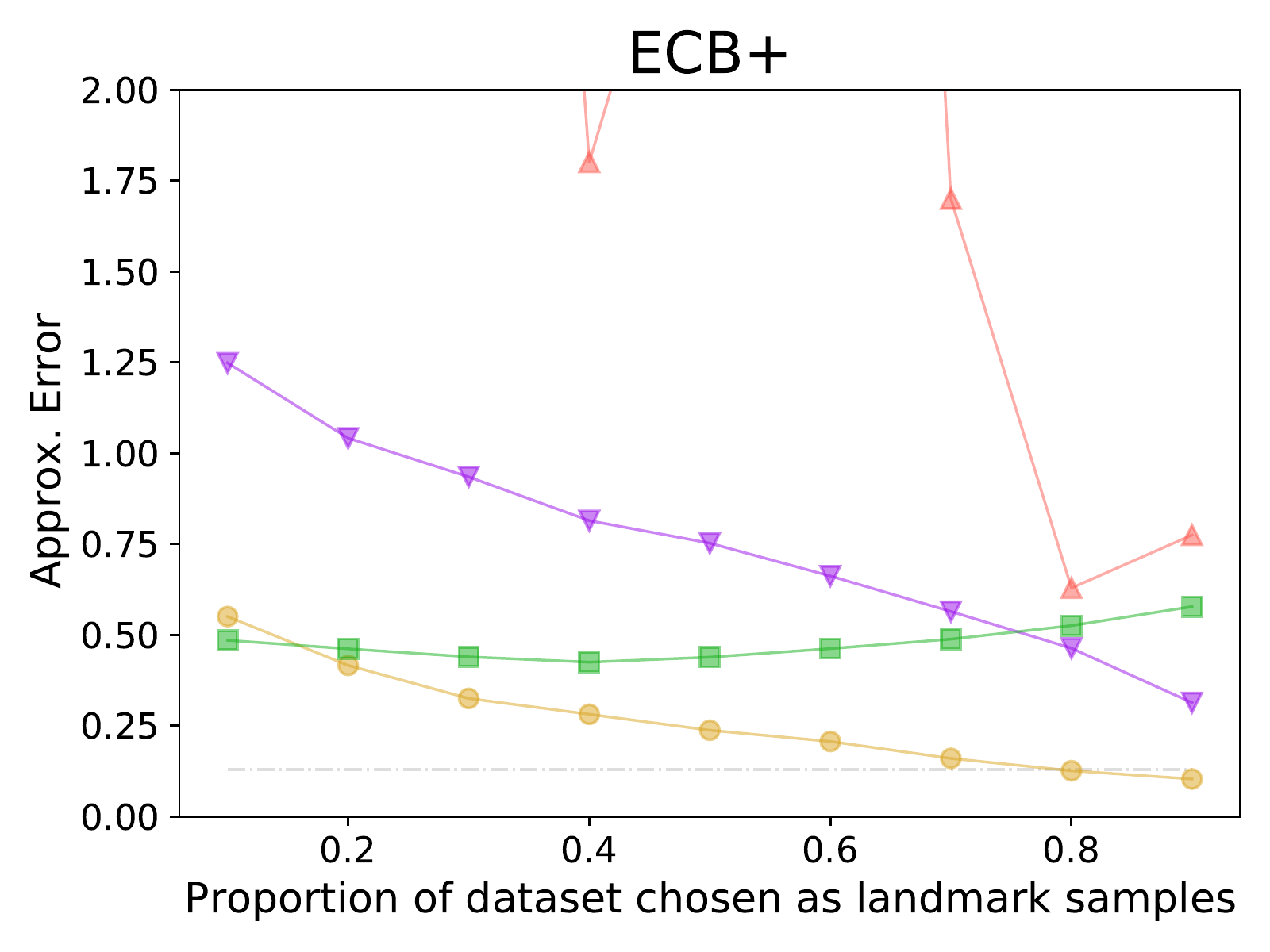}
    \includegraphics[width=0.48\textwidth]{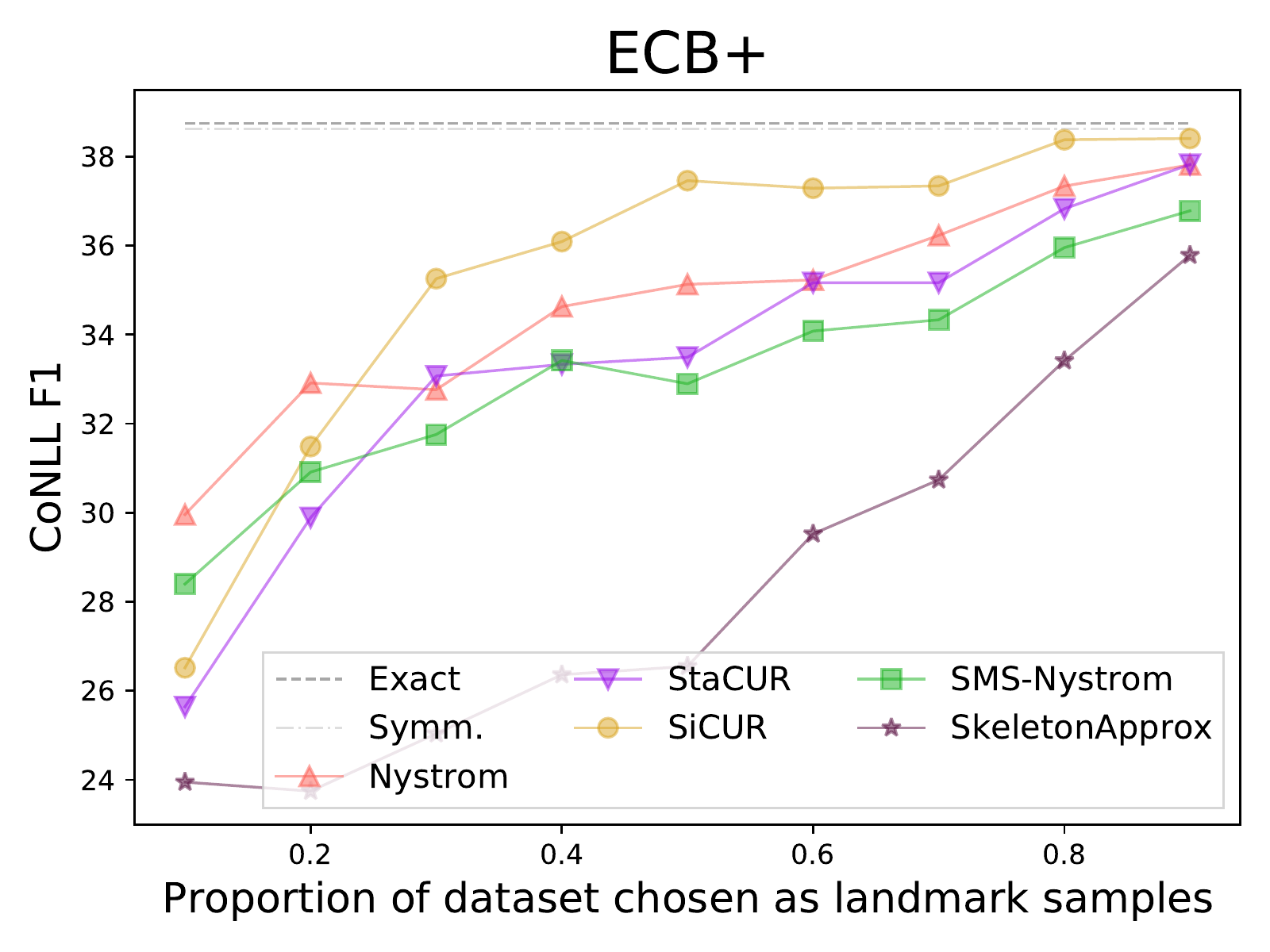}
    \caption{\textbf{Cross-document Entity \& Event Coreference Performance}. We report the downstream task F1 performance and approximation error on EventCorefBank (ECB+).}
    \label{fig:coreference_results}
\end{figure}

\subsection{Approximation of Cross-Encoder BERT Similarity Matrices}
\label{sec: cross encoder similarity details}

Our second application is to approximate similarity given by a cross-encoder BERT model \cite{devlin2018bert}. 
\smallskip

\noindent\textbf{Datasets.} We compare this task on multiple GLUE benchmark datasets. These include STS-B, MRPC and RTE. For each of the datasets we have a pair of sentences and their corresponding score. The scores are based on human judgements and the performance of the algorithm is compared by computing the correlation of human scores to model prediction. While STS-B compares similarity of two sentences by assigning scores from 0 to 5 (0 being not similar and 5 being most similar), the other datasets require binary labels only. STS-B is a sentence similarity task with 1469 pairs of sentences in the validation set. MRPC is a corpus of sentence pairs automatically extracted from online news sources, with human annotations for whether the sentences in the pair are semantically equivalent with 409 sentence pairs in validation set. RTE is a sentence entailment task checking if second sentence entails the first one and contains 278 sentence pairs in the validation set. A summary of the dataset is available in Table \ref{tab: cross-encoder datasets}. We compute Pearson (P) and Spearman-rank (S) correlation coefficients for the STS-B dataset and compute accuracy of prediction for MRPC and RTE datasets.

\begin{table*}[ht]
    \begin{center}
    \begin{small}
    \begin{tabular}{lccclc}
        \toprule
        \bf Dataset & \bf Score range & \bf Train & \bf Test & \bf Application\\
        \midrule
        STS-B & 1-5 & 5749 & 3000 & Semantic similarity of sentence pairs based on human annotations\\
        MRPC & 0-1 & 3668 & 816 & Semantic equivalence of sentence pairs\\
        RTE & 0-1 & 2490 & 554 & Text entailment of news and Wikipedia articles\\
        \bottomrule
    \end{tabular}
    \end{small}
    \end{center}
    \vspace{-1em}
        \caption{\textbf{GLUE dataset description.} Dataset description for comparison on GLUE benchmarks.}
    \label{tab: cross-encoder datasets}
\end{table*}

\smallskip

\noindent\textbf{Setup.} For each dataset we finetune the weights of BERT \cite{devlin2018bert} using the train set. Since the dataset generally contains only a few test pairs, we first grab all sentences from the test set and create all pair combinations. This gives us a $n\times n$ size test set. We then evaluate the test set on the trained transformer. Once this is evaluated it is easy to check the performance since the true labelled pairs are already available among the newly created $n \times n$ test set. Before approximating the resultant matrix we also symmetrize the matrix. The symmetrization acts as a regularizer for the datasets we have experimented with, and perhaps this is why in almost all cases the downstream performance of the symmetrized similarity matrix is better than the true similarity matrix (See Table \ref{tab: correlation comparison GLUE}). The symmetrized matrix is then approximated using any of the approximation algorithms. The performance on downstream task is measured by comparing the similarities of the approximated scores for pairs for which we have true label available.

\smallskip

\noindent\textbf{Evaluation.} We consider three GLUE benchmark datasets -- STS-B, where the goal is to detect sentence similarity, MRPC, where the goal is to detect semantic equivalence, and RTE, where the goal is to detect entailment. See Table \ref{tab: cross-encoder datasets} for further details. For each task, we first train the BERT model on the test set, using code from \cite{wolf2019huggingface}. We then compute the full BERT  similarity matrix for all sentences in the validation set, which consists of a set of sentence pairs, each with a `true' score, derived from human judgements. The similarity matrices for the datasets STS-B, MRPC and RTE are $ 3000\times 3000$, $816\times 816$, and $554\times 554$ respectively, and thus are very expensive to fully compute, motivating the use of our fast approximation methods. We compute  approximations to this full similarity matrix using SMS-Nystr\"{o}m, SiCUR, and StaCUR. In general, the BERT similarity matrices are non-PSD (see Figure \ref{fig: Eigenvalue plots}), and in fact non-symmetric. So that SMS-\nystrom can be applied, we symmetrize them as $\bar \Delta(x,\omega) = 1/2 \cdot ( \Delta(x,\omega) +\Delta(\omega,x))$. 

We use the approximate similarity matrix to make predictions on a dataset of labeled sentences for evaluation.
Performance is measured via Pearson and Spearman correlation with the human scores for STS-B, F1 score of predicted labels for MRPC, and accuracy for RTE.
We report the average scores obtained with different sample sizes,  over 50  runs. 

\smallskip

\noindent\textbf{Results.} Table~\ref{tab: correlation comparison GLUE} reports results for the approximations, the exact, and the symmetrized (SYM-BERT) approaches. 
SMS-\nystrom performs particularly well on STS-B, while SiCUR performs best on MRPC. All methods are comparable on RTE. This performance is inline with the accuracy in approximating $\bv{K}$. Comparing the relative Frobenius norm error with respective to the predicted outputs of BERT we observe similar trends we have seen in Figure \ref{fig: naive nystrom vs CUR variants}. These results are reported in Table \ref{tab: error comparison CEBERT}. For larger sample sizes the SMS-\nystrom error tends to shoot up. We suspect this is mostly because the Frobenius norm error resulting from translating $\bv S^T\bv{KS}$ with estimated eigen-correction increases as we tend to the full rank of the respective matrices. 

\begin{table}[!ht]
\setlength{\tabcolsep}{3pt}
    \centering
    \begin{tabular}{m{1em}cccc}
        \toprule
        & \textbf{Method} & \textbf{STS-B} & \textbf{MRPC} & \textbf{RTE}\\
        \midrule
        \rotatebox{90}{SMS-Nys} & \makecell[l]{$@$Rank1 \\ $@$Rank2 \\ $@$Rank3 } & \makecell{ $0.1738$@$250$ \\ $0.1402$@$350$ \\ $0.1349$@$700$} & \makecell{$0.3571$@$100$ \\ $0.3138$@$250$ \\ $0.3364$@$500$} & \makecell{$0.1186$@$100$ \\ $0.1070$@$250$ \\ $0.1289$@$450$}\\
        \midrule
        \rotatebox{90}{StaCUR} & \makecell[l]{$@$Rank1 \\ $@$Rank2 \\ $@$Rank3 } & \makecell{ $0.5353$@$250$ \\ $0.5001$@$350$ \\ $0.4511$@$700$} & \makecell{$0.3941$@$100$ \\ $0.3339$@$250$ \\ $0.2530$@$500$} & \makecell{$0.3076$@$100$ \\ $0.3398$@$250$ \\ $0.1144$@$450$}\\
        \midrule
        \rotatebox{90}{SiCUR} & \makecell[l]{$@$Rank1 \\ $@$Rank2 \\ $@$Rank3 } & \makecell{ $0.2833$@$250$ \\ $0.2264$@$350$ \\ $0.1916$@$700$} & \makecell{$0.2149$@$100$ \\ $0.1854$@$250$ \\ $0.1587$@$500$} & \makecell{$0.0658$@$100$ \\ $0.0691$@$250$ \\ $0.0503$@$450$}\\
        \midrule
        & \makecell[l]{BERT \\ SYM-BERT} & \makecell{$0.0$ \\ $0.1375$} & \makecell{$0.0$ \\ $0.1958$} & \makecell{$0.0$ \\ $0.0187$}\\
        \bottomrule
    \end{tabular}
    \caption{\textbf{Error comparisons with cross-encoder BERT}. Comparing relative Frobenius norm error for approximations of BERT outputs.}
    \label{tab: error comparison CEBERT}
\end{table}

\smallskip 

\subsection{Approximate Similarity Matrices for Entity \& Event Coreference}
\label{sec: coref details}

Cross-document entity and event coreference is a clustering problem. Ambiguous
mentions of entities and events that appear throughout a corpus of documents
are to be clustered into groups such that each group refers to the same
real world entity or event. \citet{cattan2020streamlining} present an approach that (1) learns a pairwise similarity function between ambiguous 
mentions and (2) uses average-linkage agglomerative clustering with a similarity threshold to produce the predicted clustering. The pairwise 
similarity function is a MLP which takes as input the concatenation of RoBERTa \cite{liu2019roberta}, embeddings of two mentions and their elementwise product. This induces a matrix that is asymmetric and not-PSD.  We symmetrize the matrix for the approximations.

\smallskip

\noindent\textbf{Task Description.} Entities and events are mentioned ambiguously in natural language. Cross document coreference is the task of clustering 
mentions, such that the mentions in 
each cluster correspond to the same 
real world entity or event. 
For example in Figure \ref{fig:crossdoc}, we show an example of data from ECB+ corpus \cite{cybulska2014using}, observe that the entity \emph{Doug McDermott} is mentioned in each of the three documents with a variety of name spellings (just his last name \emph{McDermott}, the pronoun \emph{him}, and as \emph{all time leading scorer}. We consider the `end-to-end' setting of this task, in which we need to select the tokens in each document that correspond to a mention of the entities as well as performing the clustering of those entity mentions. 
\begin{figure}
    \centering
    \includegraphics[width=0.475\textwidth]{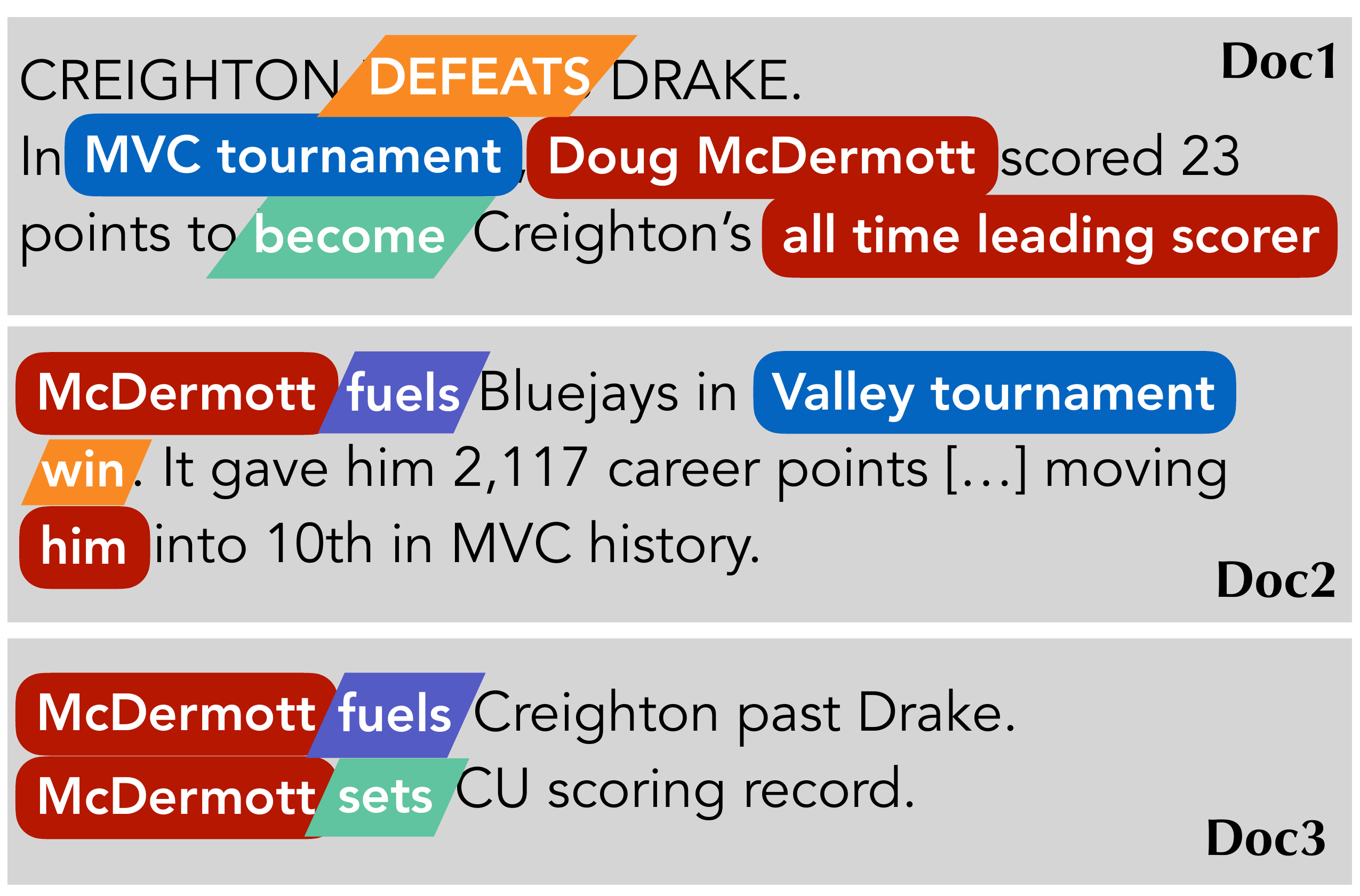}
    \caption{\textbf{Example Cross-Document Coreference} Colors Represent ground truth coreference decisions. Entities are shown as rounded boxes and events as parallelograms.}
    \label{fig:crossdoc}
    \vspace{-1em}
\end{figure}

\smallskip

\noindent\textbf{Task Formalization.} Given a corpus of documents $\mathcal{D}$, each document consists word tokens $w_1,\dots,w_T$. Performing cross-document coreference requires predicting a set of mentions $\mathcal{M}$ where each $m \in \mathcal{M}$ is a sequence of tokens $w_i,w_{i+1},\dots,w_{i+k}$. There is a ground truth set of mentions $\mathcal{M}^\star$. We evaluate the performance of the predicted $\mathcal{M}$ compared to $\mathcal{M}^\star$ according to the MUC, $B^3$, CEAF metrics and their average (als known as CoNLL) \cite{pradhan-EtAl:2014:P14-2}. For the ECB+ data, the documents are organized into a set of topics. The dataset assumes that each entity appears in only one topic. We evaluate in a setting where the topic of each document are assumed to be known.

\smallskip

\noindent\textbf{Similarity Function.} We use the state-of-the-art model described in \cite{cattan2020streamlining}. We train the model using the code provided by the authors\footnote{\url{https://github.com/ariecattan/coref}}. The model encodes tokens with RoBERTa \cite{liu2019roberta} and produces a similarity between two mentions with an MLP applied to the vector that is the concatenation of the two mentions and the element-wise product of the two. 

\smallskip

\noindent\textbf{Evaluation.} We evaluate both the approximation error
as well as the downstream coreference task performance (CoNLL F1 \cite{pradhan-EtAl:2014:P14-2})
of approximating similarity matrix of the model. 
We evaluate on the EventCorefBank+ Corpus \cite{cybulska2014using}. 

\smallskip

\noindent\textbf{Results.} Figure~\ref{fig:coreference_results} shows the downstream 
task performance measured in CoNLL F1 and the approximation
error as a function of the number of landmarks used. 
We find a similar trend as the previous two tasks. 
SiCUR performs very well in terms of both metrics,
with performance improving as more landmarks are added, achieving 
nearly the same F1 (within 1 point) performance when 90\% of the data is used for landmarks and very competitive performance (within 1.5 points) with just 50\%, a drastic reduction in time/space compared to the exact matrix. 
SMS-\nystrom required additional rescaling 
for this task likely due to sensitivity of threshold of 
agglomerative clustering. We report the rescaled version, which is quite competitive with StaCUR (see the following paragraph for more detail). The results indicate that the proposed approximation could help scale models for which the $\Omega(n^2)$ similarity computations would be intractable. 

\smallskip 

\noindent\textbf{Rescaling.} We observed that although SMS-\nystrom could approximate the similarity matrices well enough the downstream performance of the approximated matrices was not good for the datasets in our experiments. So our primary hypothesis was that because of the shift $e* \bv I_{n,s_1}$, the scores are not getting scaled properly. A natural variation to try in such scenarios would be to rescale them back to original proportions. As such we implemented a modification to SMS-\nystrom where Step 8 of Algorithm \ref{alg:generalized_nystrom} is replaced with $\bv{S}_1^T\bv{K}\bv{S}_1 = \beta(\bv{S}_1^T\bv{K}\bv{S}_1 + e* \cdot \bv{I}_{s_1 \times s_1})$, where $\beta = \|\bv{S}_1^T\bv{K}\bv{S}_1\|_2 / \|\bv{S}_1^T\bv{K}\bv{S}_1 + e* \cdot \bv{I}_{s_1 \times s_1}\|_2$. Thus $\beta$ is just a rescaling of the shifted $\bv{S}_1^T\bv{K}\bv{S}_1$ matrix. We compare the rescaled and non-rescaled versions in Figure~\ref {fig:coref_rescaled}.

\begin{figure*}[h]
    \centering
    \includegraphics[width=0.48\textwidth]{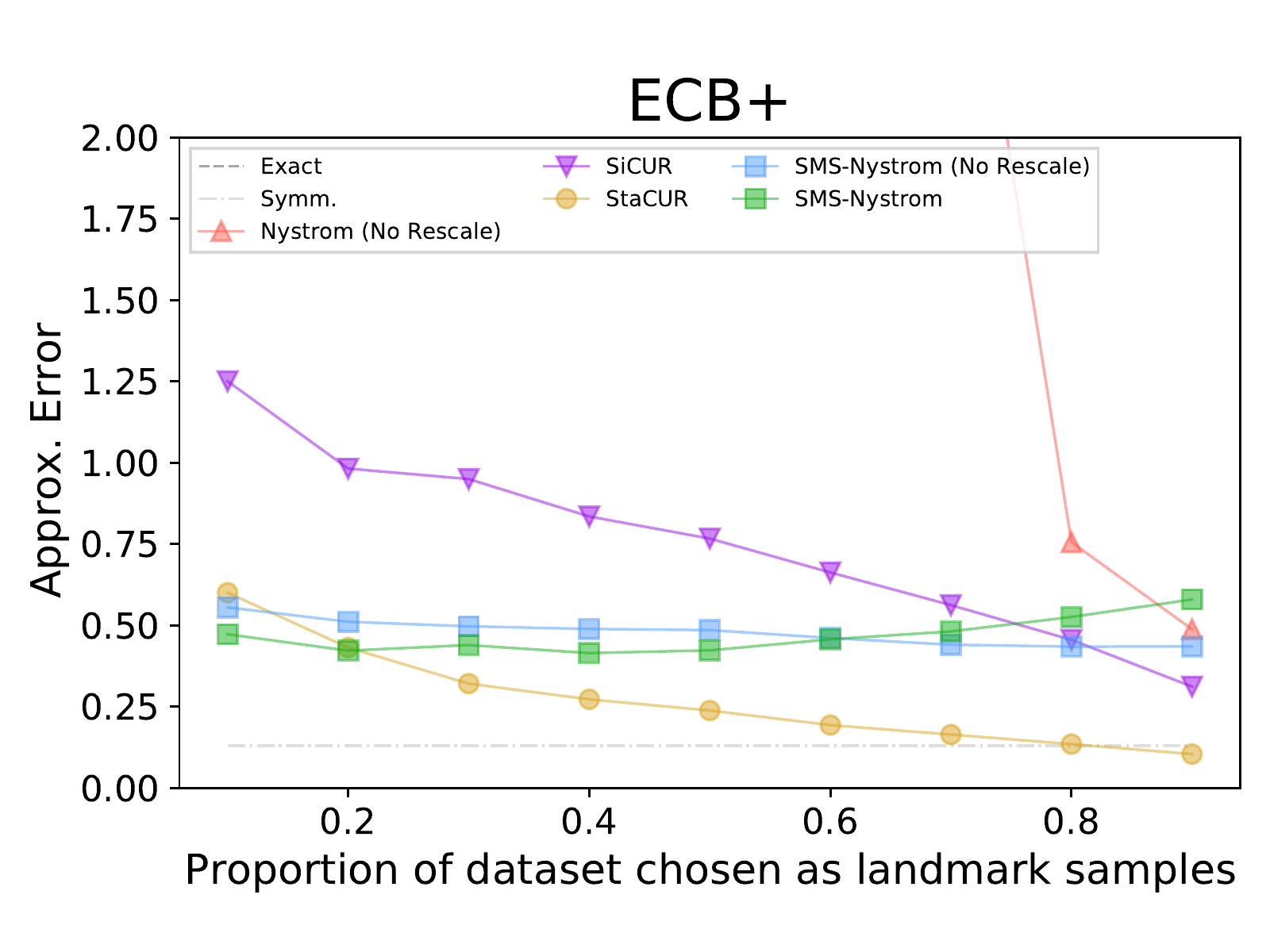}
    \includegraphics[width=0.48\textwidth]{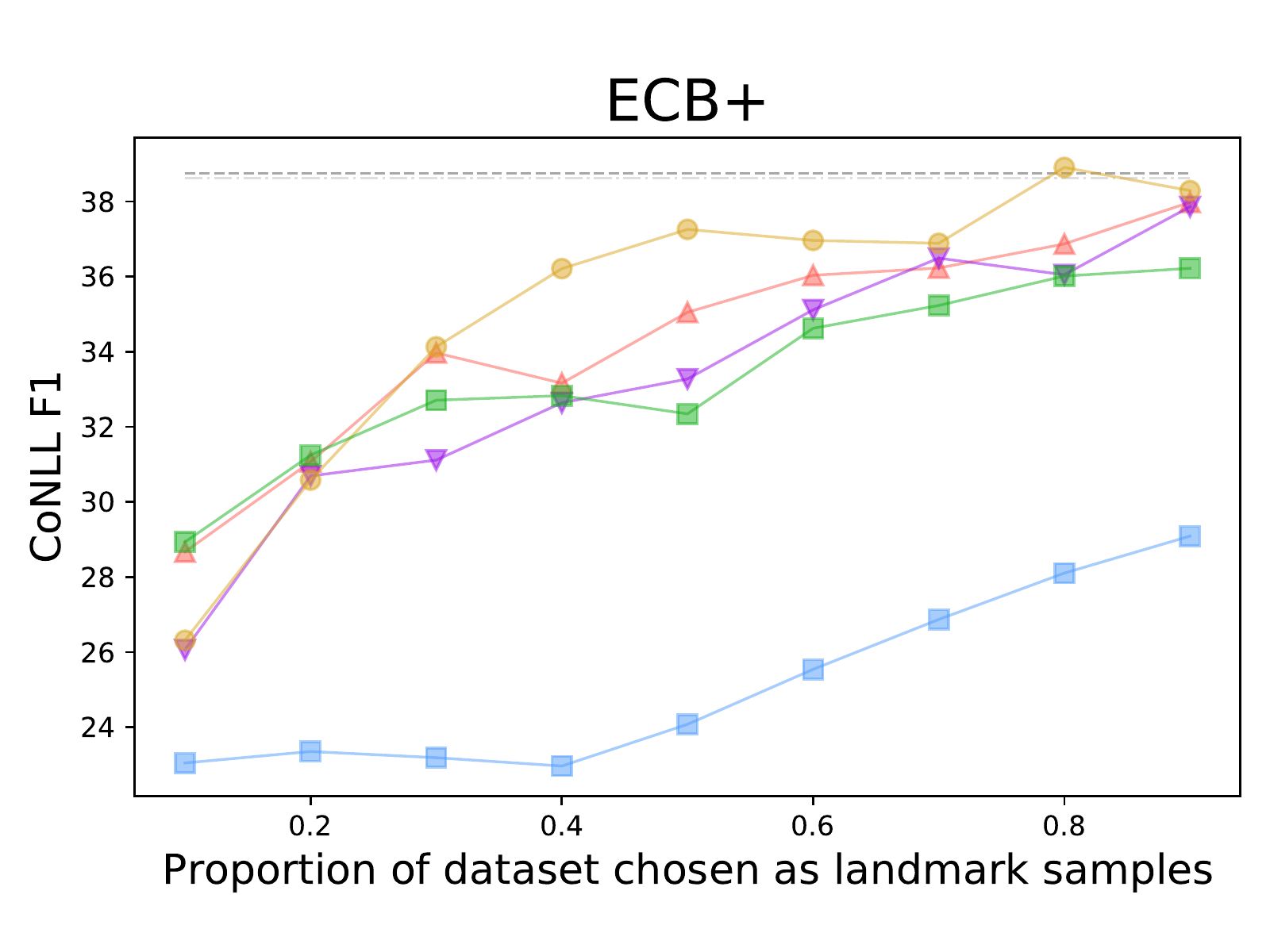}
    \caption{\textbf{Comparison of Re-scaled and Non-Rescaled Methods for Cross-Document Coreference}}
    \label{fig:coref_rescaled}
    \vspace{-2em}
\end{figure*}
\section{Conclusion}\label{sec: conclusion}

We have shown that indefinite similarity matrices arising in NLP applications can be effectively approximated in sublinear time. A simple variant of the  \nystrom method, and several simple CUR approximation methods,  all display strong performance in a variety  of tasks. We hope that in future work, these methods can be used to scale text classification and clustering  based on cross-encoder, word mover's distance, and other expensive similarity functions, to much larger corpora.

\subsection*{Acknowledgements}

This work supported in part by the Center for Data Science, in part the Center for Intelligent Information Retrieval, in part by the National Science Foundation under Grants No. 1763618 and 2046235, in part by International Business Machines Corporation Cognitive Horizons Network
agreement number W1668553, along with an Adobe Research Grant. 
Some of the work reported here was performed using high performance computing equipment obtained under a grant from the Collaborative R\&D Fund managed by the Massachusetts Technology Collaborative. 
Any opinions, findings and conclusions or recommendations expressed here are those of the authors and do not necessarily reflect those of the sponsor.

\bibliography{refs}

\clearpage

\appendix

\begin{center}
    \textbf{Supplementary: {Sublinear Time Approximation of Text Similarity Matrices}}
\end{center}

\section{Word Mover's Distance Approximation -- Omitted Details}\label{app:wme}
\smallskip

\noindent\textbf{Hyper-parameter optimization.} 
For each experiments using \nystrom and CUR, we uniformly sample documents from the dataset as landmark samples. Since \nystrom requires random choice of samples, we run it 20 times and report average accuracy. The range of values experimented for $\gamma$ is range $[0.0001, 1.58]$, $s_2$ ranges in $[100, 550]$ for small rank variation and in $[900, 4096]$ for large rank variation, and $\lambda^{-1}$ parameter for LIBLINEAR \cite{fan2008liblinear} ranges in $[1, 1e12]$. For smaller datasets like Twitter, we vary $s_2$ in range $[100, 550]$ for small rank and in range $[900, 1700]$ for large rank. The hyperparameter optimization uses Bayes hyperparameter optimization \cite{shahriari2015taking} to identify the best set of hyperparameters for each approximation algorithm. Separate searches were done to avoid Bayesian optimization from quitting smaller ranks in favor of large ranks to identify the best set of hyperparameters. 
\smallskip

\noindent\textbf{Running time.} Computing the similarity matrix requires $O(n^2)$ evaluations of the similarity metric (WMD computaiton time is $O(L^3\log(L))$, where $L$ is the size of the longest document). For SMS-Nystr\"om we only require $O(ns)$ similarity evaluations and $O(s^3)$ evaluation for the SVD and $s\ll n$. WME can be computed in $nD^2L\log(L)$, where $D$ is the length of random documents. Thus both SMS-Nystr\"om and WME are faster than evaluating true WMD. We thus compare our method to WME to show the loss of speed up using SMS-Nystr\"om. To compare runtimes we check how long each of WME and SMS-Nystr\"om take to generate the feature set of a given dataset with fixed set of hyperparameters. These are computed using the hyperparameters which give best performance for corresponding algorithms and datasets. It is important to note that SMS-Nystr\"om (SR) performs very close to WME(LR), thus comparing those numbers helps justify its use. Our findings are summarized in Table \ref{tab: runtime comp}.
\begin{table}[h]
\setlength{\tabcolsep}{2.5pt}
\small
    \begin{center}
    \begin{tabular}{lccccclc}
        \toprule
        \bf Method & \bf Twitter & \bf Recipe\_L & \bf Ohsumed & \bf 20News\\
        \midrule
        WME(SR) & $13.02$ & $5639.06$ & $85.43$ & $2712.12$\\
        SMS-N(SR) & $86.23$ & $13979.01$ & $629.54$ & $9422.05$ \\
        WME(LR) & $102.06$ & $29238.48$ & $2787.00$ & $13021.13$\\
        SMS-N(LR) & $1014.06$ & $223902.32$ & $21246.65$ & $130342.28$\\
        \bottomrule
    \end{tabular}
    \end{center}
    \vspace{-1em}
        \caption{\small\textbf{Runtime comparisons.} Comparing the total run-time of WME and SMS-Nystr\"om (in seconds).}
    \vspace{-2em}
    \label{tab: runtime comp}
\end{table}
\smallskip

\noindent\textbf{Results.} Table \ref{tab: wmd best rank comparison} compares the ranks for the best achieved scores of the corresponding errors. In general the ranks of \nystrom and CUR variant approximations are greater than WME. This is understandable as the approximation error generally goes down as we increase the number of landmark samples chosen. 

We plot the validation accuracy in Figures \ref{fig: SR validation plots} and \ref{fig: LR validation plots}. As can be observed from the plots in Figures \ref{fig: SR validation plots} and \ref{fig: LR validation plots}, \nystrom and SiCUR performs better than StaCUR in most of the datasets. Although for some of the datasets StaCUR's performance is competitive. We also observe that the ranks of approximation used for small rank regions sometimes outperforms WME with much higher rank. This will lead to actually higher cost of computing these approximations. 

\begin{table}[h]
\small
\centering
\begin{tabular}{lccccc@{}}
\toprule
 \textbf{Dataset}                     & \bf Twitter & \bf RecipeL & \bf Ohsumed & \bf 20News \\ \midrule
 \textbf{WME(SR)}                     & $128$   & $500$   & $320$   & $491$  \\ 
 \textbf{\nystrom(SR)} & $421$   & $532$   & $500$   & $550$  \\ 
 \textbf{StaCUR(SR)}                      & $382$   & $548$   & $488$   & $548$  \\ 
 \textbf{SiCUR(SR)}                   & $507$   & $550$   & $550$   & $550$  \\ 
 \textbf{WME(LR)}                     & $896$   & $4096$  & $2500$  & $4096$ \\ 
 \textbf{\nystrom(LR)} & $1631$  & $4061$  & $2500$  & $3872$ \\ 
 \textbf{StaCUR(LR)}                  & $1367$  & $4069$  & $2489$  & $3256$ \\  
 \textbf{SiCUR(LR)}                   & $1204$  & $4030$  & $2500$  & $3824$ \\ \bottomrule
\end{tabular}
    \caption{\textbf{Best performing rank comparison for WMD approximation.} Comparing the ranks of the respective algorithms to approximate exponentiated WMD matrix.}
    \label{tab: wmd best rank comparison}
    \vspace{-1em}
\end{table}

\begin{figure*}[h]
    \centering%
    \begin{subfigure}
    \centering%
    \includegraphics[width=0.24\textwidth]{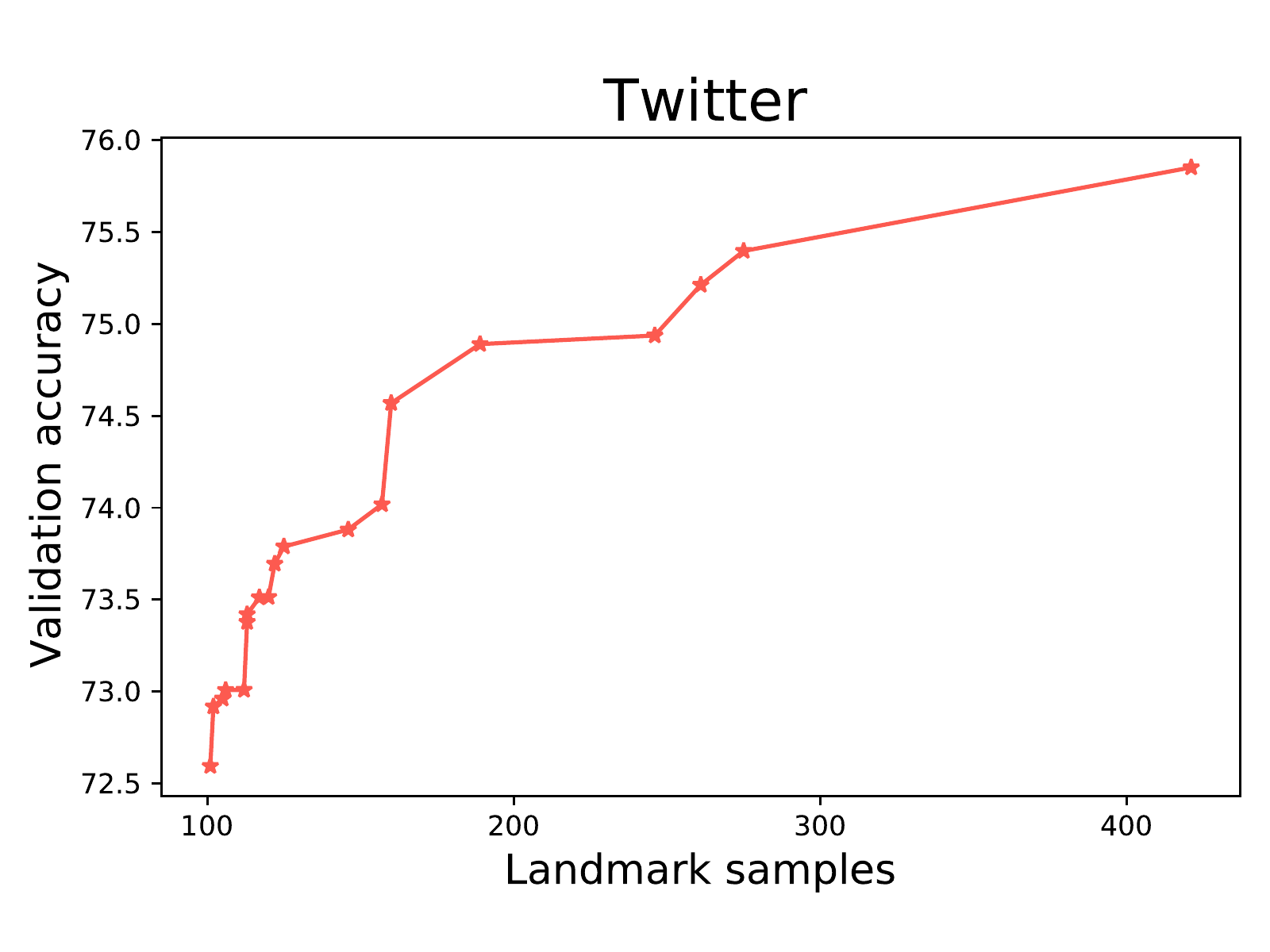}
    \includegraphics[width=0.24\textwidth]{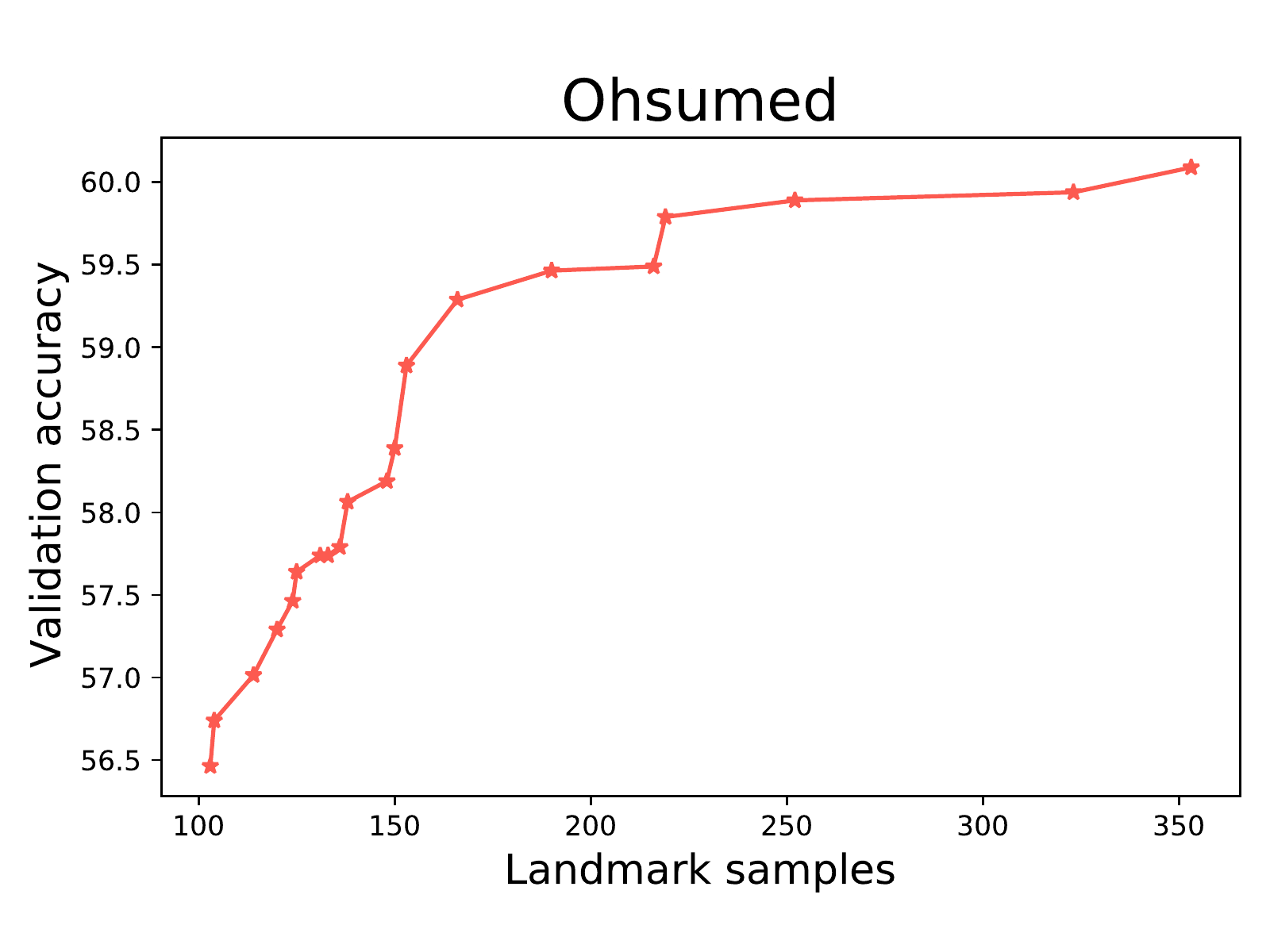}
    \includegraphics[width=0.24\textwidth]{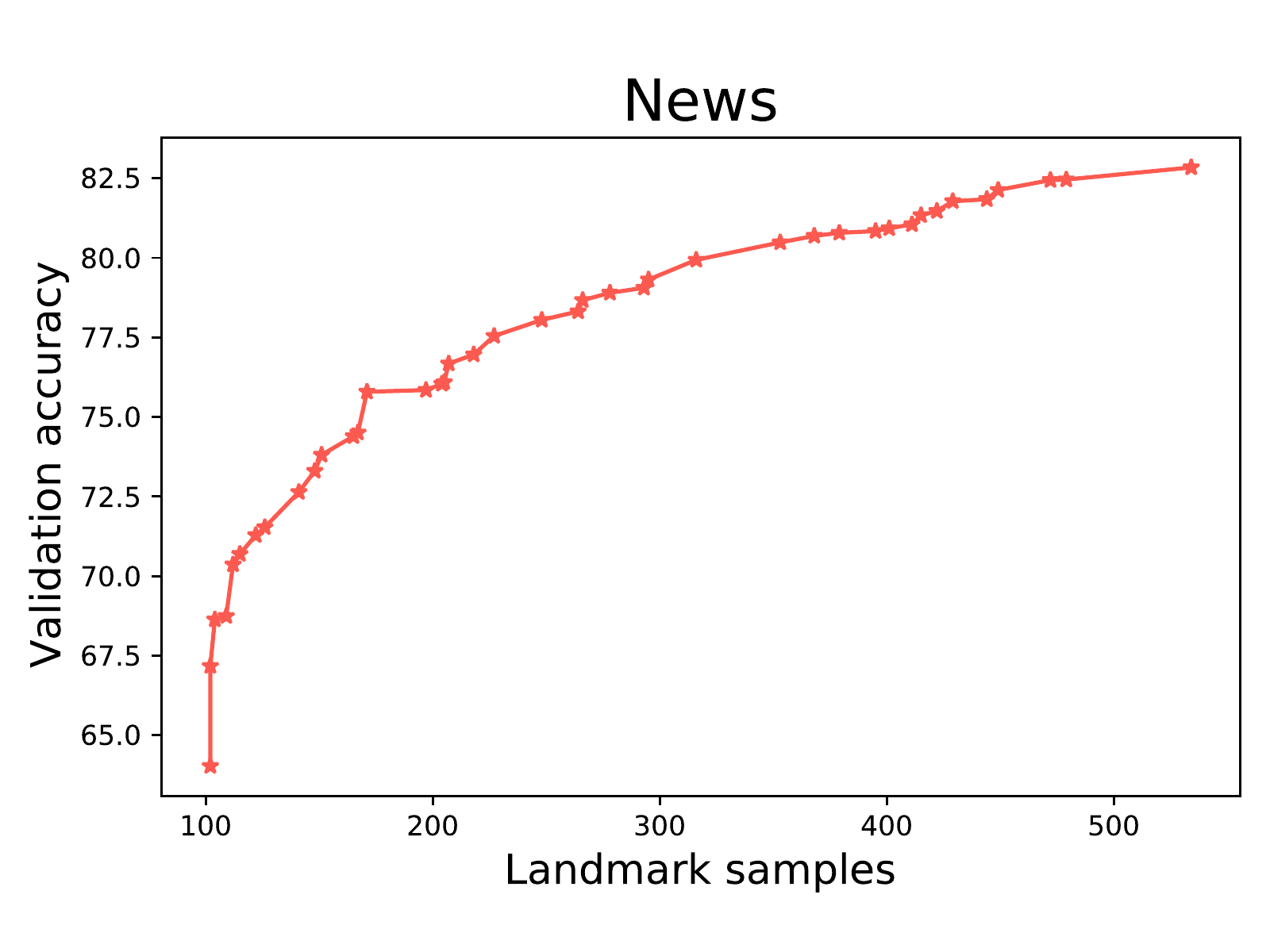}
    \includegraphics[width=0.24\textwidth]{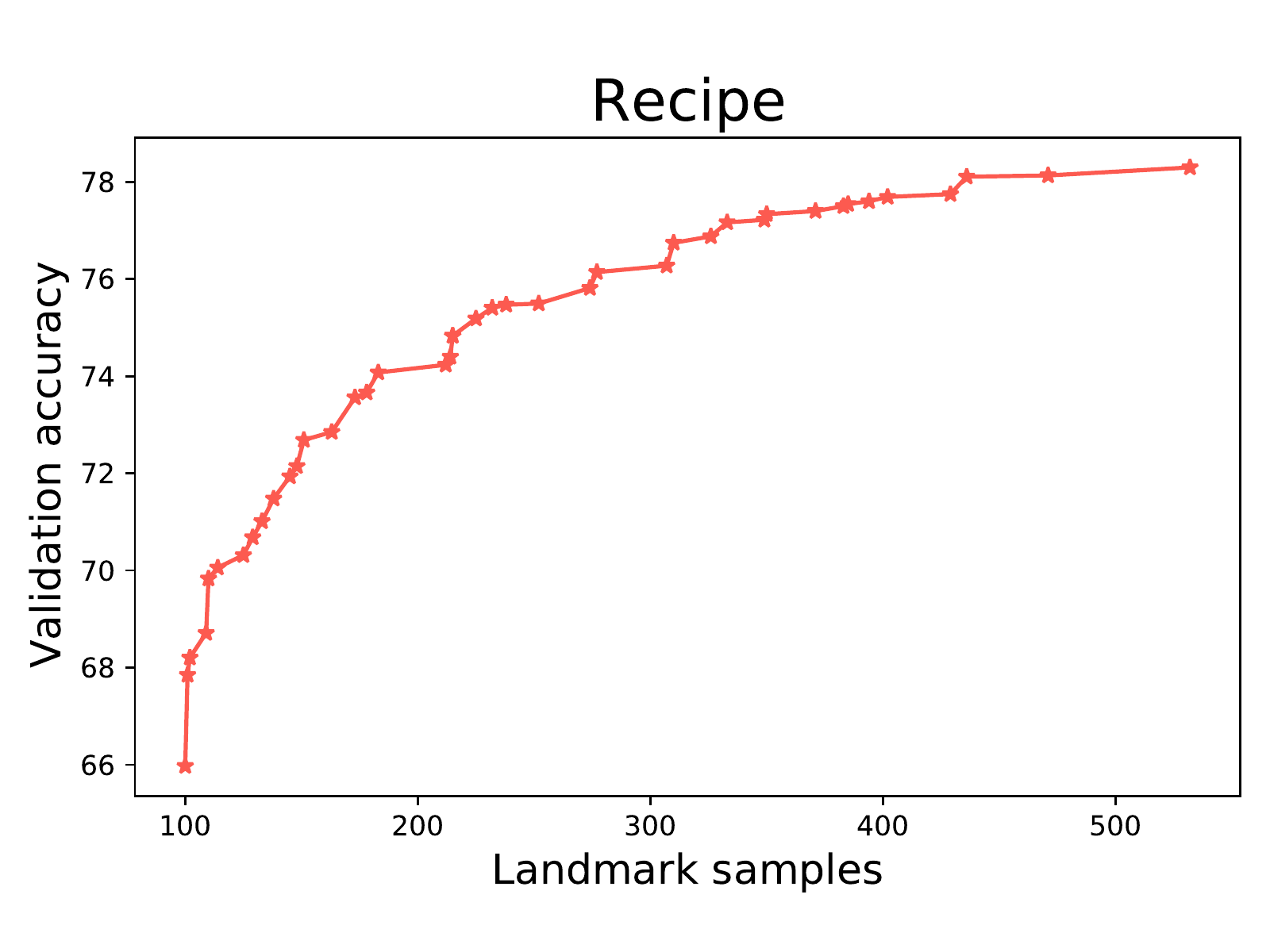}
    \end{subfigure}
    
    \vskip\baselineskip
    \vspace{-2em}
    
    \begin{subfigure}
    \centering%
    \includegraphics[width=0.24\textwidth]{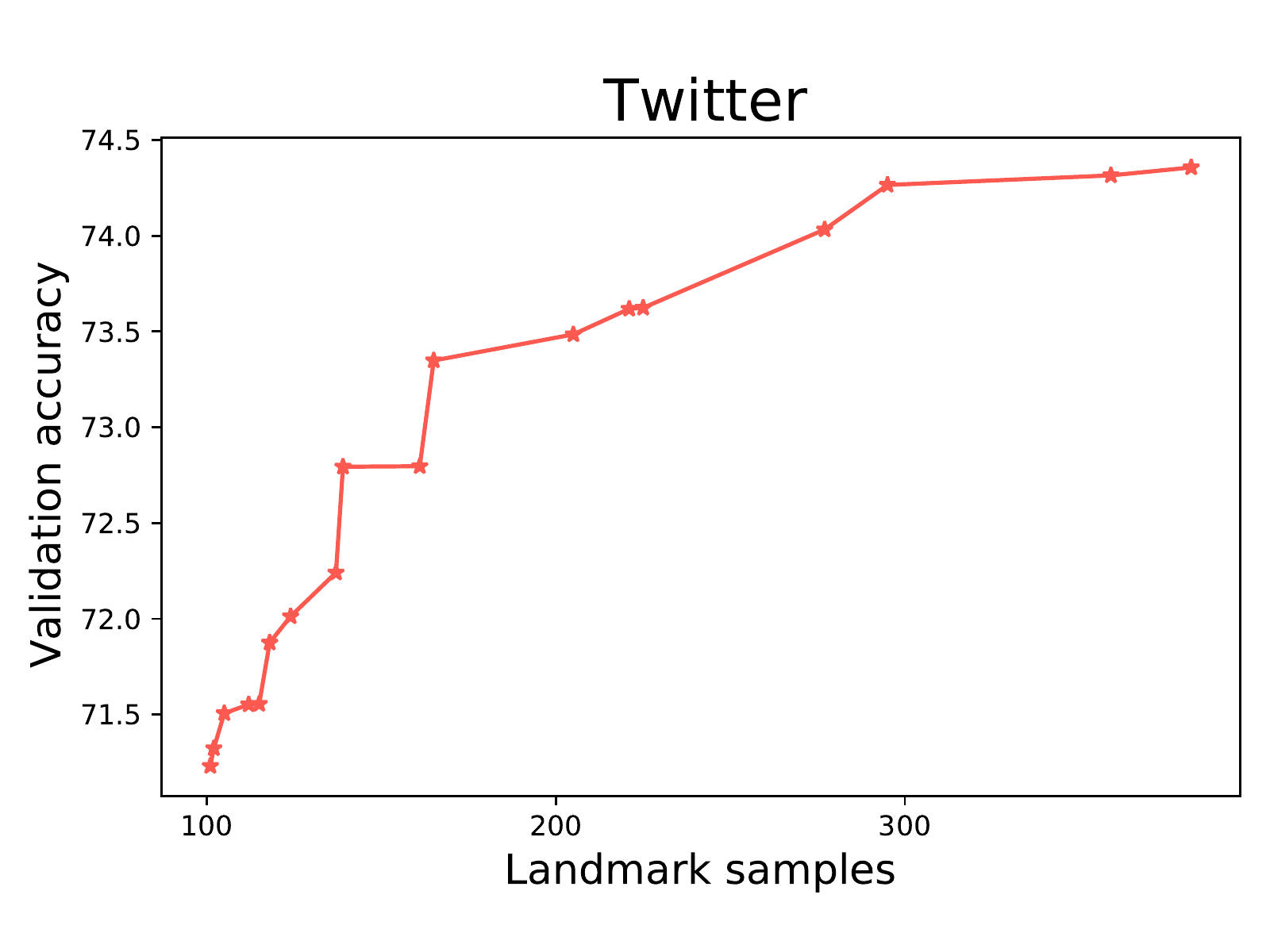}
    \includegraphics[width=0.24\textwidth]{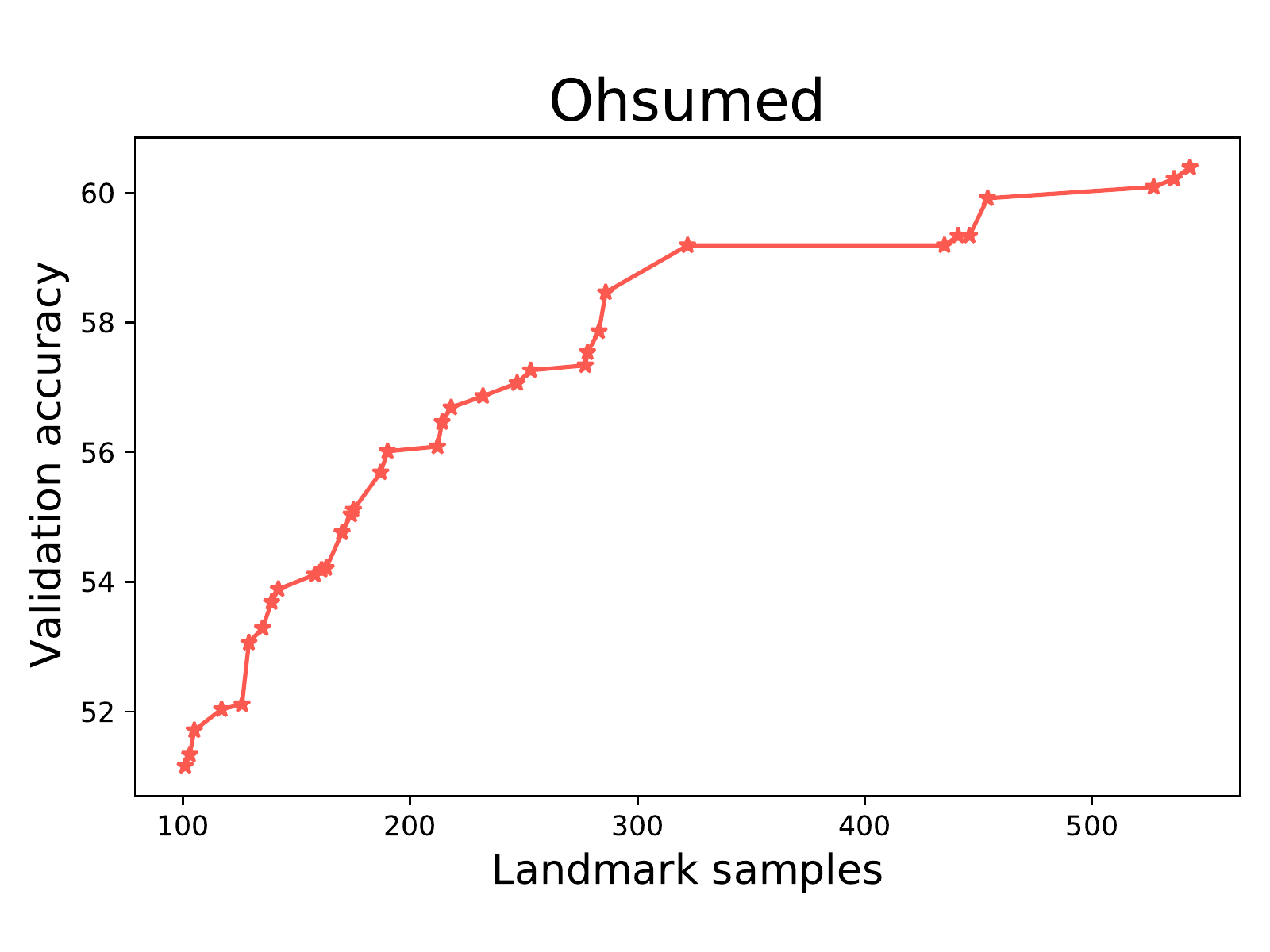}
    \includegraphics[width=0.24\textwidth]{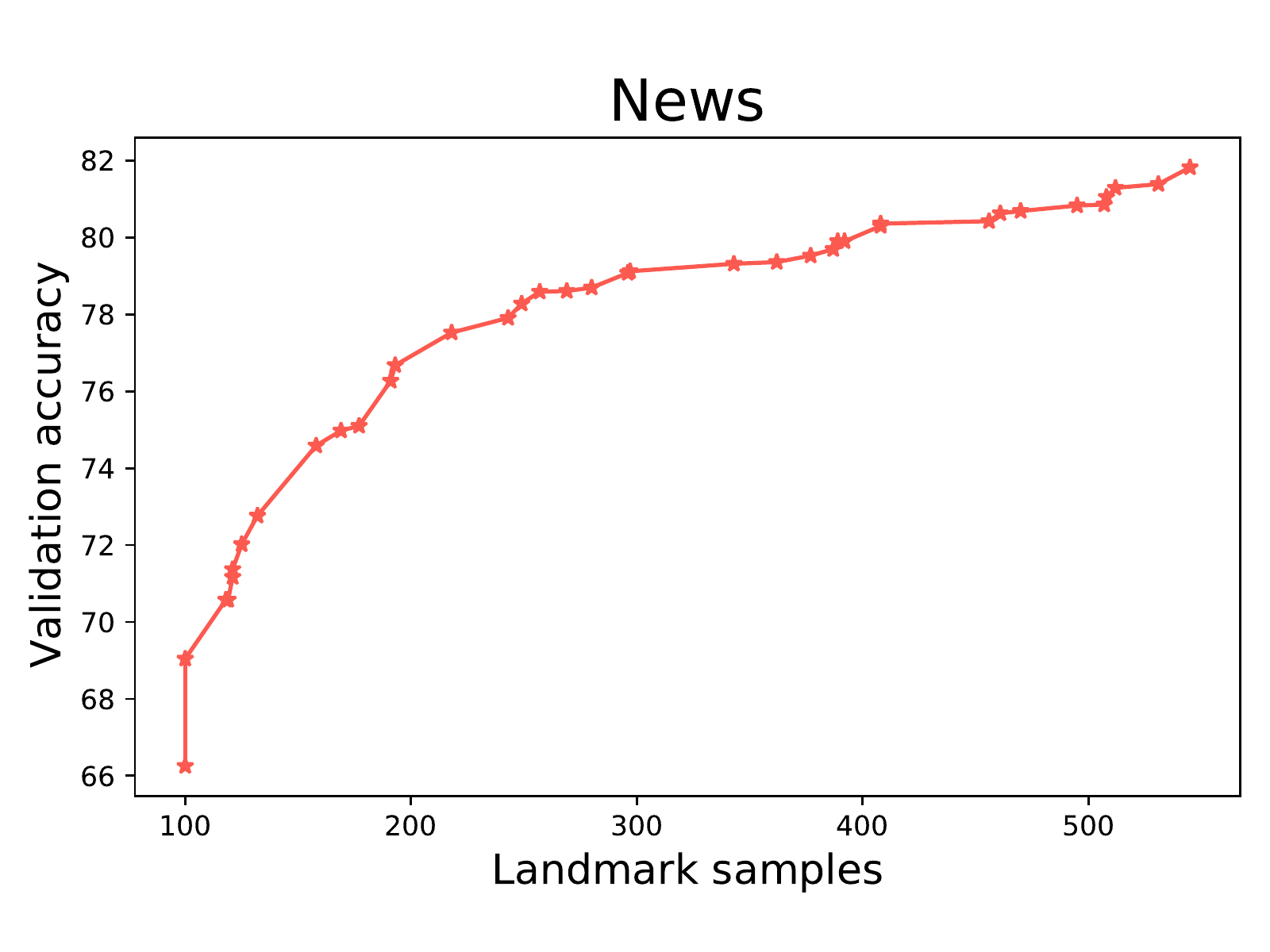}
    \includegraphics[width=0.24\textwidth]{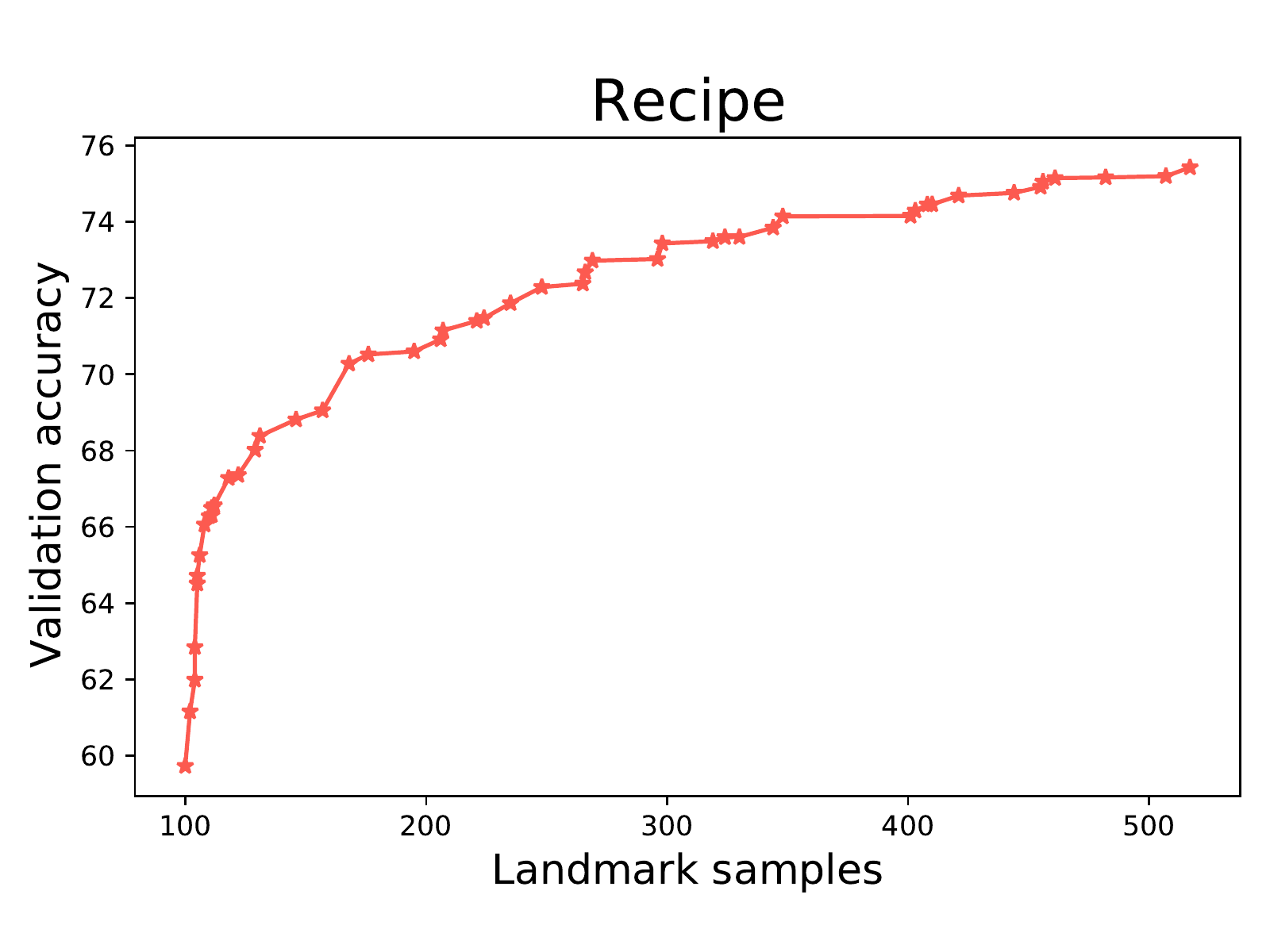}
    \end{subfigure}
    
    \vskip\baselineskip
    \vspace{-2em}
    
    \begin{subfigure}
    \centering%
    \centering%
    \includegraphics[width=0.24\textwidth]{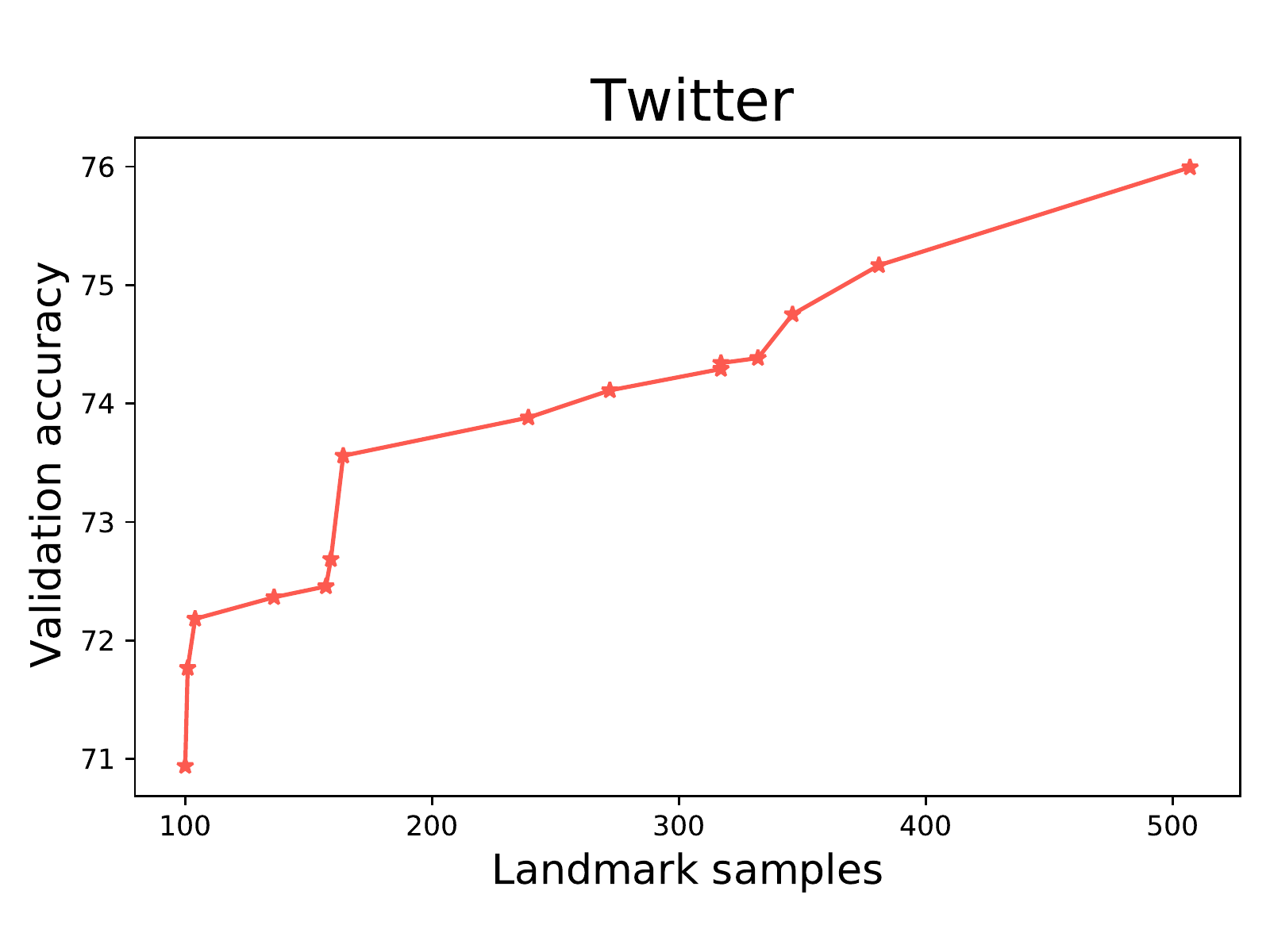}
    \includegraphics[width=0.24\textwidth]{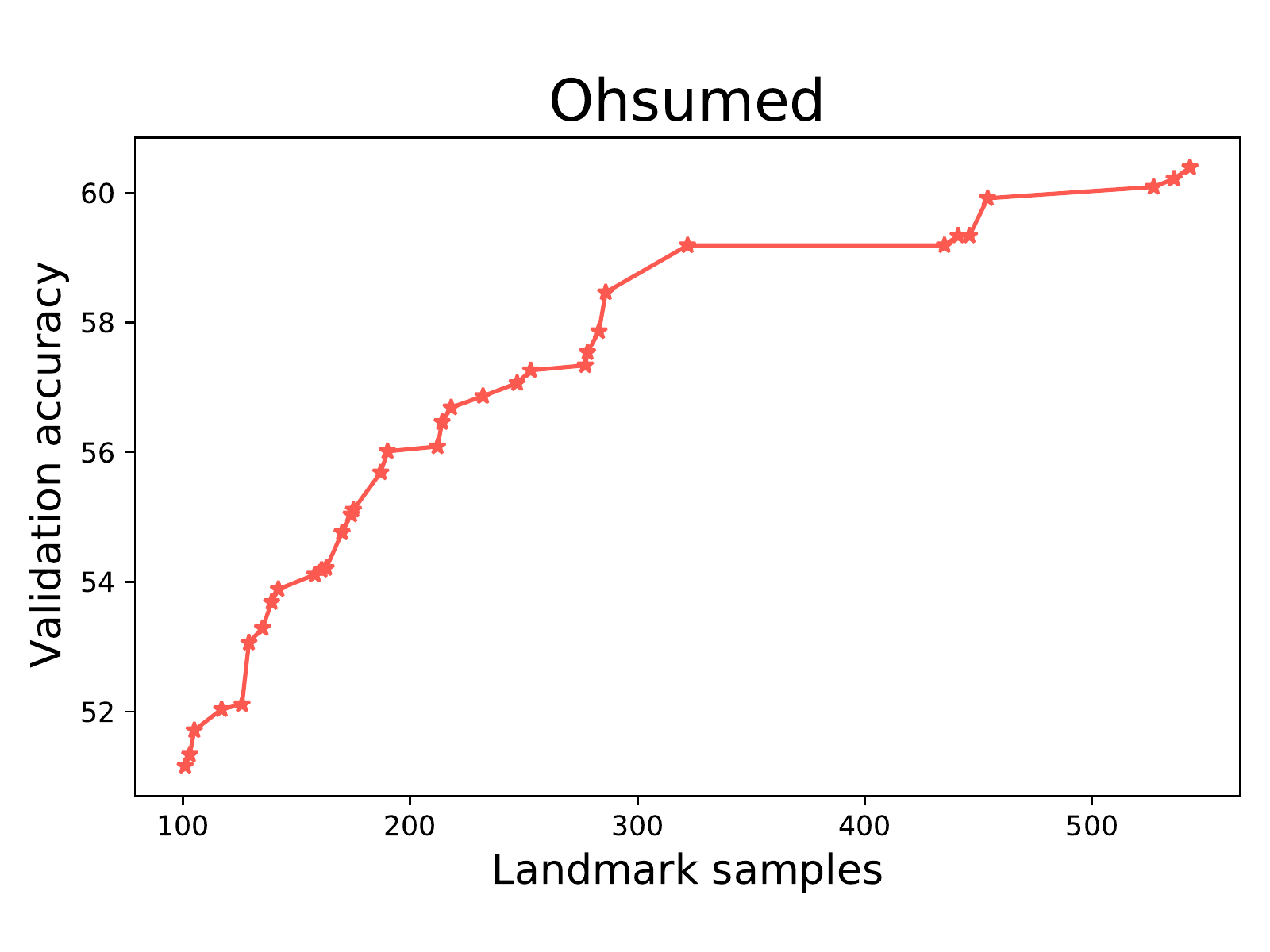}
    \includegraphics[width=0.24\textwidth]{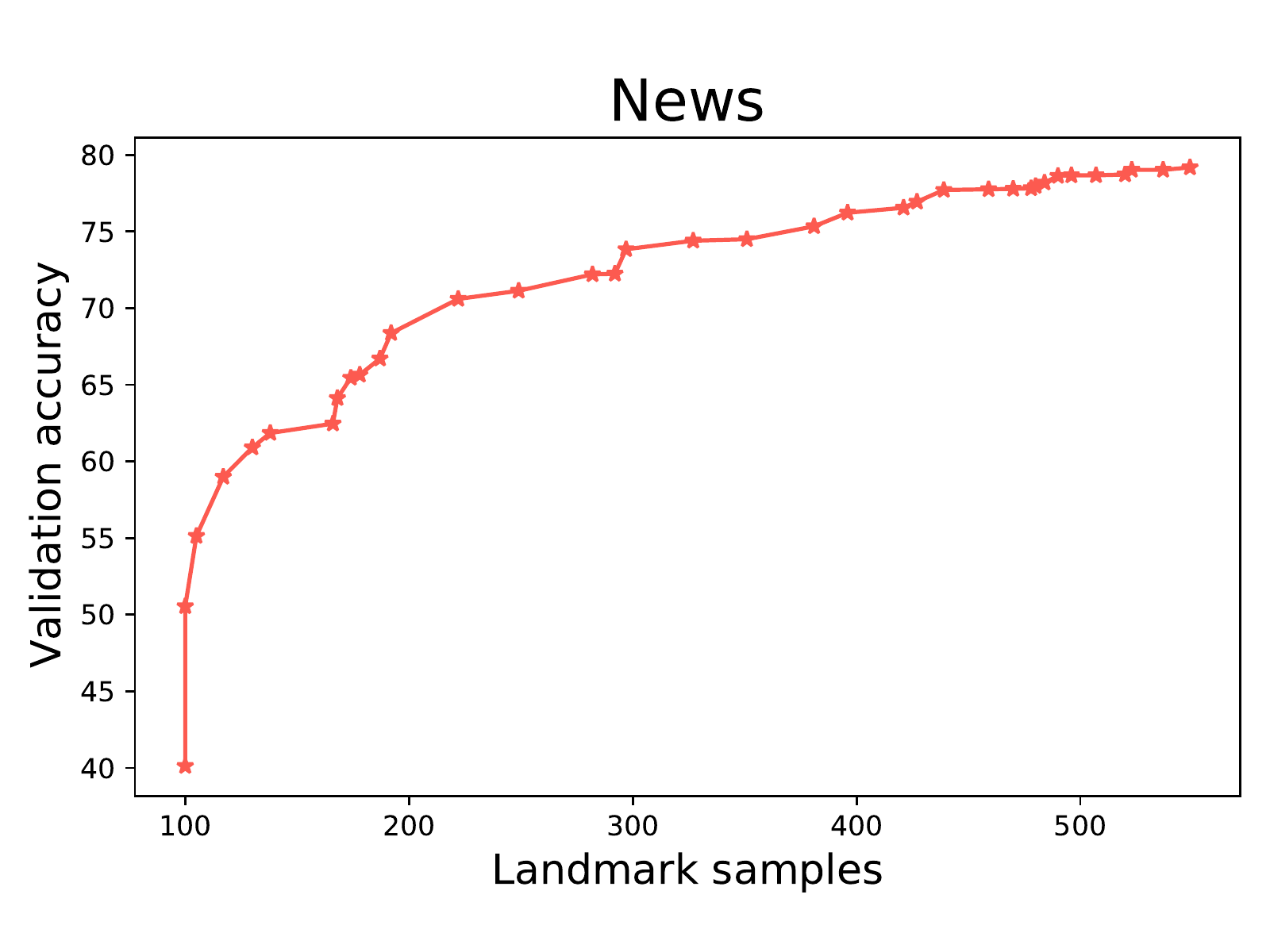}
    \includegraphics[width=0.24\textwidth]{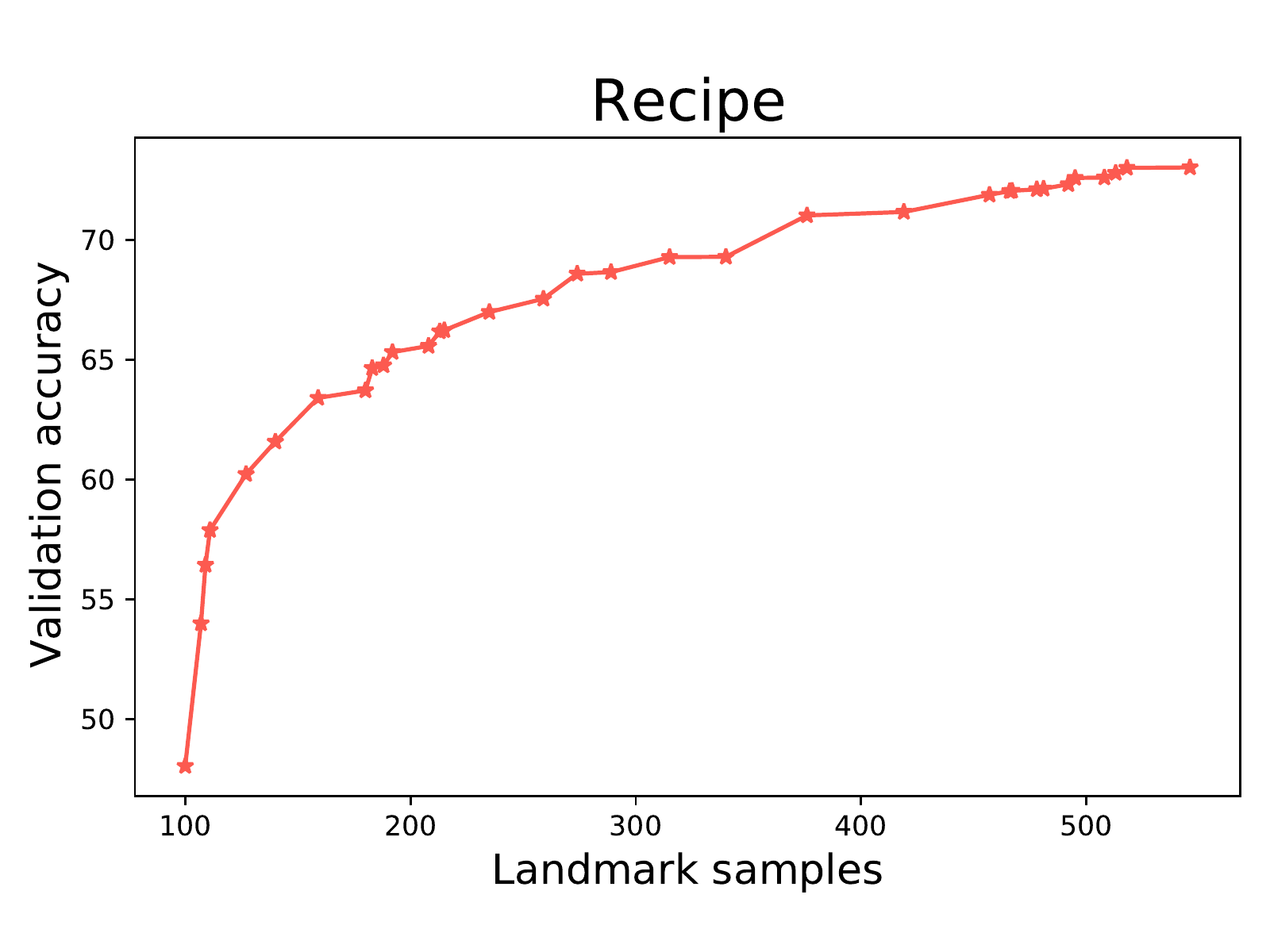}
    \end{subfigure}
    
    \vspace{-3mm}
    
    \caption{\textbf{Mean validation accuracy for small ranks}. Validation accuracy of WMD datasets plotted for hyperparameter optimization using Baye's optimization for small rank ranges. The top row is the validation plot for \nystrom, the middle row is for StaCUR(s) and the bottom row is for SiCUR. We vary the hyperparameters $\gamma, \lambda^-1$ and $s_2$. Unlike the expensive grid-search, Bayesian hyperparameter optimization only searches for viable rank ranges.}
    \label{fig: SR validation plots}
    \vspace{-1.5em}
\end{figure*}

\begin{figure*}[h]
    \centering%
    \begin{subfigure}
    \centering%
    \includegraphics[width=0.24\textwidth]{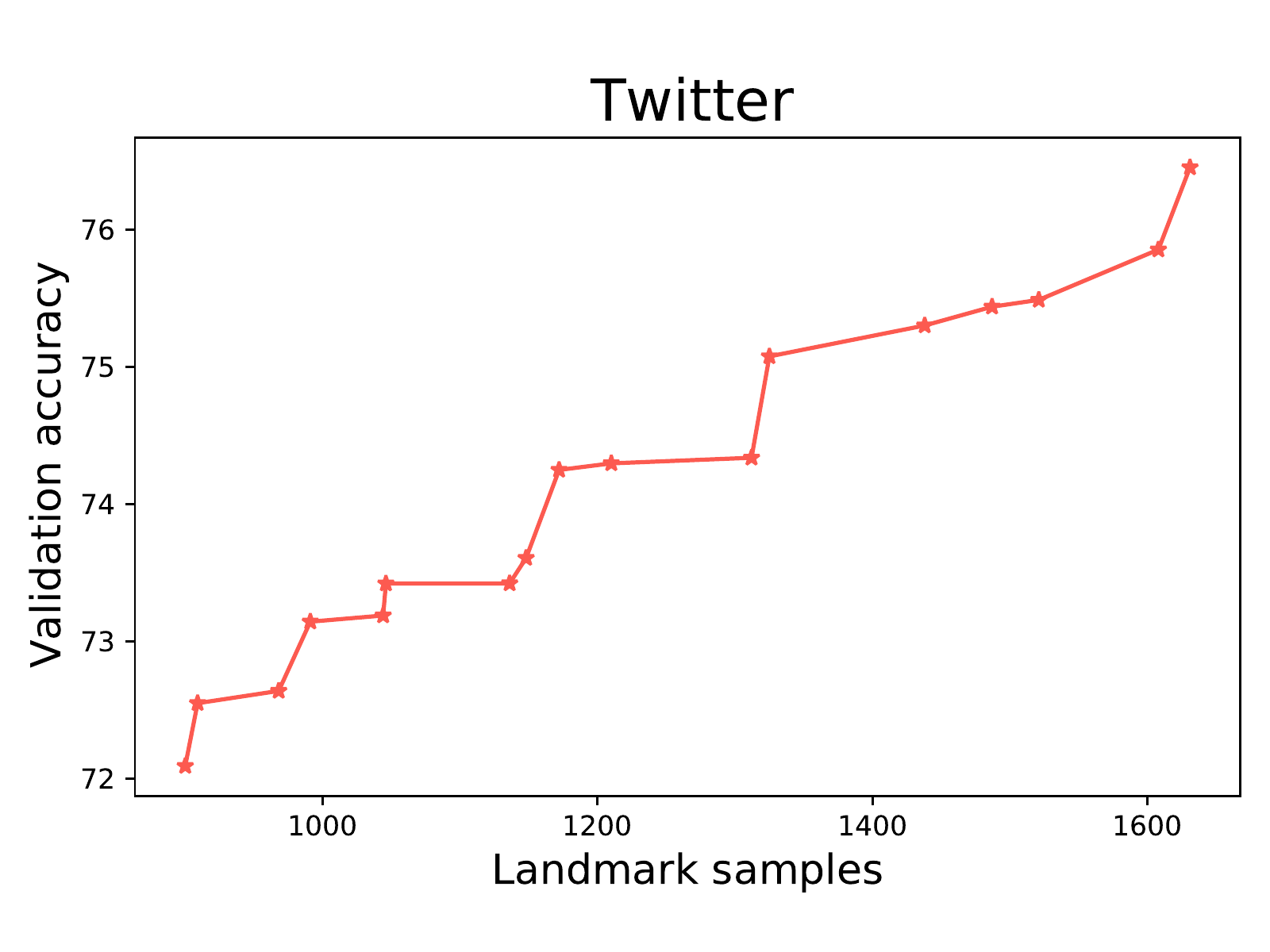}
    \includegraphics[width=0.24\textwidth]{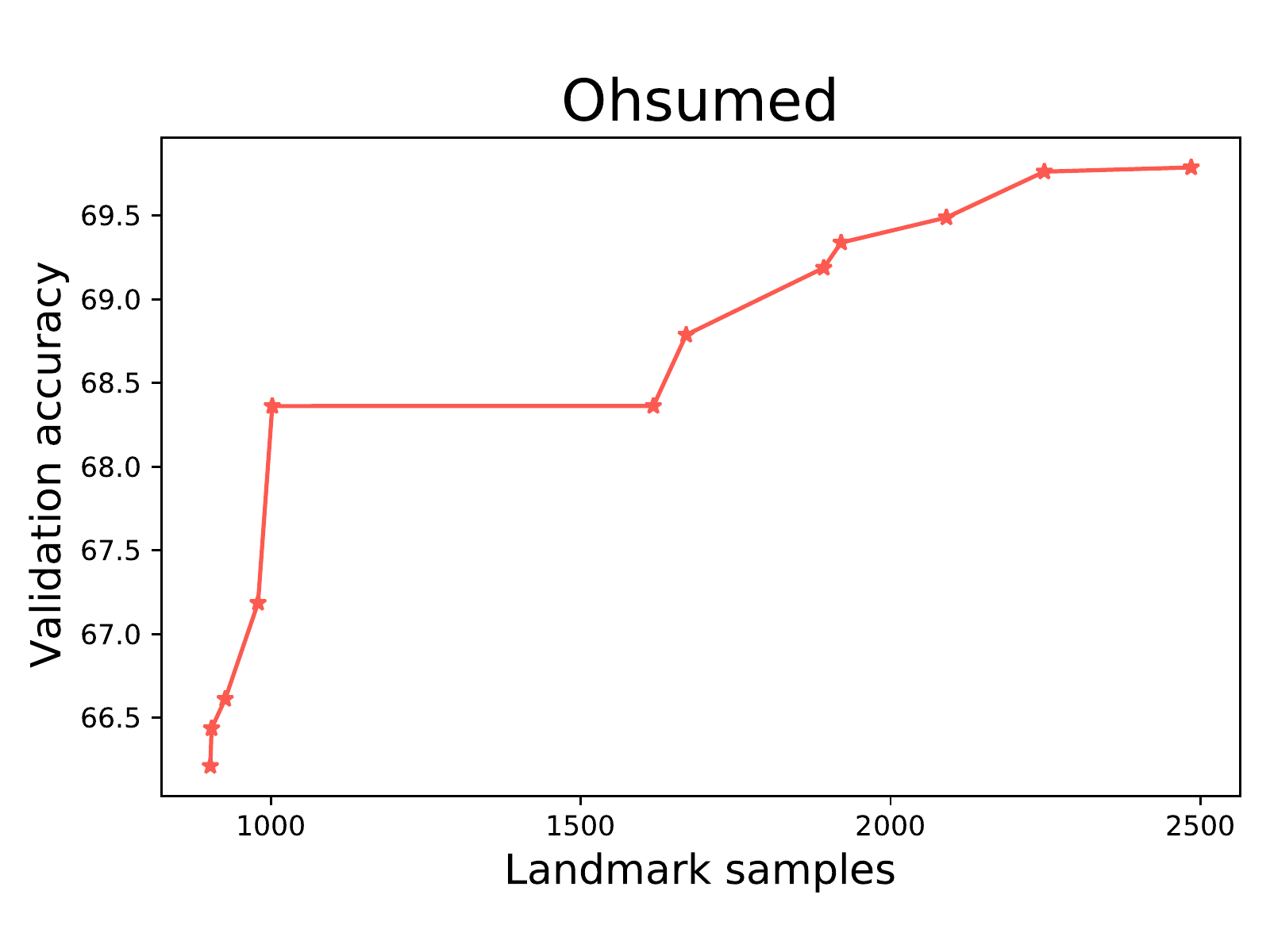}
    \includegraphics[width=0.24\textwidth]{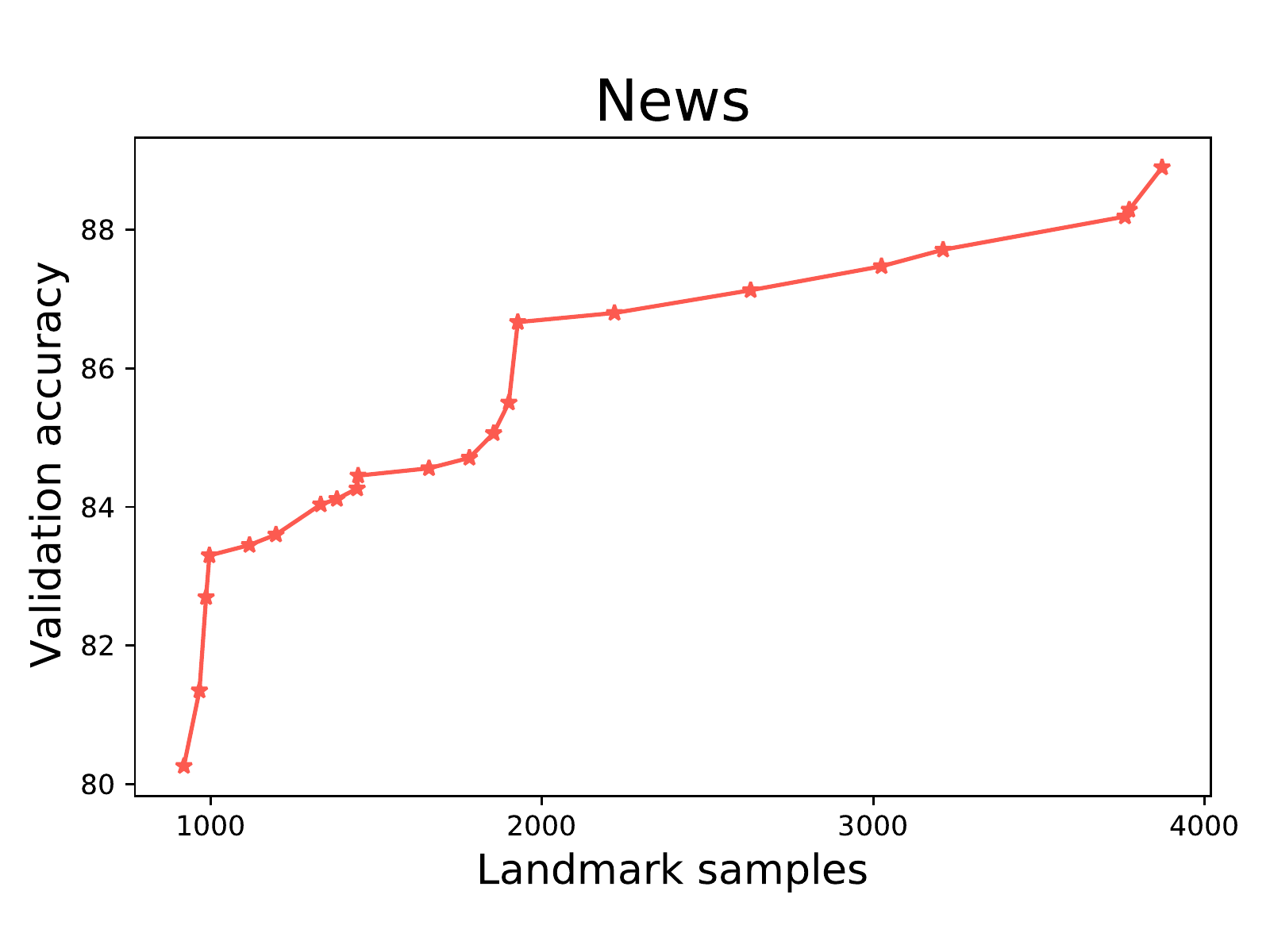}
    \includegraphics[width=0.24\textwidth]{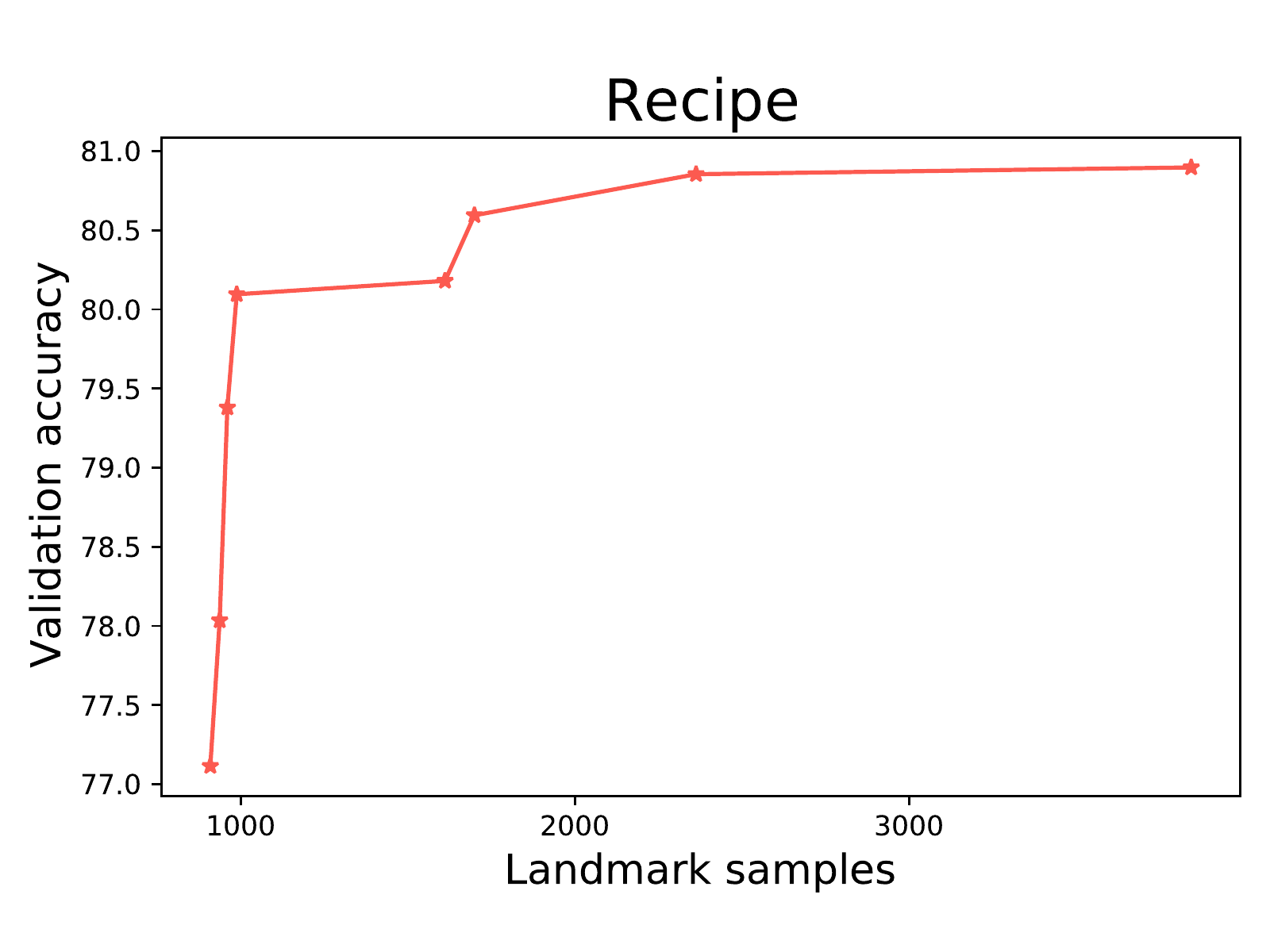}
    \end{subfigure}
    
    \vskip\baselineskip
    \vspace{-2em}
    
    \begin{subfigure}
    \centering%
    \includegraphics[width=0.24\textwidth]{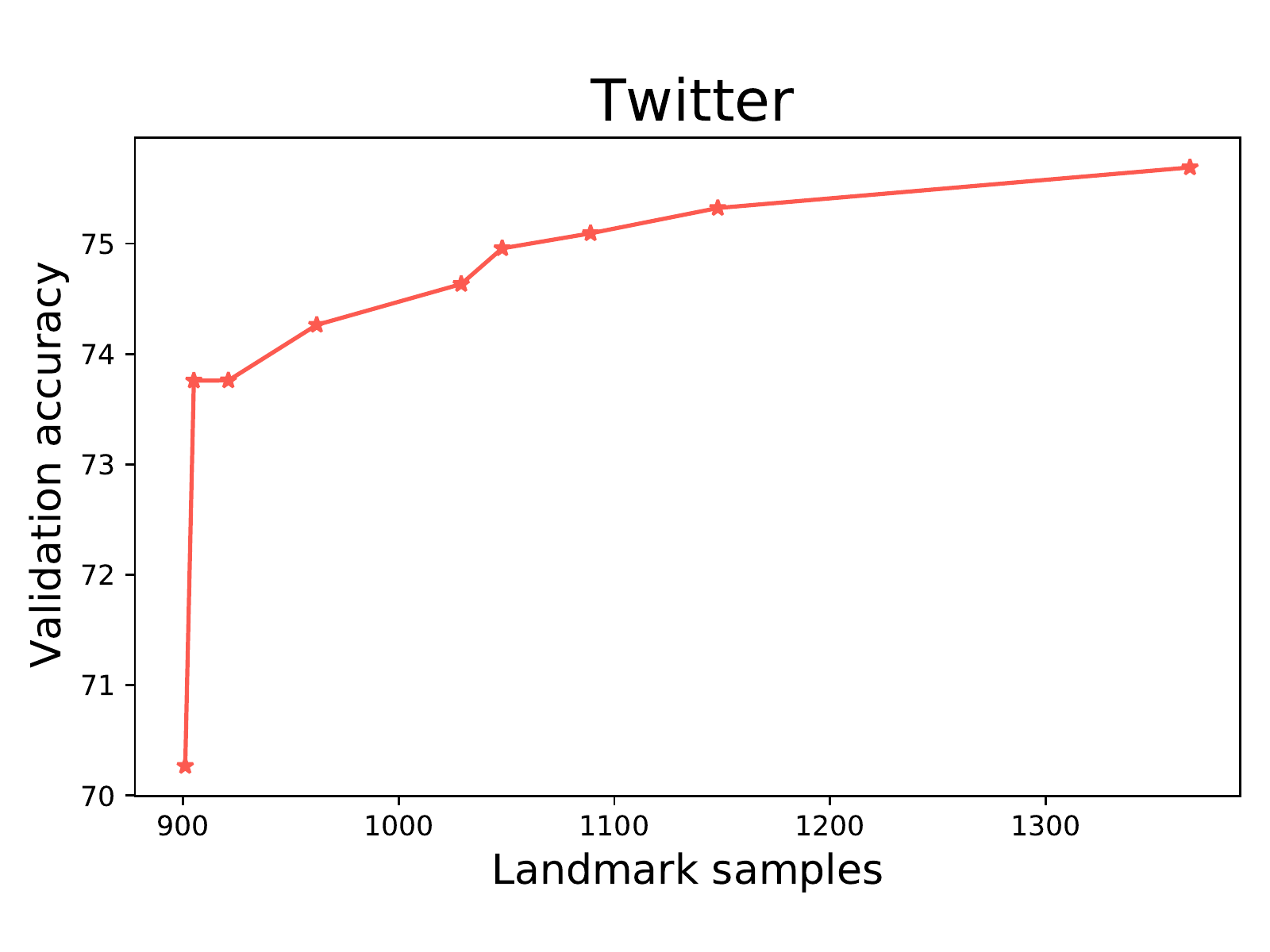}
    \includegraphics[width=0.24\textwidth]{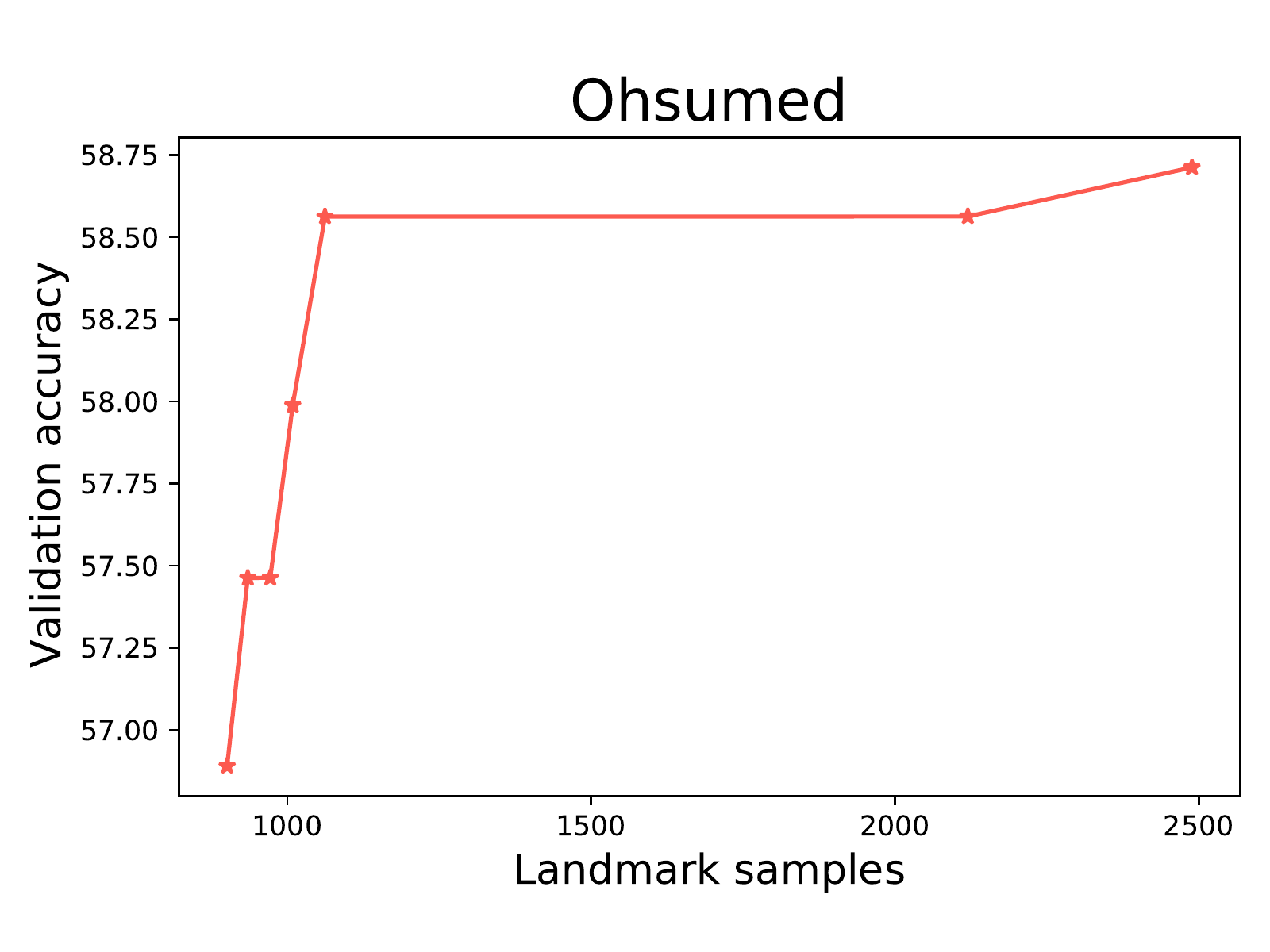}
    \includegraphics[width=0.24\textwidth]{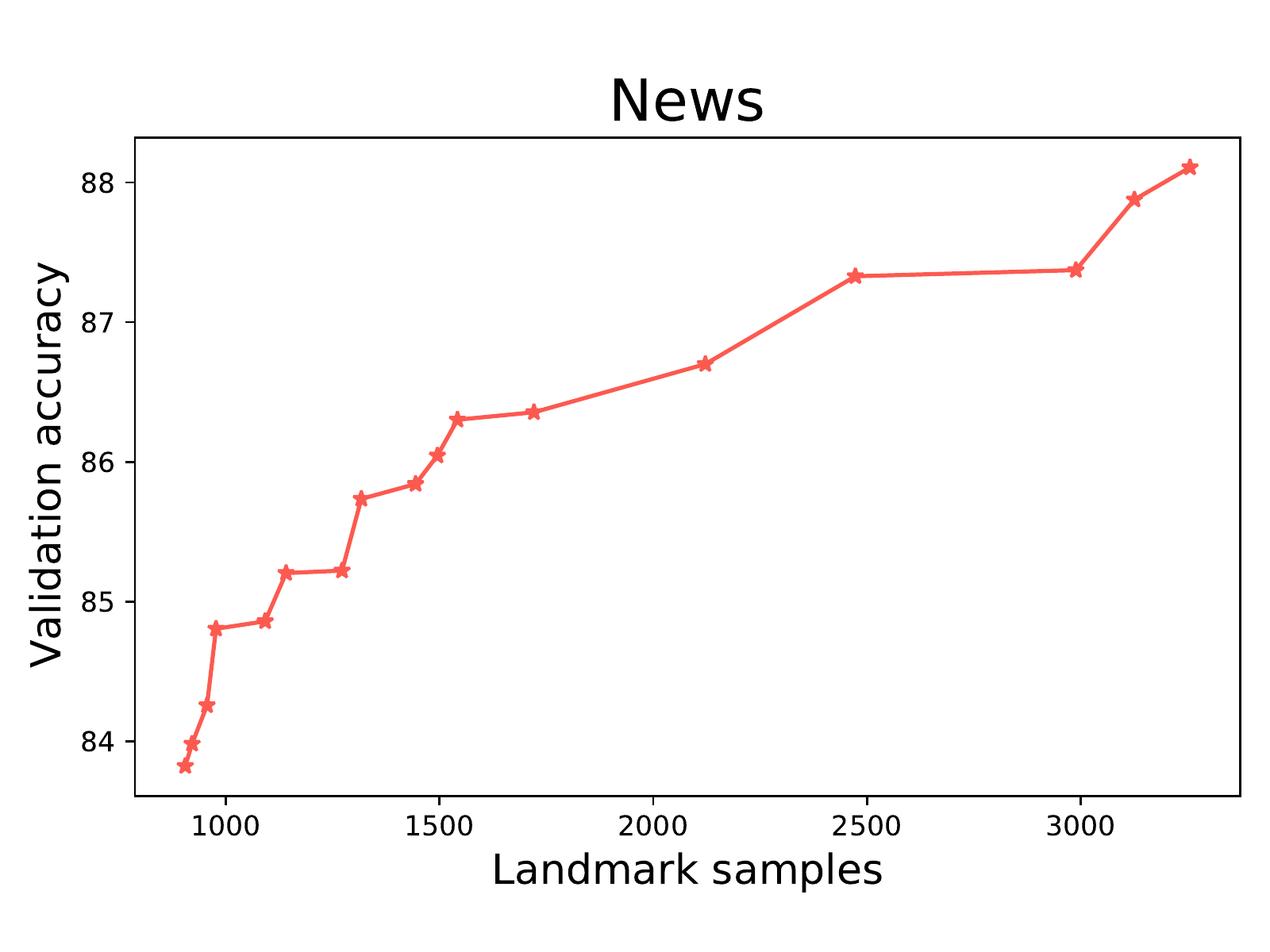}
    \includegraphics[width=0.24\textwidth]{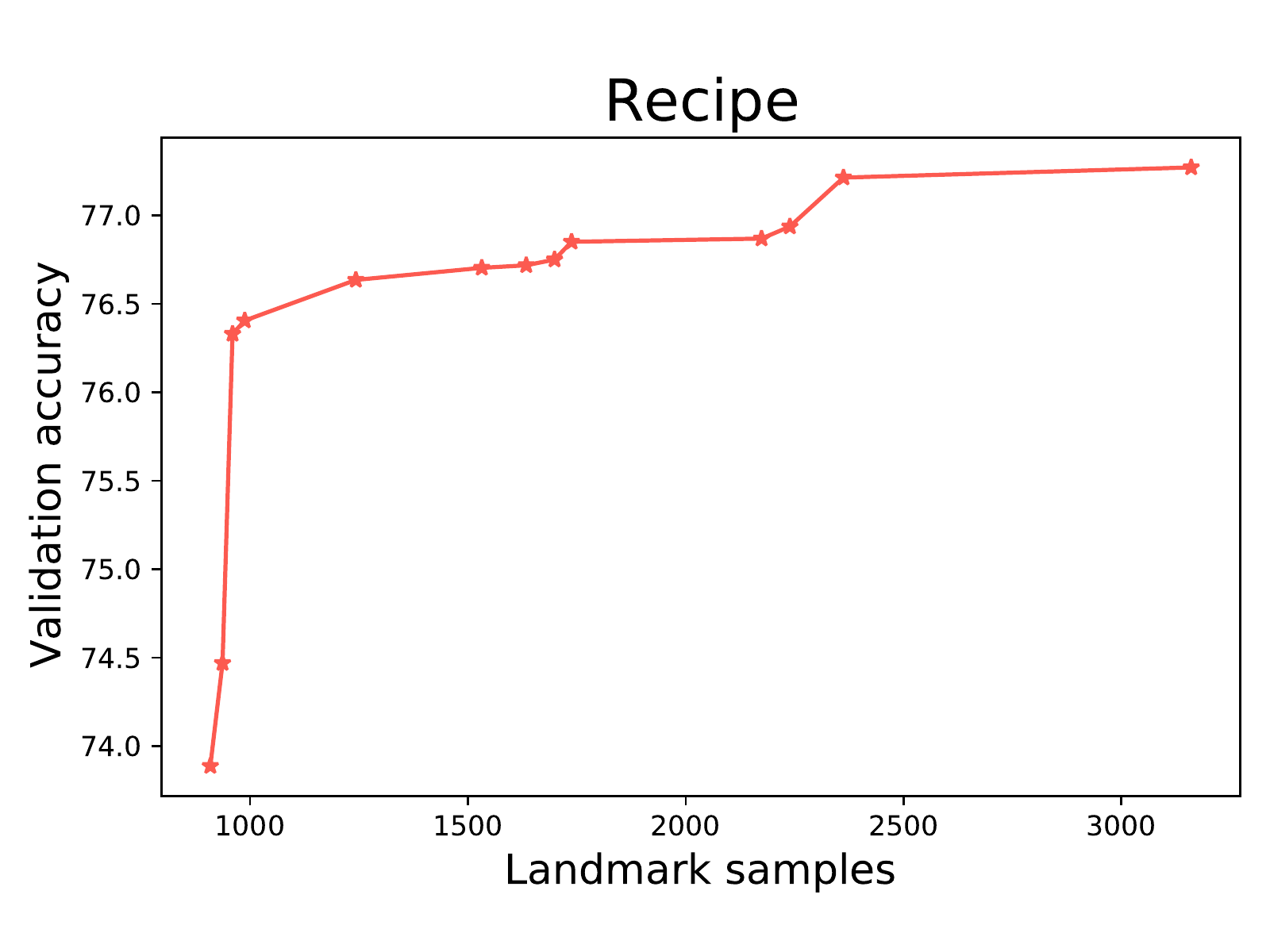}
    \end{subfigure}
    \begin{subfigure}
    \centering%
    \end{subfigure}
    
    \vskip\baselineskip
    \vspace{-2em}
    
    \begin{subfigure}
    \centering%
    \includegraphics[width=0.24\textwidth]{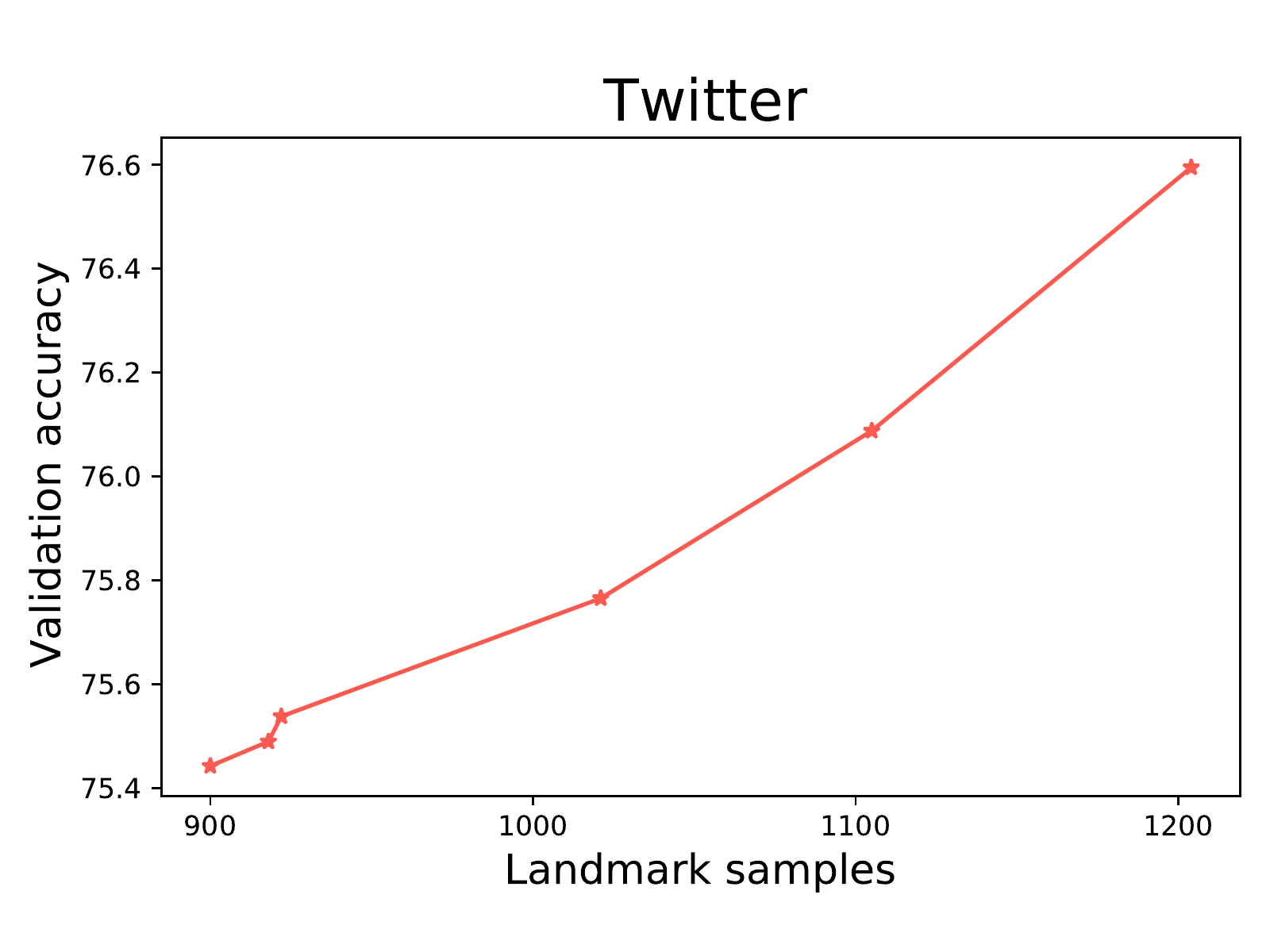}
    \includegraphics[width=0.24\textwidth]{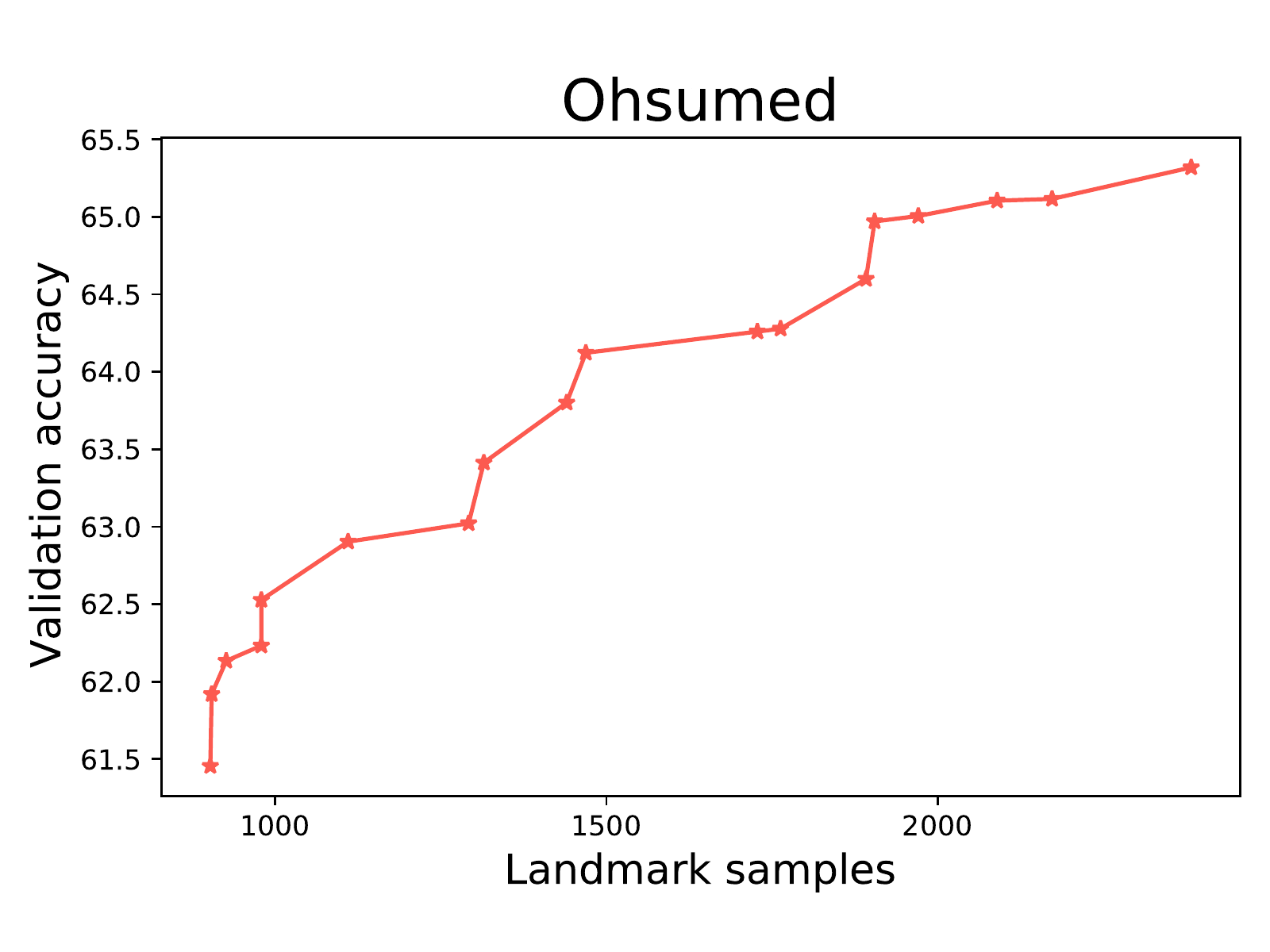}
    \includegraphics[width=0.24\textwidth]{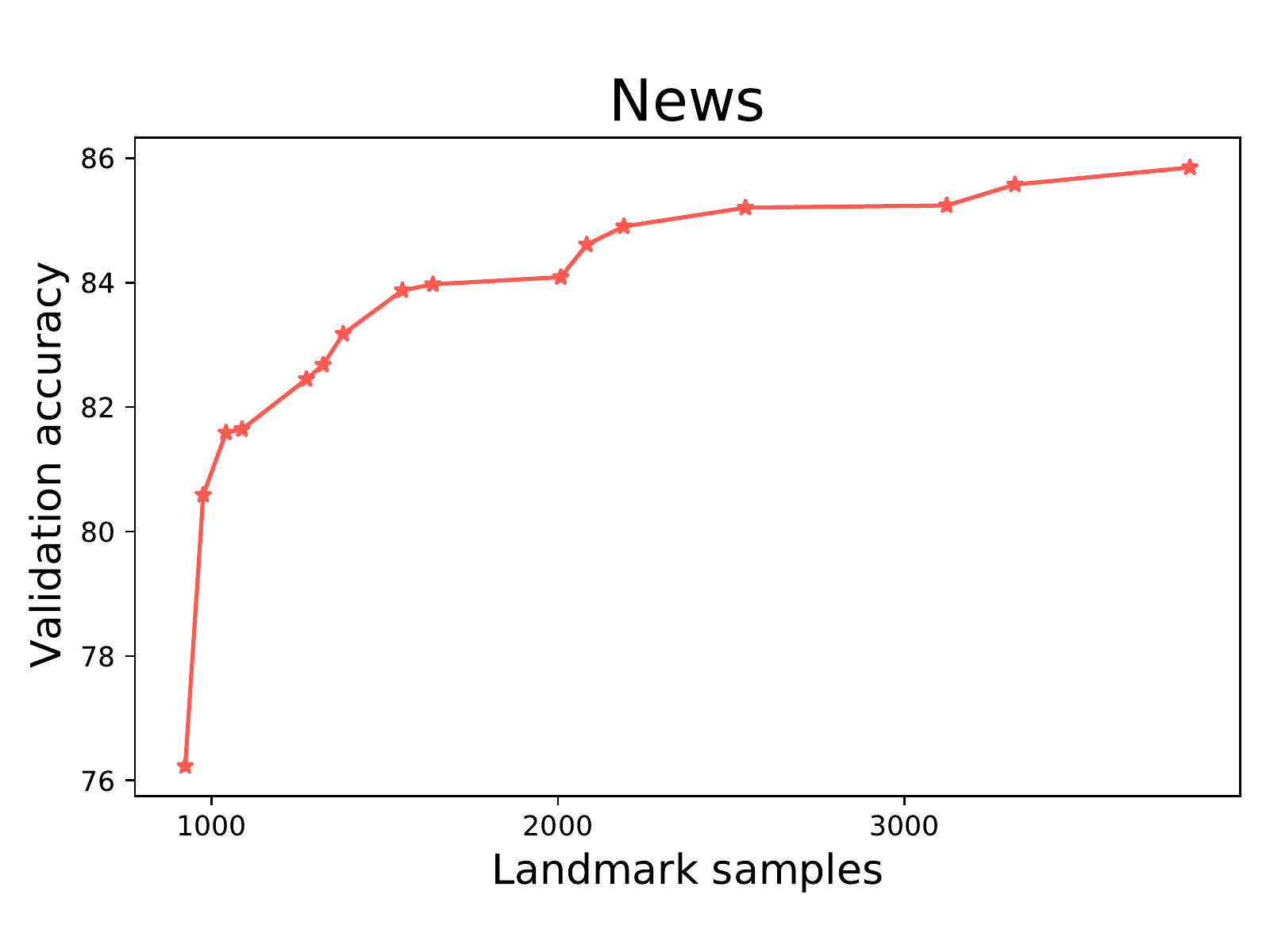}
    \includegraphics[width=0.24\textwidth]{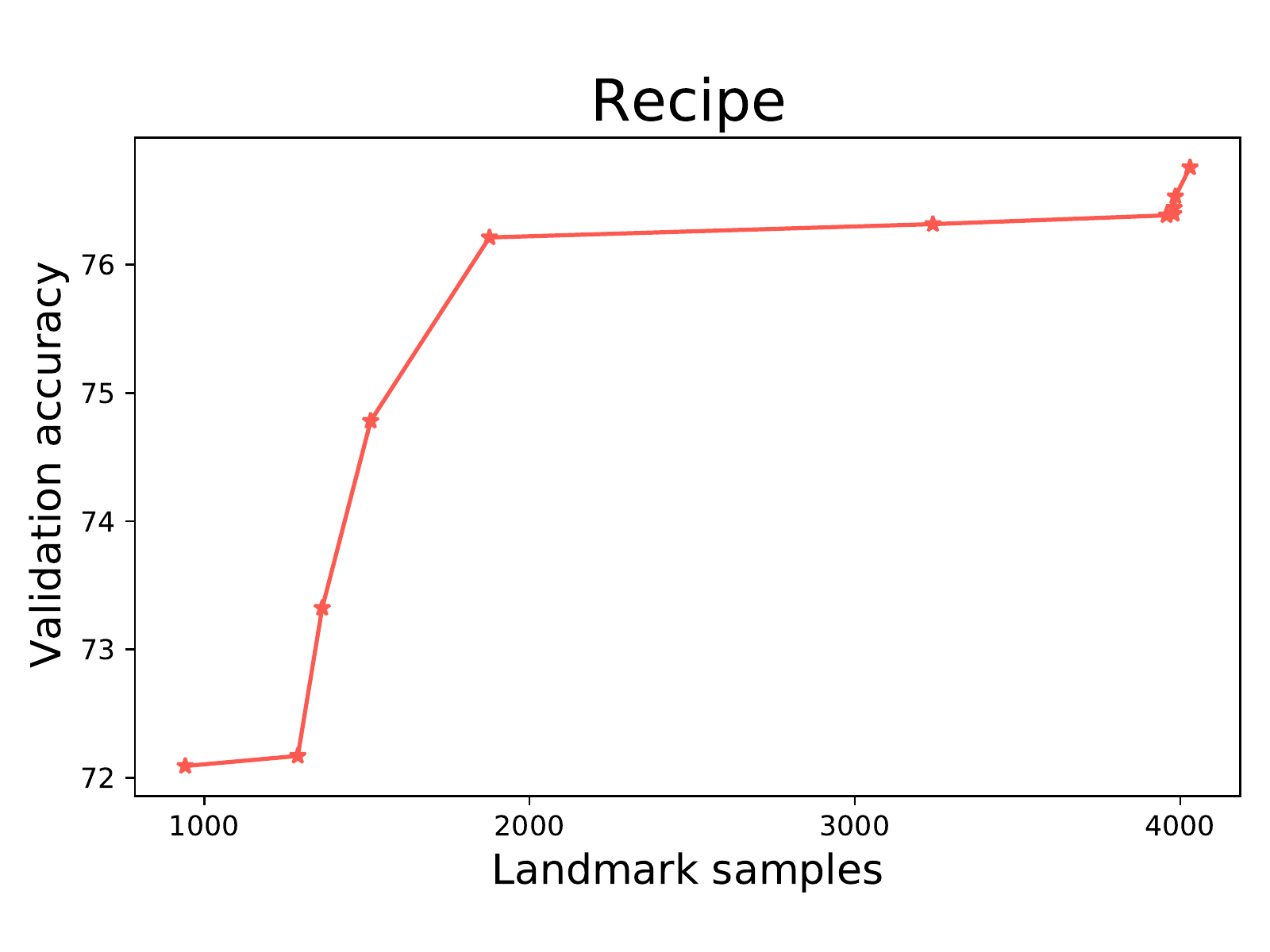}
    \end{subfigure}
    
    \vspace{-3mm}
    \caption{\textbf{Mean validation accuracy for large ranks}. Validation accuracy of WMD datasets plotted for hyperparameter optimization using Baye's optimization for large rank ranges. The top row is the validation plot for \nystrom, the middle row is for StaCUR(s) and the bottom row is for SiCUR.}
    \label{fig: LR validation plots}
\end{figure*}


\section{Choice of sample size and multiplier}\label{app:clarification}
We present here an explanation on why we choose $\alpha=1.5$ and multiplier for $s_2$ is $2$. In Figure \ref{fig:compare_z_a}, we plot the approximation error for two datasets, STS-B and MRPC. Both STS-B and MRPC datasets have a large proportion of negative eigenvalues. Thus any if \nystrom approximation works for any set of values for $\alpha$ and $z$, it should work for reliably for all other datasets. Setting $z=1$ is equivalent to estimating $\lambda_{\min}$ from the sampling matrix $\bv S_1^T\bv K\bv S_1$. In theory we expect $\{z=1, \alpha=2\}$ and $\{z=2, \alpha=1\}$ to behave similarly, but that does not happen. We believe this is because the submatrices for indefinite matrices are often ill-conditioned leading to instability in \nystrom approximation. We observe that for very low values of $z$ and $\alpha$ \nystrom approximation does not work. For values $\alpha\geq 1$ and $z\geq 2$, we can observe that approximation error improves as we increase the number of samples drawn from the dataset. Thus clearly having a two stage sampling one for the obtaining the approximation and the other for obtaining the correction term $\lambda_{\min}$ is helpful. For both the datasets approximation works moderately well for the set $\{z=2, \alpha=1.5\}$, and hence we choose these hyperparameters for all our experiments.

\begin{figure*}[h]
    \centering
    \includegraphics[width=0.48\textwidth]{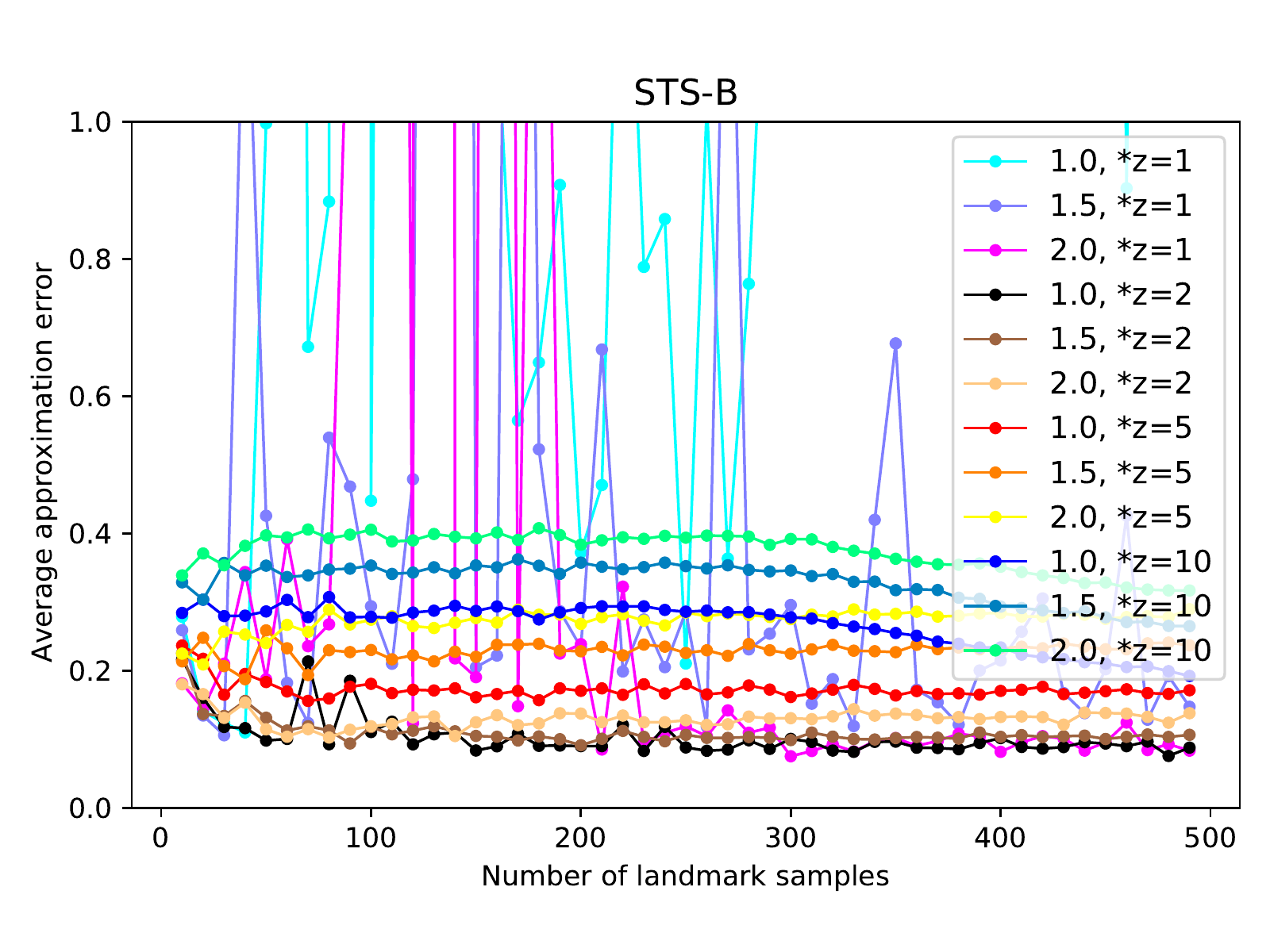}
    \includegraphics[width=0.48\textwidth]{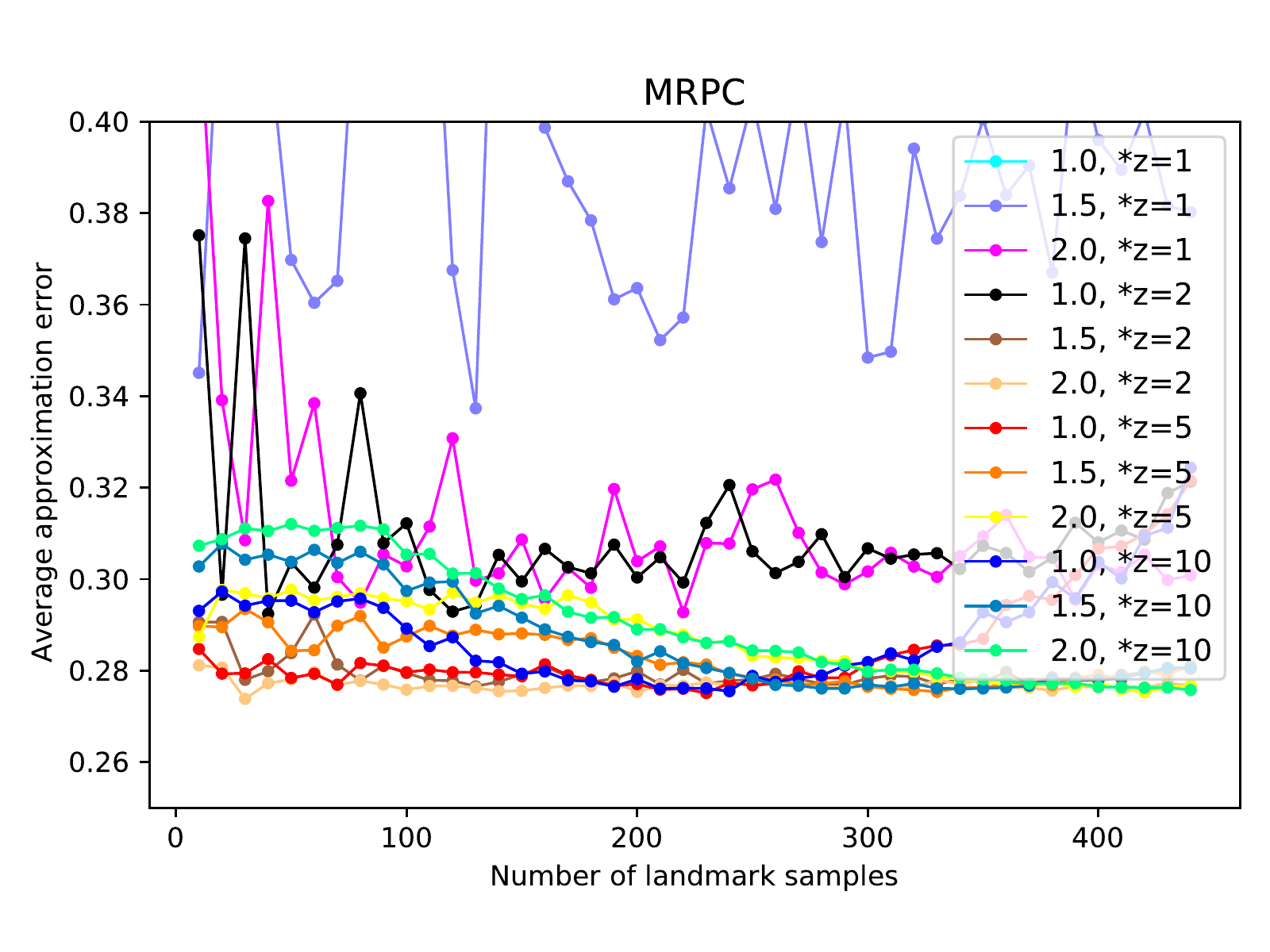}
    \caption{\textbf{Comparison of different multipliers}. Here we compare the effect of varying $\alpha$, and multiplier for $s_2$ on approximation error. In the legend ``$*z$'' is the multiplier for $s_2$, and the other number is for the varying $\alpha$. We can see that for $z=2$ and $\alpha=1.5$ the approximation works for both the datasets.}
    \label{fig:compare_z_a}
\end{figure*}



\section{Code \& Compute Resources}\label{app:code}

\textbf{Code.} The experiments of WMD are done partially in Matlab and Python. The source code of the WME experiments are taken from \url{https://github.com/IBM/WordMoversEmbeddings} \cite{wu2018word}. This uses a backend of C Mex for fast computation of Earth mover's distance\footnote{code from: \url{https://robotics.stanford.edu/~rubner/emd/default.htm}} \cite{rubner2000earth}. SVM is implemented using LIBLINEAR \cite{fan2008liblinear}. The codes for BERT  is implemented by using the Transformers library made available in \cite{wolf2019huggingface}. The rest of the code uses Python.

The parent directory of the anonymized codes can be found in Github\footnote{\url{https://github.com/archanray/approximate_similarities.git}}. The code base is split up into three parts: 1. WME: contains all files for experiments of approximating WMD matrices, 2. matrix\_approximations: contains all files for cross-encoder experiments and 3. cross\_doc\_cor: contains all files for entity and event conference experiments. Each folder is associated with a README.md to help run the codes.

\noindent\textbf{Compute resources.} For fast computation and parallel execution we used a server of 110 compute nodes with 28 core Xeon processors. Of course we only used partial resources for our execution. The maximum amount of RAM required was about 27GB for the largest dataset in WMD experiments and about 300MB for the smallest dataset in BERT experiments. The Transformers architecture was finetuned in a four NVIDIA GeForce GTX 1080 Ti GPUs with 2 Xeon Silver processors of 6 cores each.

\end{document}